\documentclass[10pt,journal,compsoc]{IEEEtran}
\usepackage[nocompress]{cite}
\usepackage{url}
\usepackage{soul}
\usepackage{bigfoot}
\usepackage{graphicx}
\usepackage{caption}
\usepackage{amsmath}
\usepackage{amssymb}
\usepackage{amsfonts}
\usepackage{natbib}
\usepackage{mwe}
\usepackage{hyperref}
\usepackage{booktabs}  
\usepackage{array}     
\usepackage{floatrow}
\usepackage{multirow}
\usepackage{floatrow}
\usepackage{adjustbox}   
\usepackage[dvipsnames,table]{xcolor}
\usepackage{bm}
\usepackage{subcaption}
\usepackage{tabularx} %
\usepackage{rotating} %
\usepackage{makecell}
\begin{document}
\newcommand{\etal}{{~et al.}}
\newcommand{\rev}[1]{\textcolor{red}{\st{#1}}} %
\newcommand{\salvit}{SalViT360}
\newcommand{\avsal}{SalViT360-AV}
\newcommand{\bounvr}{YT360-EyeTracking}
\newcommand{\nparticipants}{102}
\definecolor{bostonuniversityred}{rgb}{0.8, 0.0, 0.0}
\definecolor{darkspringgreen}{rgb}{0.01, 0.75, 0.24}

\definecolor{niceorange}{RGB}{255, 214, 171} 
\definecolor{niceblue}{RGB}{192, 231, 237} 
\definecolor{nicegreen}{RGB}{227, 247, 202} 
\definecolor{niceblue2}{RGB}{228, 234, 241}

\title{Spherical Vision Transformers for Audio-Visual Saliency Prediction in 360$\bm{^\circ}$ Videos}
\definecolor{revision}{HTML}{000000}
\newcommand{\revise}[1]{\textcolor{revision} {#1}}
\newcommand{\erkut}[1]{\textcolor{000000} {Erkut: #1}}

\author{Mert~Cokelek\IEEEauthorrefmark{1}, Halit Ozsoy\IEEEauthorrefmark{1}, Nevrez Imamoglu, Cagri Ozcinar, Inci Ayhan, Erkut Erdem, Aykut Erdem%
\IEEEcompsocitemizethanks{
\IEEEcompsocthanksitem \IEEEauthorrefmark{1} These authors contributed equally to this work.
\IEEEcompsocthanksitem M. Cokelek is with the Department
of Computer Science and Engineering, Koç University, Istanbul,
TR, 34450.\protect\\

E-mail: mcokelek21@ku.edu.tr
\IEEEcompsocthanksitem H. Ozsoy is with the Department of Psychology, Boğaziçi University, Istanbul, TR, 34342.%
\IEEEcompsocthanksitem N. Imamoglu is with the National Institute of Advanced Industrial Science and Technology (AIST), Intelligent Platforms Research Institute, Tokyo, JP. N. Imamoglu is also with AIST, CNRS-AIST Joint Robotics Laboratory, Tsukuba, JP. 
\IEEEcompsocthanksitem C. Ozcinar is with MSK.AI, London, UK.
\IEEEcompsocthanksitem I. Ayhan is with the Department of Psychology, Boğaziçi University University, Istanbul, TR, 34342.
\IEEEcompsocthanksitem E. Erdem is with the Department of Computer Engineering, Hacettepe University, Ankara, TR, 06800.
\IEEEcompsocthanksitem A. Erdem is with the Department of Computer Science and Engineering, Koç University, and KUIS AI Center, Istanbul, TR, 34450.

}
}

\IEEEtitleabstractindextext{%
\begin{abstract}
Omnidirectional videos (ODVs) are redefining viewer experiences in virtual reality (VR) by offering an unprecedented full field-of-view (FOV). This study extends the domain of saliency prediction to 360$^\circ$ environments, addressing the complexities of spherical distortion and the integration of spatial audio. Contextually, ODVs have transformed user experience by adding a spatial audio dimension that aligns sound direction with the viewer's perspective in spherical scenes. Motivated by the lack of comprehensive datasets for 360$^\circ$ audio-visual saliency prediction, our study curates YT360-EyeTracking, a new dataset of 81 ODVs, each observed under varying audio-visual conditions. Our goal is to explore how to utilize audio-visual cues to effectively predict visual saliency in 360$^\circ$ videos. Towards this aim, we propose two novel saliency prediction models: \salvit, a vision-transformer-based framework for ODVs equipped with spherical geometry-aware spatio-temporal attention layers, and \avsal, which further incorporates transformer adapters conditioned on audio input. Our results on a number of benchmark datasets, including our YT360-EyeTracking, demonstrate that SalViT360 and SalViT360-AV significantly outperform existing methods in predicting viewer attention in 360$^\circ$ scenes. Interpreting these results, we suggest that integrating spatial audio cues in the model architecture is crucial for accurate saliency prediction in omnidirectional videos. Code and dataset will be available at: \href{https://cyberiada.github.io/SalViT360/}{\texttt{https://cyberiada.github.io/SalViT360/}}.%

\end{abstract}

}
\maketitle

\IEEEdisplaynontitleabstractindextext
\IEEEpeerreviewmaketitle

\section{Introduction}\label{introduction}
\IEEEPARstart{T}{he} rapid proliferation of virtual reality (VR) and multimedia streaming platforms has led to an unprecedented surge in the popularity and usage of omnidirectional or 360$^\circ$ videos (ODVs). These videos offer viewers a fully immersive experience by providing a complete field-of-view (FOV), fundamentally reshaping the way users interact with visual content. Accurately predicting viewer attention, termed \textit{visual saliency prediction}, is increasingly critical in optimizing various multimedia tasks such as video compression~\cite{7962230,mishra2021multi,8692638,zhu2018spatiotemporal}, perceptually-driven super-resolution \cite{hou2022multi}, quality assessment~\cite{guan2017visual, yang2019sgdnet, zhu2021saliency,qiu2021blind,jabbari2023st360iq}, and adaptive streaming techniques like foveated rendering in VR environments \cite{patney2016towards}. The success of these applications hinges on a precise understanding of where users naturally fixate their attention during video viewing.

Predicting visual saliency in ODVs, however,  presents unique challenges not encountered with traditional 2D video content. The inherent spherical geometry of ODVs introduces significant spatial distortions when conventional projection methods, such as equirectangular projection (ERP) or cubemaps, are employed. ERP, while computationally simple, severely distorts content near the poles, and cubemaps introduce discontinuities at the edges, both negatively impacting model prediction accuracy. To overcome these challenges, in our recent work, we proposed SalViT360~\cite{cokelek2023spherical}, a vision transformer (ViT)~\cite{dosovitskiy2020image} based model explicitly tailored for omnidirectional videos. SalViT360 uniquely addressed spherical distortions using locally undistorted tangent images generated through gnomonic projection~\cite{eder2020tangent}, and incorporated spherical geometry-aware spatio-temporal transformer attention. This model established state-of-the-art performance for visual-only saliency prediction in ODVs. Though, a critical limitation remained: the absence of audio modality integration, despite its significant role in guiding viewer attention in immersive scenarios.

\begin{figure}[!t]
    \includegraphics[width=\textwidth]{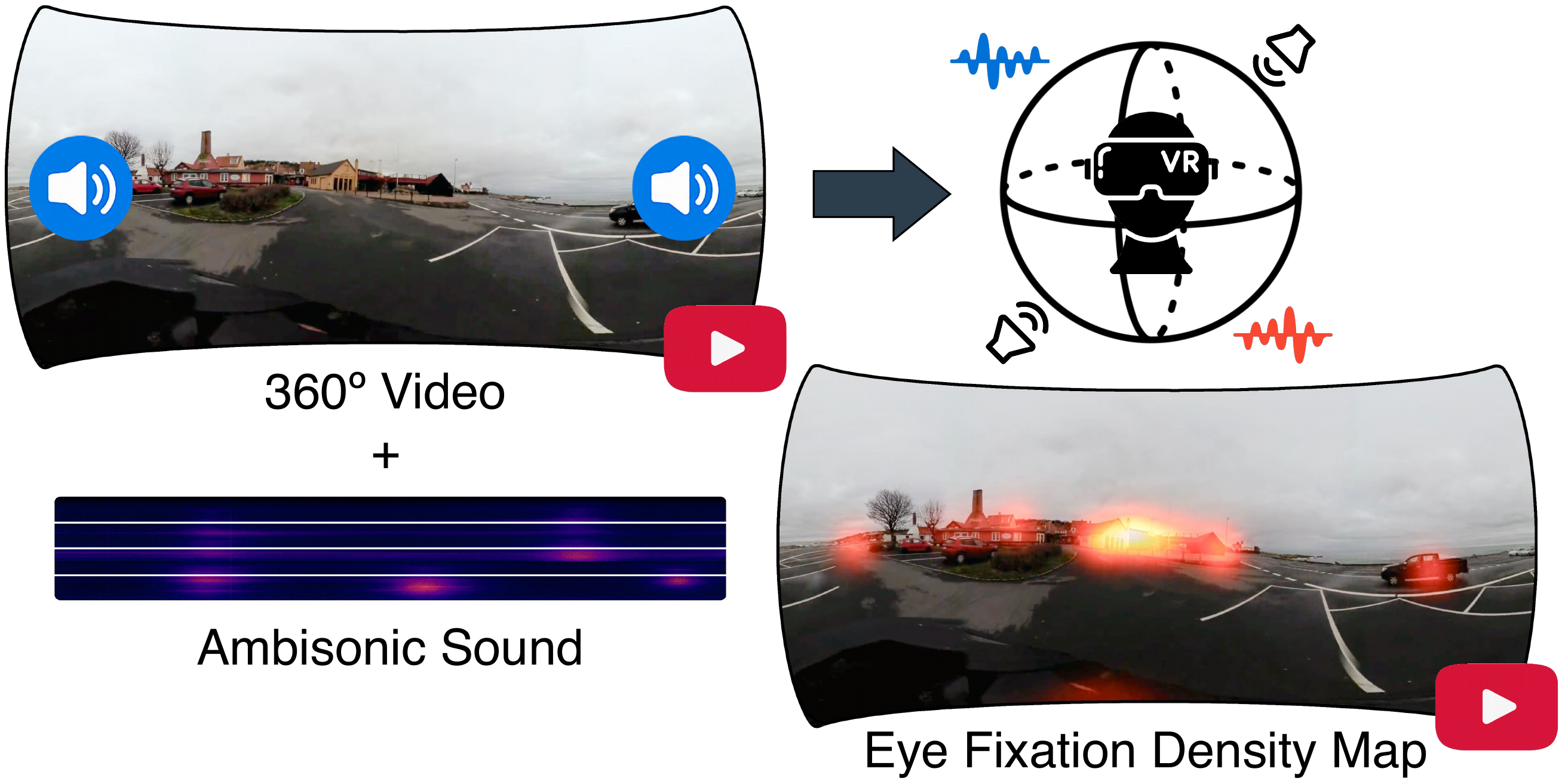}
    \caption{\textbf{Audio-Visual Saliency in 360$^\circ$ Videos.}  Illustration of how spatial audio cues influence visual attention in omnidirectional videos. In this example, spatial audio highlights salient regions by directing viewer attention towards audio-emitting objects such as a passing car and birds singing in the trees, emphasizing the necessity of integrating audio modalities into saliency prediction models.}
    \label{fig:teaser}
\end{figure}

Indeed, human perception naturally and continuously integrates auditory and visual stimuli. Numerous studies confirm that audio cues significantly modulate visual attention, guiding eye movements and enhancing viewer immersion \cite{tsiami2020stavis, Tavakoli2019,min2020multimodal}. In ODVs, spatial audio provides directional auditory cues aligned with visual elements, profoundly influencing where viewers focus their attention. Despite its recognized importance, existing computational models and datasets rarely consider audio-visual interactions comprehensively. To illustrate and emphasize the impact of spatial audio, we introduce an illustrative example in Fig.~\ref{fig:teaser}, which demonstrates how integrating spatial audio significantly alters visual attention patterns in omnidirectional contexts.

Motivated by the significant yet under-explored influence of spatial audio on visual attention, we present a novel, large-scale dataset named YT360-EyeTracking. This dataset is explicitly curated for audio-visual saliency prediction in ODVs, offering extensive annotations across various audio conditions, including mute, mono, and spatial (ambisonic) audio. YT360-EyeTracking addresses existing limitations by providing a robust foundation to systematically investigate how different audio modalities affect viewer gaze behavior in omnidirectional video scenarios.

In addition, we introduce SalViT360-AV, a novel audio-visual saliency prediction framework tailored specifically for 360$^\circ$ content. SalViT360-AV significantly expands upon our earlier SalViT360 model by incorporating a parameter-efficient transformer-adapter architecture inspired by AdaptFormer \cite{chen2022adaptformer}. This enables effective integration of spatial audio cues into the transformer-based visual saliency prediction pipeline. By capturing complex interactions between audio and visual stimuli, SalViT360-AV achieves substantially improved prediction accuracy, setting a new state-of-the-art benchmark for audio-visual saliency prediction in omnidirectional contexts.

In summary, our contributions to the domain of audio-visual saliency prediction for 360$^\circ$ videos are as follows:
\begin{itemize}
    \item We introduce YT360-EyeTracking, the first large-scale dataset explicitly designed for audio-visual saliency prediction in ODVs. It uniquely enables the systematic analysis of audio-visual interactions under varying audio modalities (mute, mono, ambisonics).

    \item We propose SalViT360-AV, a novel audio-visual saliency prediction framework that innovatively extends our previous visual-only model, SalViT360, by incorporating spatial audio through transformer adapters. This approach efficiently captures intricate audio-visual interactions, significantly improving model accuracy.

   \item We conduct comprehensive experiments across four benchmark datasets, including our newly introduced YT360-EyeTracking dataset, demonstrating that SalViT360-AV achieves state-of-the-art performance in predicting viewer attention, effectively highlighting the critical role of spatial audio integration in saliency prediction tasks.
\end{itemize}

\section{Background and Related Work}\label{relatedwork}
Saliency prediction models aim to predict the eye fixations of humans exploring visual stimuli by producing saliency maps resembling ground-truth maps. The ground-truth annotations are collected from subjects with eye-tracking hardware and represented as fixation and saliency maps. Fixation maps contain the actual eye gaze data of humans. Fixations are binary for each pixel in the scene, corresponding to whether an observer has fixated on that location or not. Saliency maps are heatmaps generated by convolving the fixation maps with a Gaussian kernel to create a continuous measure of saliency or likelihood that a location attracts human attention. In this section, we provide an overview of 360$^\circ$ video saliency prediction models and datasets.

\subsection{Overview of Saliency Prediction Models}
Saliency prediction models can be divided into two categories based on the scene characteristic, namely, \textit{static} and \textit{dynamic} saliency prediction. Static saliency considers input in the image level and ignores temporal information between frames. Conversely, dynamic saliency includes both spatial and temporal cues in the videos. For $360^\circ$ saliency prediction, previous works primarily focused on addressing the representation of $360^\circ$ scenes, with each method trying to balance representative power and computational complexity. Nevertheless, most works ignore the audio modality, which contains essential stimuli that drive the human visual attention system. First, we review the $360^\circ$ static and dynamic saliency literature and give an overview of the audio-visual saliency models for 2D and $360^\circ$ domains.

\vspace{0.15cm}\noindent\textbf{360$\bm{^\circ}$ Static Saliency Prediction.}
Chao et al.\cite{chao2018salgan360} introduced SalGAN360, utilizing cubemap projection to reduce ERP distortion. They fine-tuned the SalGAN model\cite{Pan_2017_SalGAN} for cubemap faces, then reprojected the outcomes to ERP for final predictions. While this method lessens local distortion, it faces edge discontinuities due to CNNs' inability to recognize adjacency across cube faces, leading to prediction errors at edges. To address this, the authors suggest employing multiple cubemap sets with horizontal and vertical shifts, resulting in $81$ projections per image, which significantly increases computational demand. In the follow-up work MV-SalGAN360~\cite{9122430}, the authors extended SalGAN360 with multi-view fusion. Unlike the single-scale cubemap projection, MV-SalGAN360 uses three scales. The first scale corresponds to the cubemap of six faces with $90^\circ$ FOV each. The second scale consists of five faces, each with an FOV of $120^\circ$. The last scale is the original ERP, with $180^\circ\times360^\circ$ FOV. While the multi-view representation is important for local-global scene encoding, MV-SalGAN360 requires training a separate set of parameters for each level, which is computationally expensive.
\begin{table*}[t!]
    \centering
    \begin{adjustbox}{width=1\textwidth}
        \begin{tabular}{l@{$\;\;$}c@{$\;\;$}c@{$\;\;$}c@{$\;\;$}c@{$\;\;$}c@{$\;\;$}c@{$\;\;$}c@{$\;\;$}c}
            \toprule
            \textbf{Dataset} & 
            \rotatebox{0}{\textbf{\shortstack{Num. \\Subjects}}} & 
            \rotatebox{0}{\textbf{\shortstack{Num. \\Videos}}} & 
            \rotatebox{0}{\textbf{\shortstack{Avg. \\Dur. (s)}}} & 
            \rotatebox{0}{\textbf{\shortstack{Frame \\Rate}}} & 
            \rotatebox{0}{\textbf{\shortstack{Min. \\Res.}}} & 
            \rotatebox{0}{\textbf{\shortstack{Audio\\ Categories}}} & 
            \rotatebox{0}{\textbf{\shortstack{Audio\\ Modality}}} &
            \rotatebox{0}{\textbf{\shortstack{Color\\ Modality}}} \\
            \midrule
            Salient360!~\cite{david2018dataset}       & 57 & 20 & 20s & 25-30 & 4K & n/a & Mute & Colored\\
            PVS-HMEM~\cite{xu2018predicting}          & 58 & 76 & 30s & 24-60 & 2K & n/a & Mute & Colored\\
            360-EM Dataset~\cite{agtzidis2019dataset} & 13 & 14 & 60s & 25-60 & 4K & Misc. & Mono & Colored\\
            VR-EyeTracking~\cite{xu2018gaze}    & 45 &208 & 30s & 24-60 & 4K & Speech, Music, Vehicle, Noise & Mono & Colored\\
            SVGC-AVA~\cite{yang2023svgc}          & 63 & 57 & 28s & 25-50 & 2K & Speech, Music, Sports & Ambisonics & Colored\\
            PAV-SOD~\cite{zhang2023pav}          & 40 & 67 & 30s & 25-60 & 4K & Speech, Music, Misc. & Ambisonics & Colored\\
            360AV-HM~\cite{chao2020audio}          & 45 & 15 & 25s & 25-60 & 4K & Speech, Music, Environment & Mute, Mono, Ambisonics & Colored\\
            AVS-ODV~\cite{11003418} & 60 & 162 & 15s & 30 & 8K & Speech, Music, Vehicle & Mute, Mono, Ambisonics & Colored\\
            \midrule
            {\bounvr}           & {102} & {81} & {30s} & {24-30} & {4K} & {Speech, Music, Vehicle} & {Mute, Mono, Ambisonics}  & {Colored, Grayscale}\\
            \bottomrule
        \end{tabular}
    \end{adjustbox}
    \caption{\textbf{A Comparison of Existing 360$^\circ$ Video Saliency Datasets.} Our YT360-EyeTracking dataset contains a large number of sequences with eye-tracking data collected from the highest number of subjects for both mute, mono, and ambisonics audio modalities.}
    \label{tab:dataset-comparison}
\end{table*} 

Dahou\etal~\cite{dahou2021atsal} proposed ATSal, a two-stream model to compute global and local saliency in $360^\circ$ videos. They use global prediction as a rough attention estimate, and the local stream on cube faces predicts local saliency. The local predictions are projected back to ERP and linearly combined with the global map for the final saliency prediction.
Zhang et al.~\cite{Zhang_2018_ECCV} developed a spherical U-Net for saliency prediction, applying modified kernels directly on ERP in spherical coordinates. These convolution kernels adapt to their location on the sphere, with dynamic receptive fields and a parameter-sharing approach for translation equivariance. However, computing a unique spherical crown for each convolution increases the model's runtime complexity. Djilali\etal~\cite{Djilali_2021_ICCV} used a self-supervised pre-training based on learning the association between several different views of the same scene and trained a supervised decoder for $360^\circ$ saliency prediction as a downstream task. Their framework maximizes the mutual information between different views on the spherical image, and the decoder is trained with these embeddings. Although their approach can model the global relationship between viewports, it ignores the crucial temporal dimension for video understanding.

Zhu et al.\cite{zhu2018spic} suggested a framework for generating separate head and eye saliency maps for static panoramas using some hand-crafted features and graphical models. Their follow-up~\cite{zhu2020tmm} extended this pipeline by combining visual equilibrium, uncertainty, and object cues to refine head and eye saliency predictions. A deep reinforcement learning agent was later introduced in~\cite{zhu2020acmtmcca}, encoding viewport content (CNN features, spherical coordinates, visited maps) and employing a free-energy-based reward to predict head movement trajectories in static 360$^\circ$ images.

\vspace{0.15cm}\noindent\textbf{360$\bm{^\circ}$ Dynamic Saliency Prediction.}
Cheng\etal~\cite{cheng2018cube} propose cube-padding to address discontinuities at cube face boundaries, facilitating information transfer across faces through overlapping regions. Particularly for CNN models, the information exchange increases with deeper layers. The model also utilizes ConvLSTMs to capture temporal features. 
Qiao\etal~\cite{qiao2020viewport} found that eye fixation distribution varies with viewport locations and treated $360^\circ$ video saliency prediction as a multi-task challenge, aiming to predict global saliency on the sphere and local saliency within viewports. This insight into fixation biases based on viewport locations inspired us to incorporate \textit{spherical position information} into our models. Zhu \etal~\cite{zhu2022ieeetcsvt} fuse RGB and motion cues over high-clarity viewports with a graph-based approach to predict saliency, demonstrating that combining appearance and dynamic information at select viewports improves temporal consistency. Yun et al.~\cite{yun2022panoramic} developed PAVER, utilizing deformable convolutions to correct spherical distortion, assigning specific offsets per convolution. This setup allows for spatial and temporal self-attention in vision transformers using undistorted patch features.

Several recent works attempt to reduce spherical distortion through projections or modified kernels while performing dense prediction tasks. For instance, Eder\etal~\cite{eder2020tangent} introduced tangent image representations using gnomonic projection to convert spherical images into multiple overlapping patches, each tangent to an icosahedron face. Their method, which also inspired our approach, addresses distortion at a local level but lacks explicit modeling of global interactions across tangent viewports. Similarly, Cheng\etal~\cite{cheng2018cube} utilize cube projections but do not incorporate mechanisms to capture spatial dependencies across cube faces.

While such techniques do facilitate localized predictions, they fall short in capturing global saliency cues critical for comprehensive scene understanding. In particular, Zhu\etal~\cite{zhu2022ieeetcsvt} and SalViT360 do process the whole field-of-view by integrating multiple partial views, but their fusion strategies differ: Zhu et al. rely on late or local fusion of viewport-specific outputs, whereas SalViT360 explicitly models long-range dependencies across all viewports at each layer. Panoramic Vision Transformer~\cite{qiao2020viewport} does perform spatio-temporal aggregation via deformable convolutions, but its windowed attention remains restricted to local regions. In contrast, our method enhances the tangent image approach by integrating spatial and temporal attention across viewports, allowing the model to reason globally while maintaining distortion-aware local representations. We note, however, that Yun\etal~\cite{yun2022panoramic} perform spatio-temporal modeling through spatially deformable convolutions.

In this work, we propose using tangent images to process undistorted local viewports and develop a transformer-based model, \salvit, to learn their global and spatio-temporal association for 360$^\circ$ video saliency prediction. Our transformer module is inspired by TimeSformer~\cite{gberta_2021_ICML}, which is developed for 2D videos. TimeSformer proposes Divided Space-Time Attention, which computes spatio-temporal attention in two stages to optimize for model complexity. We present Viewport Spatio-Temporal Attention, which extends the Divided Space-Time Attention idea to $360^\circ$ videos.

\vspace{0.15cm}\noindent\textbf{2D Audio-Visual Saliency.} Min\etal~\cite{7457921} proposed one of the earliest audio-visual saliency prediction methods, emphasizing the role of audio in guiding gaze. Their method computes separate spatial, temporal, and audio attention maps where the audio attention is generated through canonical correlation analysis between motion patterns and audio features—to produce an integrated audio-visual saliency map. Subjective eye-tracking experiments confirmed that this fusion improves prediction quality over visual-only models.
Tavakoli\etal~\cite{Tavakoli2019} proposed DAVE, an audio-visual encoder framework for 2D video saliency prediction. DAVE comprises a two-stream encoder architecture for audio and visual modalities. The video clips and audio mel-spectrograms are encoded by 3D ResNets in each stream, concatenated in the feature space, and then decoded with a 2D CNN to obtain the saliency predictions. Tsiami\etal~\cite{tsiami2020stavis} introduced STAViS, a spatio-temporal audio-visual saliency model that also adopts a two-stream CNN-based encoder. They further proposed a bilinear audio localization module that fuses the audio and visual features before decoding them into saliency maps.

Min\etal~\cite{min2020multimodal} presented the MMS model for fusing audio-visual cues under the assumption of high audio-visual correspondence. Their framework emphasizes the motion-consistent audio sources and performs late fusion with existing saliency maps, though it is less effective when visual and auditory cues diverge. Chen\etal~\cite{chen2021audiovisual} proposed an alternative architecture for two-stream feature extraction, semantic interaction, and their fusion for auditory and visual inputs. Semantic interaction brings the audio-visual features from lower and higher levels to the same dimensionality, demonstrating superior performance than fusing the features from only the final layers.

Zhu et al.~\cite{Dandan2023} propose LAVS, which extracts visual features using a VGG-16 backbone with ConvLSTM and audio features via separable convolutions on MFCC, CFCC and CQT features. Cross-modal alignment is achieved through deep canonical correlation analysis, and the resulting audio map is fused with the visual map, yielding competitive accuracy with minimal computational overhead. More recently, Zhu \etal~\cite{10422835} proposed MTCAM, a Transformer-based weakly supervised model for audio-visual saliency prediction. MTCAM jointly models audio and visual cues using separate encoders and a unified transformer-based fusion mechanism. Xiong\etal~\cite{Xiong_2024_CVPR} introduced DiffSal, a diffusion-based audio-visual saliency model that unifies saliency learning with generative modeling. DiffSal jointly learns audio and visual representations through a multi-stage diffusion process and enables fine-grained and high-resolution prediction with improved temporal coherence.

\vspace{0.15cm}\noindent\textbf{360$\bm{^\circ}$ Audio-Visual Saliency.} 
Spatial audio has been increasingly recognized as a critical cue for saliency prediction in omnidirectional videos (ODVs), particularly due to its alignment with immersive VR experiences. Several recent works have explored this direction.

Chao\etal~\cite{chao2020towards} proposed a two-stream pipeline where ERP frames are projected to padded cube maps and processed via a 3D ResNet, while mono audio mel-spectrograms are encoded separately. The resulting features are fused and enhanced with an audio energy map (AEM) that estimates the direction of arrival based on first-order ambisonics. Cokelek\etal~\cite{cokelek2021leveraging} introduced an unsupervised, audio-only method called MCSR, which computes directional audio saliency via mel-cepstrum features across six primary directions, projecting this as a bias onto ERP-based maps.
More recently, Zhu~\etal~\cite{10109890} proposed AVPS, a unified saliency prediction model integrating both spatial audio localization and content-level audio features. Their approach combines improved group-equivariant CNNs with E3D-LSTMs to capture visual dynamics and computes spatial audio localization via AEM, further complemented by SoundNet-based attribute-aware audio embeddings. In another work, Zhu~\etal~\cite{11003418}
presented OmniAVS, a fully convolutional framework that introduces a learnable spatial audio map (SAM) based on ambisonic energy. OmniAVS combines this SAM with visual features using an attention-based fusion module that models temporal and directional saliency consistency. Yang\etal~\cite{yang2023svgc} introduced \mbox{SVGC-AVA}, a graph-based model leveraging spherical vector-based graph convolution and audio-visual attention fusion. Their model encodes spherical geometry and spatial audio jointly to predict fixations.

It is worth noting that the aforementioned recent works are concurrent with ours and some of them have not yet released code or pretrained models. As such, we are unable to provide direct quantitative comparisons. Nevertheless, we include all these works here to present a comprehensive overview of the current research landscape in 360$^\circ$ audio-visual saliency prediction.
Our SalViT360-AV model advances this research line by integrating spatial audio via transformer adapters conditioned on ambisonic input, allowing for dynamic, fine-grained cross-modal fusion within an end-to-end trainable architecture. Compared to previous efforts, it captures deeper semantic and spatial correlations, leading to superior or competitive performance on a large number of benchmark datasets.

\subsection{360${^\circ}$ Video Saliency Datasets}\label{bg:datasets}
Several datasets have been proposed for visual saliency prediction in omnidirectional videos (ODVs), capturing varying levels of realism, scale, and audio modality.

\vspace{0.15cm}\noindent\textbf{Salient360 dataset~\cite{david2018dataset}} is one of the earliest datasets to provide head movement (HM) and eye fixation (EM) annotations over ERP or cubemap projections. While it includes diverse content, it remain relatively small in size.

\vspace{0.15cm}\noindent\textbf{PVS-HMEM dataset~\cite{xu2018predicting}} includes head movement and eye-tracking data collected from 76 $360^\circ$ video sequences from 58 subjects. The videos vary between $10-80$~secs, with a minimum resolution of 2K and frame rate between $24- 60$fps. They includes indoor, outdoor, movie, and sports categories. The data collection is done in a muted modality. 

\begin{figure*}[!t]
    \centering
    \includegraphics[width=0.9725\textwidth, trim={10 1900 980 0}]{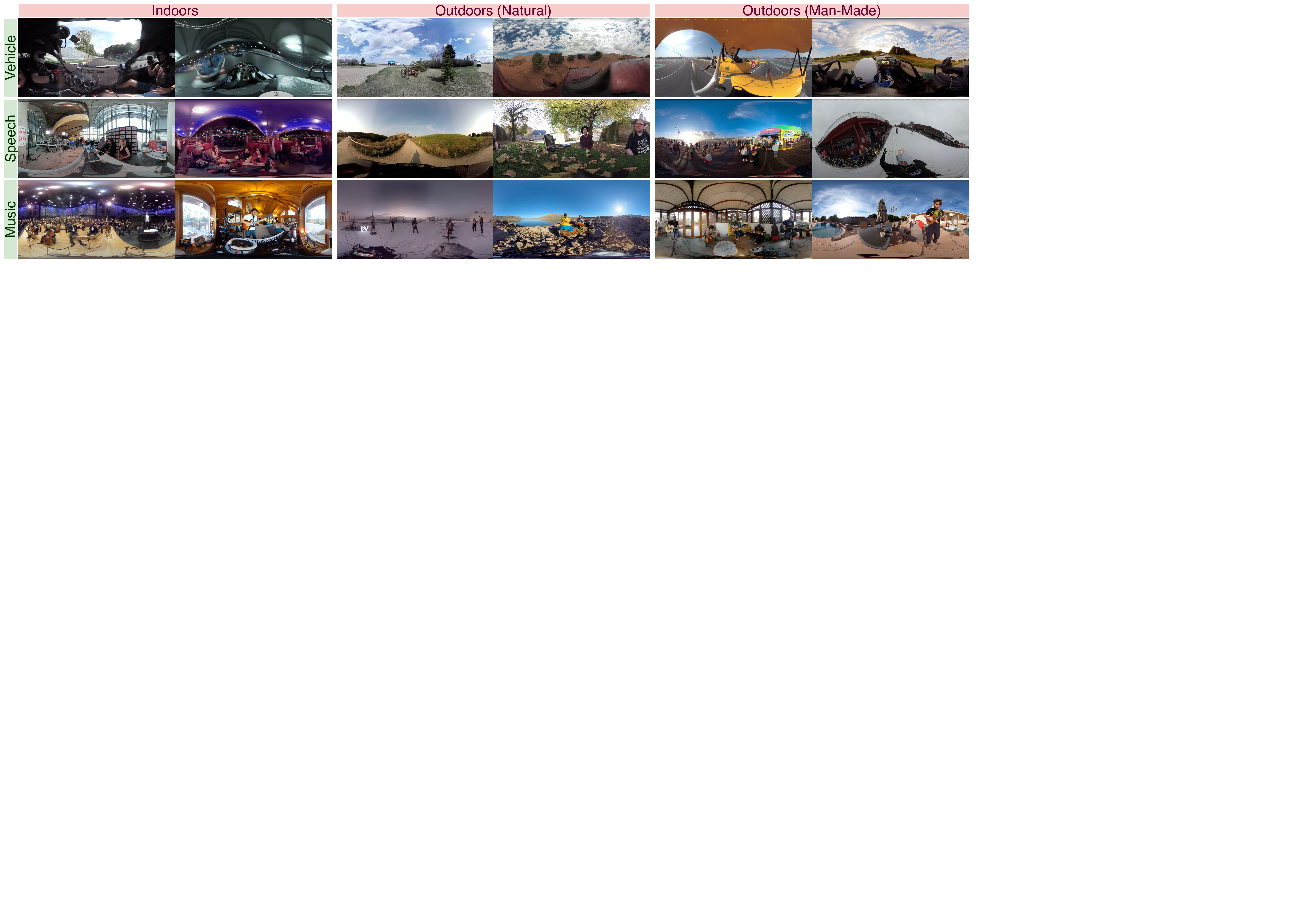}
    \caption{\textbf{Sample Input Frames from Our YT360-EyeTracking Dataset.} Rows and columns respectively represent different \textcolor{YellowGreen}{\textbf{audio}} and \textcolor{Maroon}{\textbf{visual}} categories.}
    \label{fig:dataset_overview}
\end{figure*}

\vspace{0.15cm}\noindent\textbf{VR-EyeTracking dataset~\cite{xu2018gaze}} consists of eye tracking data collected for 208 videos (134 train, 74 test) lasting between 20-60 seconds. The videos have a minimum resolution of 4K in ERP format with frame rates between 24-60 fps. There are 20 female and 25 male subjects aged between 20-24 years. At least 31 subjects view each video, with each subject watching around 35 videos. Videos are picked under various scene categories, including indoors, outdoors, sports, games, documentation, short movies, and music shows. Subjects have viewed the videos with mono audio modality, as in traditional 2D videos. Mono modality delivers sound sources equally from all directions. Hence, the viewers cannot be guided by directional sound sources. In contrast, $360^\circ$ videos can be delivered with spatial audio, where the sound sources are precisely located, and the perceived sound changes with head movement (roll, pitch, yaw). 

\vspace{0.15cm}\noindent\textbf{SVGC-AVA dataset~\cite{yang2023svgc}} is a recent benchmark proposed to evaluate audio-visual saliency in ODVs. It comprises 57 360$^\circ$ videos recorded with spatial audio, annotated with eye-tracking data from 63 subjects in free-viewing conditions, supporting both ERP and cubemap formats. The scenes include complex audio-visual events such as overlapping sounds and directionally distinct audio sources, providing a challenging benchmark for multimodal attention modeling.

\vspace{0.15cm}\noindent\textbf{360AV-HM dataset~\cite{chao2020audio}} consists of 15 ODVs with spatial audio lasting 25 seconds. Each video has a resolution of $3840\times1920$ in ERP format, with frame rates between $24-60$fps. The videos are collected from YouTube under three categories,  Conversation, Music, and Environment. The videos are viewed under mute, mono, and spatial audio settings by three subject groups, where each subject has only viewed the videos with one modality. Each video was viewed by fifteen subjects, and there are a total of 16 female and 29 male subjects aged between $21-40$ years.

\vspace{0.15cm}\noindent\textbf{AVS-ODV~\cite{11003418} dataset}, proposed concurrently with our work, includes 162 ODV clips spanning human activities, natural scenes, vehicles categories and provides saliency annotations under three audio conditions (mute, mono, spatial). The clips are sourced from YouTube and feature real-world audio-visual correlations. The dataset is recorded with 15 participants using a VR headset and includes a thorough psychophysical analysis of how different audio configurations influence gaze behavior. 

Together, these datasets illustrate the progression of 360$^\circ$ saliency research from smaller-scale, visual-only datasets to larger-scale benchmarks integrating multimodal annotations, particularly spatial audio. As detailed in Table~\ref{tab:dataset-comparison}, our YT360-EyeTracking dataset, as compared to the recently introduced SVGC-AVA\cite{yang2023svgc} and AVS-ODV\cite{11003418}, employs a significantly larger number of subjects. Additionally, YT360-EyeTracking explicitly explores the influence of audio modality by systematically providing annotations under three audio conditions (mute, mono, ambisonics), closely matching the concurrent AVS-ODV dataset in experimental conditions. But it also ensures a balanced gender representation among participants, featuring 53 female and 49 male subjects. These comparative strengths make our dataset particularly suitable for training deep-learning-based multimodal models such as SalViT360-AV, facilitating detailed analyses of audio's role in guiding visual attention in omnidirectional video scenarios.

\section{\bounvr~Dataset}

\begin{figure}[!t]
    \includegraphics[width=0.6\textwidth]{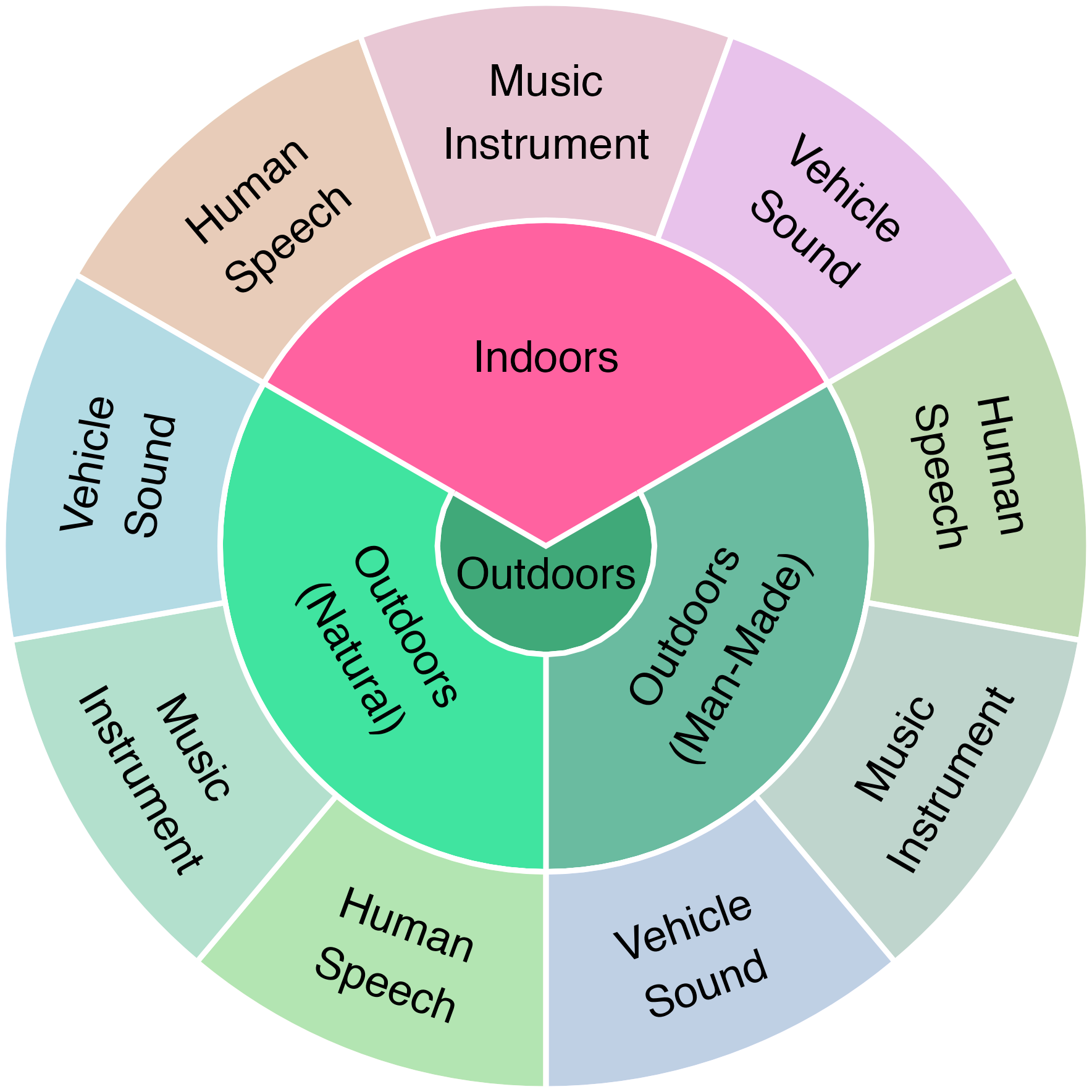}
    \caption{\textbf{Observation Sessions in \bounvr~by Category and Format.} The dataset includes 9 clips per scene-audio combo, with 7-29 participants viewing each audio/color version, averaging 16.77 participants per clip.}
    \label{fig:dataset_categories}
\end{figure}

\textbf{Overview.} We picked 81 videos from YouTube-360~\cite{morgado2020learning} by carefully considering nine audio-visual semantic categories and selecting nine videos per category, viewed under three audio conditions (mute, mono, spatial audio) and two color conditions (colored, grayscale\footnote{Including grayscale video allows isolating luminance-driven attention from chromatic effects, offering a clearer view of the role of color in visual saliency. Psychophysical studies have shown that color can influence gaze allocation and scene interpretation in meaningful ways~\cite{nuthmann16eye,frey08what}, making grayscale conditions a valuable complementary modality for exploring cross-modal and cross-condition saliency patterns. See~\cite{ozsoy2024msc} for related preliminary psychophysical analyses.}). The audio semantic categories are (1) speech, (2) music, (3) vehicle, and visual categories are (1) indoors, (2) outdoors-natural, and (3) outdoors-human-made. Sample frames from each audio and scene category are given in Fig.~\ref{fig:dataset_overview}. The distribution of samples for each audio and scene category is given in detail in Fig.~\ref{fig:dataset_categories}. Each video in our dataset contains dynamic scenes rather than static images and does not have cutscenes. The video clips last 30 secs, and have a resolution of 3840$\times$1920 pixels in ERP format, with frame rates between $24-30$fps. Each video is watched by at least 15 observers under specific audio-visual conditions, with each observer viewing the video only once. YouTube video IDs, segment start timestamps, raw fixation data, as well as data preparation and post-processing scripts, will be made available to the public.

\subsection{Stimuli}
\textbf{Videos.} All clips within the YT-360 dataset were extracted as 10-second segments from longer videos, and there are multiple consecutive clips from the same videos. After re-downloading the original videos and trimming the reference sequences to 30 seconds, we obtained our new stimuli for ODV saliency. Moreover, we chose to exclusively incorporate high-quality videos, with a minimum resolution of $3840\times1920$, and spatial audio modality. To facilitate the data filtering process while also keeping valuable semantic segmentation information in our new dataset, we deliberately chose the subset of the original sequences in YT-360.

\noindent\textbf{Data Filtering and Labeling.} We employed learning-based models \cite{chen2020vggsound,gemmeke2017audio,zhou2018places360} for initial video annotation, and selected the top 300 videos based on their annotation scores for a more detailed evaluation. For each of these videos, we manually labeled both the audio and scene categories. To streamline this process, we developed a website equipped with an intuitive interface specifically designed for video annotation.
The annotation interface prompted researchers to identify the predominant type of audio in the video. The options provided were: vehicle sound, musical instrument, human speech, human vocalizations, other, and noise. For the environmental context, the interface asked whether the video was recorded indoors or outdoors, with further distinctions made between natural and human-made outdoor environments. We also incorporated several control questions to improve the overall stimuli quality. These questions aimed to identify issues such as black screens, sudden scene changes, and the origin of the audio (whether it was captured live or added in post-production). Lastly, we evaluated the ability to localize sound sources within the videos and the presence of text overlays, which helped us eliminate any erroneous content from our dataset. Following the initial annotation and filtering process, we are left with videos divided into three audio (speech, music, vehicle) and three visual (indoors, outdoors-human-made, outdoors-natural) categories.

To determine the number of stimuli for the dataset, \textit{a priori} power analysis was conducted using G*Power version 3.1~\cite{faul2009gpower}. Results indicated that to achieve $95\%$ power for detecting a medium effect ($\mathit{f}=0.25$), at a significance criterion of $\alpha=.05$, the required sample size was $\mathit{N} = 63$ for a mixed model ANOVA (3$\times$3 between factors of sound and scene types, 3$\times$2 within factors of sound and color modalities) corresponding to at least $\mathit{n}=7$ videos per audio-visual category. Consequently, we chose nine representative videos for each audio-scene combination, resulting in a curated set of 81 videos, ensuring a diverse and representative dataset.

\noindent\textbf{Pre-processing.} Videos only available in higher resolutions were downsized to $3840\times1920$ pixels via FFmpeg \cite{tomar2006converting_ffmpeg} using bicubic interpolation to ensure consistency. In addition to the original first-order ambisonics, we altered all audio files to mono and mute versions by removing the audio channels. Moreover, we converted videos to grayscale by keeping only the Y (luma) component of the YUV format. As a result of these augmentations, we obtained a total of 486 distinct stimuli from 81 clips.

\subsection{Eye-Tracking Data Collection}
\begin{figure}[!t]
    \centering
    \includegraphics[width=\textwidth]{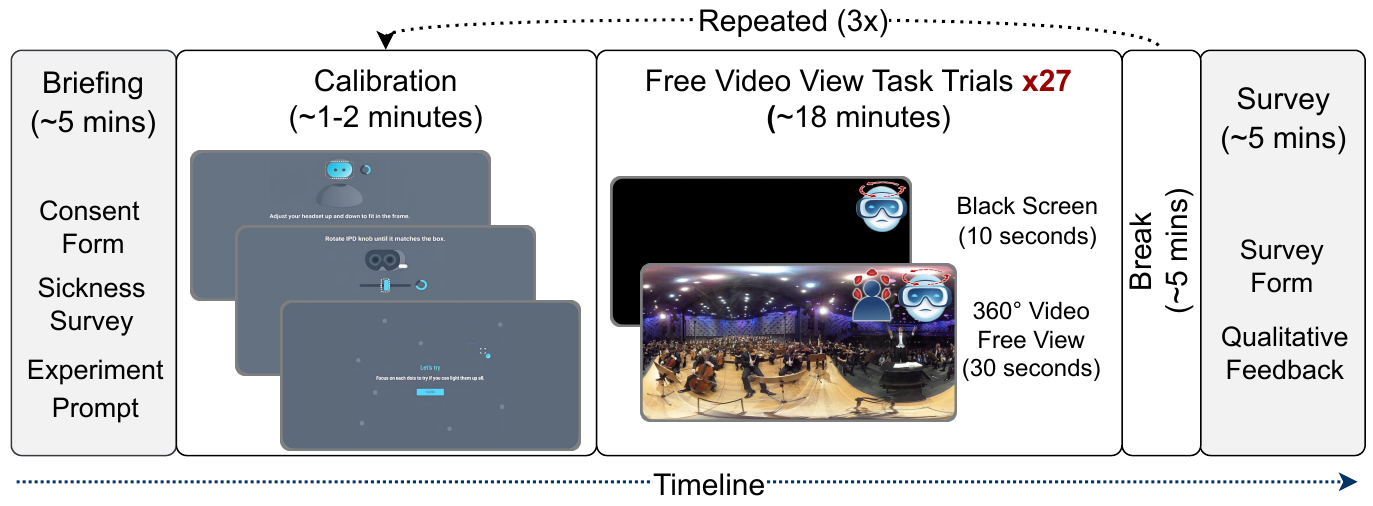}
    \caption{\textbf{\bounvr~Data Collection Procedure.} 
    Participants completed a free-viewing task across three sessions with 81 videos, incorporating a 5-min. break between sessions to lessen fatigue. Sessions started with eye calibration, followed by 27 audio-visual stimuli, each being a 30-sec. video clip in one of three pseudo-randomized audio (mute, mono, ambisonics) and color (grayscale, colored) conditions, separated by 10-sec. black screen intervals to reduce carry-over effects.}
    \label{fig:datacollectionsetup}
\end{figure}

\begin{figure*}[t]
    \centering
    \includegraphics[width=0.98\textwidth, trim={0 85 25 0}]{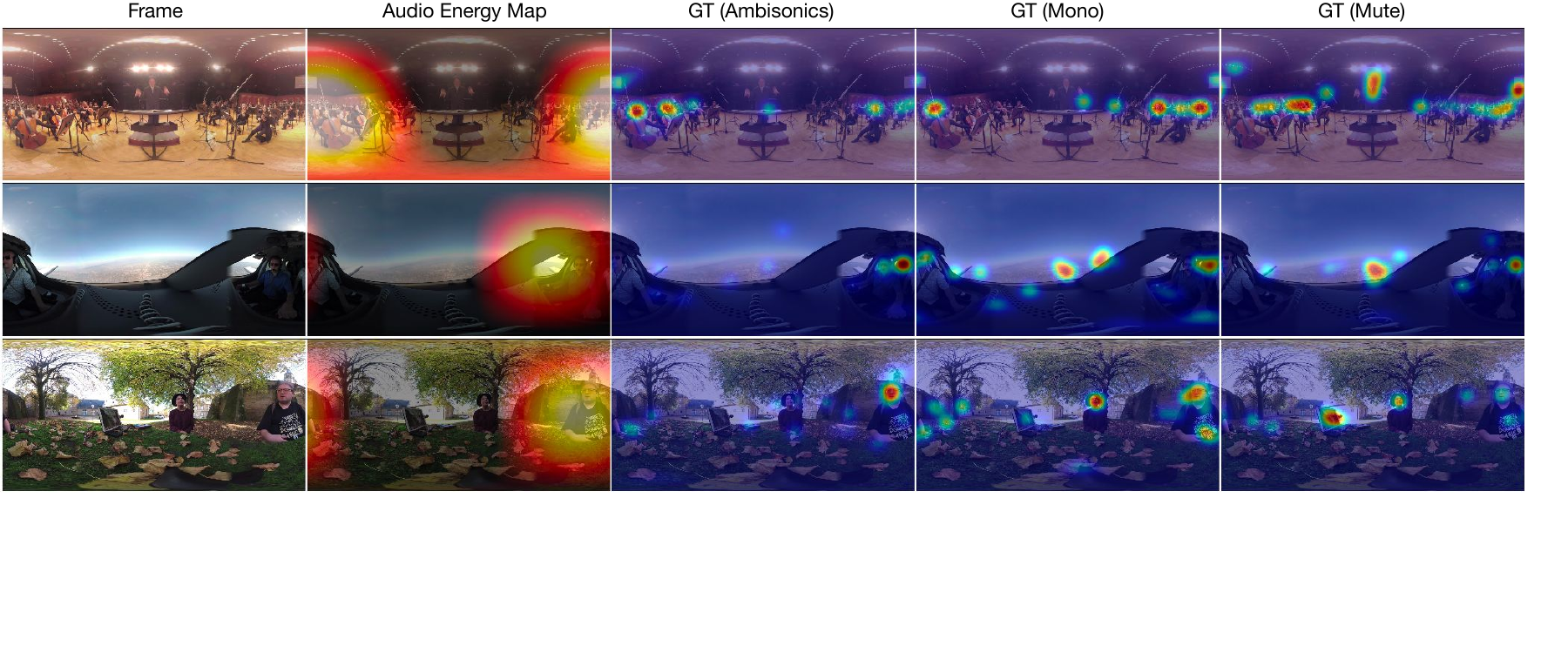}
    \caption{\textbf{Fixation Density Maps Collected from Three Subject Groups Under Mute, Mono, and Ambisonic Audio Conditions in Our Dataset.} The analysis underscores a progressive decrease in correlation between the fixation density maps (ground-truth saliency maps) and audio energy maps, moving from ambisonics to mute settings. This trend distinctly illustrates the influence of spatial audio on guiding participants' attention.}
    \label{fig:dataset_gtmaps}
\end{figure*}

As illustrated in Fig.~\ref{fig:datacollectionsetup}, our study employed a free-viewing task structured into three trials, with a 5-min. intermission between each to minimize participant fatigue. Each session began with eye calibration to ensure accuracy in tracking. Participants were then presented with a sequence of 27 stimuli in each trial, comprising a 30-second video clip accompanied by corresponding audio. To mitigate any carry-over effects, each stimulus was followed by a 10-second interval displaying a black screen. Overall, each video was viewed by each participant under one of six audio-visual conditions.  Fig.~\ref{fig:dataset_gtmaps} illustrates the variation in fixation density maps across these different audio modalities.

Before commencing the trials, we provided participants with guidelines to standardize the viewing experience. They were instructed to stand and view the 360° videos, allowing rotation but maintaining their position. Participants were informed about the lengths of trials, video clips, and intervening black screen periods. Although the specifics of the video content were not detailed, they were told to expect various presentations, including grayscale, colored, mute, and audiovisual formats, with an assurance of no distressing content to reduce surprises.

To examine the influence of audio and visual variables, a crossed experimental design
was utilized. This design ensured that each participant experienced every video clip under specified audio and color conditions, with the sequence of these presentations randomized to control for order effects. Consistency in the viewing experience was maintained by standardizing the initial viewing angle, resetting the participant's yaw rotation to 0 degrees at the onset of each stimulus presentation.

\noindent\textbf{Apparatus.} We used the HTC Vive Pro-Eye\footnote{\url{https://vive.com/sea/product/vive-pro-eye/overview/}} headset with a Tobii Eye Tracker\footnote{\url{https://tobii.com/products/integration/xr-headsets/}} for precise eye movement tracking. The virtual viewing environment was engineered in Unity3D\footnote{\url{https://unity.com/}}. To facilitate free movement and rotation without cable entanglement, the headset cable was connected to a ceiling-mounted extensible and adjustable mechanism.

\noindent\textbf{Participants.} The study engaged \nparticipants~participants (53 females and 49 males), aged 18 to 33, predominantly affiliated with Boğaziçi University, Istanbul, TR. All had normal or corrected-to-normal vision, with corrections via contact lenses or glasses where necessary. Participant anonymity and data confidentiality were upheld through the use of randomly assigned IDs. The research protocol was approved by the Boğaziçi University Ethics Coordinating Committee\footnote{\url{https://bogazici.edu.tr/en_US/Content/About_BU/Governance/Councils_Boards_and_Committees/Ethics_Committees}}, ensuring ethical compliance. Informed consent was obtained from all participants, who were briefed on the study's procedures and their right to withdraw at any time.

\subsection{Post-processing}
Initially, the experiment records were synchronized with the corresponding tracking data, providing gaze and head direction details for each stimulus per participant. Fixations were extracted using the I-DT algorithm, as described in \cite{agtzidis2019eq_idt}. Fixation parameters included a maximum dispersion of 1.5 visual angles and a minimum duration of 0.1 seconds. These fixation characteristics were then saved in comma-separated value text files. The chosen thresholds considered both the device's accuracy range of 0.5-1.1 degrees and established guidelines for I-DT thresholds \cite{david2023salient360, schuetz2022vive_accuracy, blignaut2009dispersion_thresh, salvucci2000fixation_extraction}. 

\begin{figure*}[!t]
    \centering
    \includegraphics[trim={0 375 70 0},clip,width=0.98\textwidth]{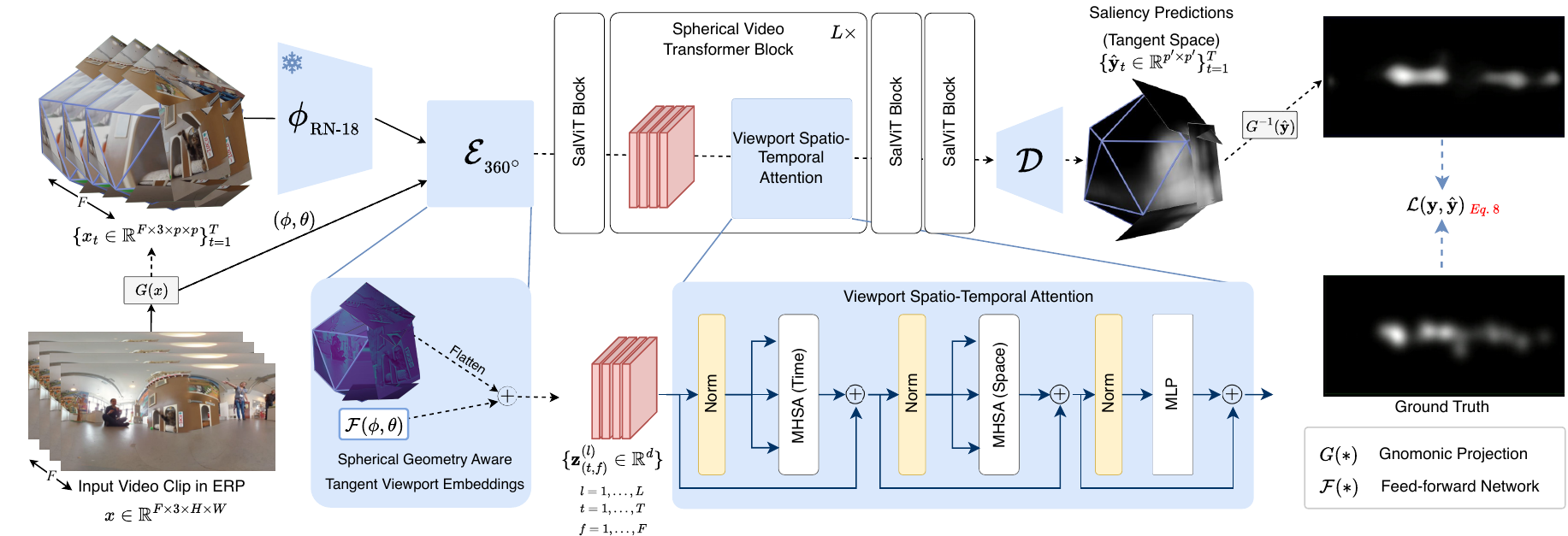}
    \caption{\textbf{Overview of \salvit.} The ERP video clip of $F$ frames (1) is projected to $F\times T$ tangent images per set (2). Each tangent image is encoded and fused with spherical-geometry-aware position embeddings (3) for the 360$^\circ$ video transformer to aggregate global information (4). The outputs are decoded into saliency predictions in tangent space (5), which are projected back to ERP, giving the final saliency map (6). In addition to the supervised loss, the model is trainable with $\mathcal{L}_{\text{VAC}}(P,\ P')$ to minimize the tangent artifacts (7). During test time, the model works with a single tangent set. For simplicity, only one set of tangent images is shown.}
    \label{fig:salvit_overview}
\end{figure*}

\section{Methodology}
In this section, we first describe \salvit, our vision-transformer-based 360$^\circ$ video saliency model. Then, we present \avsal, the unified audio-visual saliency prediction pipeline for 360$^\circ$ videos. \avsal~is a two-stream architecture for audio and visual modalities. Both streams are pre-trained and frozen, and we introduce transformer adapters conditioned on audio features for parameter-efficient fine-tuning of the video stream. 

\subsection{Spherical Vision Transformers for 360$^\circ$ Video Saliency Prediction}
In Fig.~\ref{fig:salvit_overview}, we present the overview of \salvit. We start with gnomonic projection~\cite{eder2020tangent} to obtain tangent images for each frame in the input video clip. The tangent images are passed to an encoder-transformer-decoder architecture. 
The image encoder is used to extract local features for each tangent viewport and reduce the input dimension for the subsequent self-attention stage. 
We map the pixel-wise angular coordinates to produce the proposed spherical geometry-aware position embeddings $\mathcal{F}(\phi,\theta)$ for the 360$^\circ$ transformer, enabling enriched spatial representations. 
The transformer utilizes our proposed Viewport Spatio-Temporal Attention (VSTA) to capture inter and intra-frame information across tangent viewports in a temporal window of $T=8$ frames. 
The transformed embeddings are then fed into a 2D CNN-based decoder, which predicts saliency on the tangent images. We then apply inverse gnomonic projection on the tangent predictions to obtain the final saliency maps in ERP. We propose an unsupervised consistency-based Viewport Augmentation Consistency Loss to mitigate the discrepancies after inverse gnomonic projection. The learnable parameters of the network are in tangent space, allowing us to leverage large-scale pre-trained 2D models (e.g., ResNet-18) for feature extraction while the rest of the network is trained from scratch.

\vspace{0.15cm}\noindent\textbf{Gnomonic Projection and Encoder.} We project the input ERP clip $x\in\mathbb{R}^{F\times3\times{H}\times{W}}$ to a set of tangent clips $\{x_\textit{t}\in\mathbb{R}^{F\times3\times p\times p}\}_{t=1}^T$, with $F$, $3$, $H$, and $W$ indicating the number of frames, channel dimension (RGB), height, and width of the video, respectively. These tangent images have a patch size of $p\times p=224\times224$ pixels. 
We select $T=18$ tangent images per frame and a $80^\circ$ field-of-view based on~\cite{li2022omnifusion}. After downsampling and flattening the encoder features, we obtain tangent feature vectors $\{{z_{\textit{t}}\in\mathbb{R}^{F\times d}}\}_{t=1}^T$.
In a parallel stream, we map the angular coordinates $(\phi, \theta)$ for each pixel of the tangent viewports to the feature dimension $d=512$ using an FC layer and element-wise add these embeddings with encoder features to obtain the proposed spherical geometry-aware tangent embeddings $\left\{\mathbf{z}_{\textit{t}}\in\mathbb{R}^{F\times d}\right\}_{t=1}^T$ that are used in the transformer. 

\begin{figure}[!t]
    \centering
    \includegraphics[width=\textwidth, trim={0 0 0 0}]{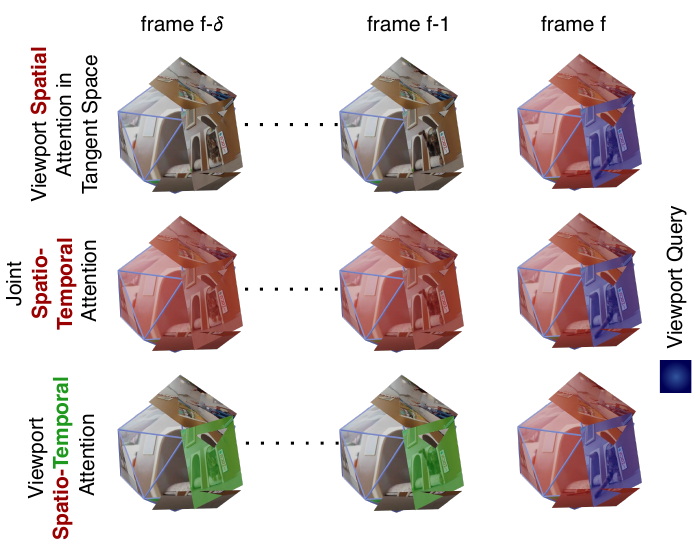}
    \caption{\textbf{Viewport Spatio-Temporal Attention (VSTA)} ({right}), compared to Viewport Spatial Attention (VSA) ({left}) and Joint Spatio-Temporal Attention ({middle}) approaches. {\color{bostonuniversityred}Red} and {\color{darkspringgreen}Green} viewports denote the self-attention neighborhood for each scheme.}
    \label{fig:vsta}
\end{figure}

\vspace{0.15cm}\noindent\textbf{Viewport Spatio-Temporal Attention.}\label{approach:vsta} While the pre-trained encoder extracts rich spatial features for each tangent image locally, aggregating the global context in the full FOV is essential for 360$^\circ$ scene understanding. We propose a self-attention mechanism on tangent viewport features to achieve this. However, since incorporating the temporal dimension of the videos increases the number of tokens, thus the computational complexity, inspired by the divided space-time attention mechanism in TimeSformer~\cite{gberta_2021_ICML}, we approximate spatio-temporal attention with two stages: we apply temporal attention (1) among the same tangent viewports from consecutive $F$ frames, then, spatial attention (2) among $T$ tangent viewports in the same frame. This reduces the overall self-attention complexity from $F^2\times{T^2}$ to $F^2 + T^2$, effectively capturing the required global context for 360$^\circ$ video analysis. Our {Viewport Spatio-Temporal Attention} for 360$^\circ$ videos illustrated in Fig.~\ref{fig:vsta} is defined as: 
\begin{equation}
    \begin{aligned}
        \text{VSTA}(\mathbf{z}_{(t, f)}^{(l)}) &= \text{VSA}(\text{VTA}(\mathbf{z}^{(l)}_{(t, f)}))\\
        \text{VTA}(\mathbf{z}_{(t, f)}^{(l)}) &= \text{SM}\left(\mathbf{q}^{(l)}_{(t, f)} \cdot \left\{\mathbf{k}^{(l)^\top}_{(t, f')}\right\}\right) \cdot \left\{\mathbf{v}^{(l)}_{(t, f')}\right\}_ {f'=1\dots F}\\
        \text{VSA}(\mathbf{z}_{(t, f)}^{(l)}) &= \text{SM}\left(\mathbf{q}^{(l)}_{(t, f)} \cdot \left\{\mathbf{k}^{(l)^\top}_{(t', f)}\right\}\right) \cdot \left\{\mathbf{v}^{(l)}_{(t', f)}\right\}_ {t'=1\dots T}\\
    \end{aligned}
\end{equation}
where $\mathbf{z}^{(l)}_{(t, f)}\in \mathbb{R}^{d}$ denote the tangent features of viewport $t$ in frame $f$ at $l$-th transformer block, $\mathbf{q}, \mathbf{k}, \mathbf{v}$ are the query, key, and value projections of $\mathbf{z}$, and $\text{SM}$ is the softmax operator. Attention heads and the scale multiplication of dot-product attention are not given for presentation clarity.

\begin{figure*}[!t]
    \centering
    \includegraphics[trim={0 375 70 0},clip,width=0.98\textwidth]{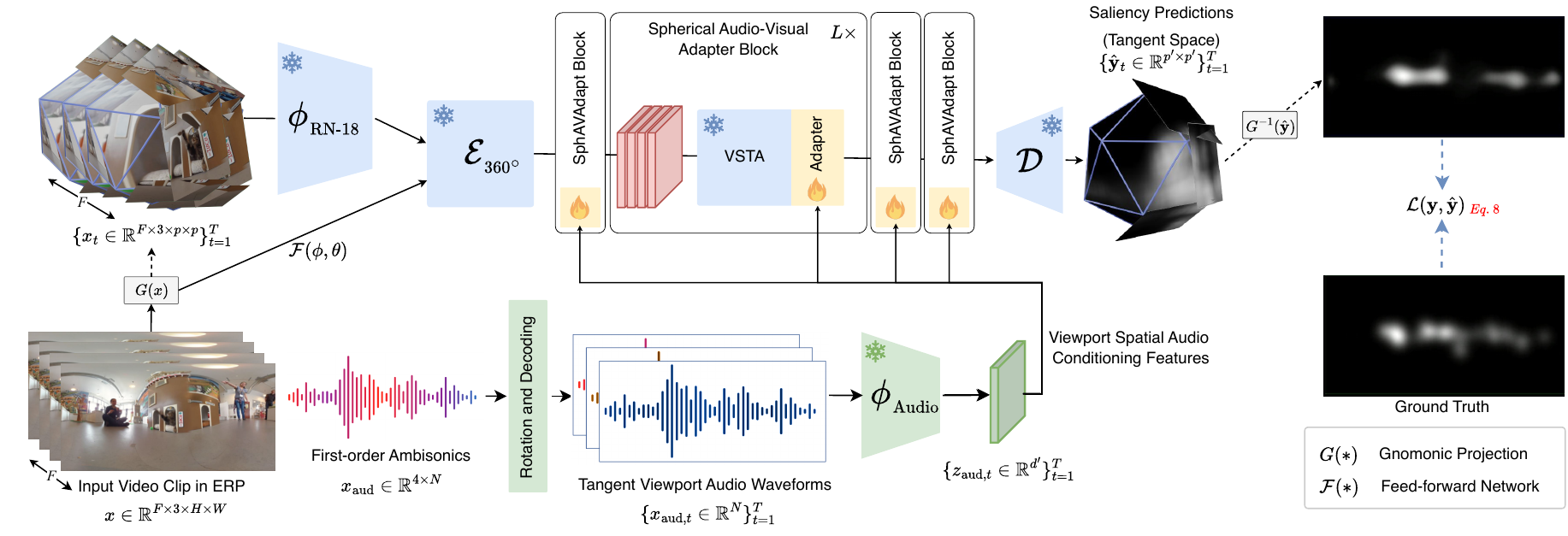}
    \caption{\textbf{Overview of Our Proposed \avsal~Pipeline.}
    We use the \salvit~model as the video saliency module {(top)}, and our implementation allows for any audio model as the audio backbone {(bottom)}. The audio stream takes input spatial audio waveforms $x_{\text{aud}}\in\mathbb{R}^{4\times N}$ encoded as first-order ambisonics in 4-channel B-format. To simulate what the subjects are hearing while looking at a particular location, we rotate the ambisonics depending on the angular coordinates $(\theta, \phi)$ for each tangent viewport $\{x_{\text{t}}\}_{t=1}^T$ (1). The rotated waveforms are mono, which enables us to use any pre-trained audio backbone for feature extraction (2). The extracted features are passed to the adapter layers in each upgraded transformer block (3) for audio-visual tuning. While the total number of parameters in the video pipeline is $37M$, the additional adapter layers require only $600k$ parameters for fine-tuning.} 
    \label{fig:avsal_av_overview}
\end{figure*}

\vspace{0.15cm}\noindent\textbf{Decoder.} The decoder comprises four upsample layers followed by $3\times3$ convolutions and normalization layers. For a set of tangent clips, it takes the skip connection of encoder and transformer features $\{\mathbf{\hat{z}}_{\textit{t}}\in\mathbb{R}^{512\times7\times7}\}_{t=1}^T$ of the last frame as input and outputs saliency prediction $\{\mathbf{\hat{y}}_{\textit{t}}\in\mathbb{R}^{p'\times p'}\}_{t=1}^T$ on tangent planes. The final ERP saliency maps are obtained by passing the tangent predictions to inverse gnomonic projection. While we aggregate global information among tangent images through the transformer, each tangent plane is predicted separately in the decoder. This causes discrepancies in the overlapping regions on ERP. Prior work on omnidirectional depth estimation~\cite{li2022omnifusion} proposes iterative refining to compensate for these artifacts, requiring the whole model to run twice per sample during training and inference. To address this, we propose Viewport Augmentation Consistency, an unsupervised loss strategy as a regularizer that requires zero parameters and has no time overhead. It is trained on multiple tangent scales in parallel and requires only one tangent set for inference. 

\vspace{0.15cm}\noindent\textbf{Viewport Augmentation Consistency.} 
Our model learns saliency distribution with a supervised loss but faces discrepancies in ERP saliency maps due to separate predictions for each tangent plane. To address this, we propose an unsupervised \textit{Viewport Augmentation Consistency} (VAC) loss to enhance prediction consistency across tangent projections. Specifically, we generate the second tangent set by applying different configurations, such as horizontally shifting the tangent planes on the sphere, using a larger FOV for the same viewports, and varying the number of tangent images at different viewports. We provide a detailed comparison of these approaches in the appendix. VAC does not require any additional memory or time overhead since it uses the shared parameters of the whole model, and the forward pass is done in parallel. Furthermore, since the ERP predictions from the original $P$ and augmented $P'$ tangent sets are expected to be consistent, only one tangent set is sufficient for testing. The VAC loss is defined as:
\begin{eqnarray}
\label{eqn:VACLoss}
    \mathcal{L}_{\text{VAC}}(\hat{\mathbf{y}},\ \bar{\mathbf{y}}) =& \mathcal L^{\text{weighted}}_{KLD}(\hat{\mathbf{y}},\ \bar{\mathbf{y}})\ +\ \mathcal L^{\text{weighted}}_{CC}(\hat{\mathbf{y}},\ \bar{\mathbf{y}}),\\
     \mathcal{L}^{\text{weighted}}_{\text{KLD}}(\hat{\mathbf{y}},\ \bar{\mathbf{y}}) =& \sum_{\substack{i, j}} {\hat{\mathbf{y}}_{i, j}\log{(\epsilon+\frac{\hat{\mathbf{y}}_{i, j}}{\bar{\mathbf{y}}_{i, j}+\epsilon}})}\cdot {w}_{i, j}, \\
     \mathcal{L}^{\text{weighted}}_{\text{CC}}(\hat{\mathbf{y}},\ \bar{\mathbf{y}}) =& 1 -\frac{\sum(\hat{\mathbf{y}} \cdot \bar{\mathbf{y}})\cdot w_{i,j}} {\sum(\hat{\mathbf{y}} \cdot \hat{\mathbf{y}}) \cdot\ \sum(\bar{\mathbf{y}} \cdot \bar{\mathbf{y}})}
\end{eqnarray}
where $\hat{\mathbf{y}},\ \bar{\mathbf{y}}$ are the saliency predictions from original and augmented viewports, and ${w}$ is an optional weight matrix obtained from gnomonic projection to weigh the overlapping pixels of gnomonic projection on ERP predictions. Details for viewport augmentation approaches and the weighting operation are provided in the appendix.

\subsection{Spherical Vision-Transformer Adapters for 360$^\circ$ Audio-Visual Saliency}
In our prior work, \salvit, we successfully leveraged spatio-temporal features for understanding $360^\circ$ scenes. However, \salvit~solely relied on visual stimuli, overlooking the influential role of audio cues. Motivated by the strong impact of audio in guiding human visual attention, as discussed in Section~\ref{relatedwork}, we have enhanced \salvit~with spatial audio encoding capabilities, leading to the development of \avsal. As illustrated in Fig.~\ref{fig:avsal_av_overview}, \avsal~integrates two streams for jointly processing audio and video: the video branch (top) uses the original \salvit~pipeline, while the audio branch (bottom) can flexibly incorporate any pre-trained audio backbone. It is important to note that when ambisonics are not available, the spatial audio stream in \avsal~can be conveniently substituted with mono audio.

\vspace{0.15cm}\noindent\textbf{Audio Backbone.} 
The audio stream receives spatial audio waveforms $x_{\text{aud}}\in\mathbb{R}^{4\times N}$ encoded as first-order ambisonics in 4-channel B-format containing audio streams $(W, Y, Z, X)$. To simulate what a subject would hear while gazing in a specific direction, we rotate the ambisonics for each tangent viewport, defined by angular coordinates $\{(\theta^t, \phi^t)\}_{t=1}^T$, producing mono directional waveforms $\{x_t\}_{t=1}^T$. This is achieved by applying a spherical harmonics rotation transformation to the ambisonics channels~\cite{kronlachner2014spatial, morgado2018self}. This approach enables the use of any pre-trained audio model such as PaSST~\cite{koutini22passt}, EZ-VSL~\cite{mo2022localizing} or AVSA~\cite{morgado2020learning} for feature extraction. The resulting audio features $\mathbf{x}^t_{\text{aud}}\in\mathbb{R}^{\hat{d}}$ are then aligned with their corresponding visual tokens for each viewport.

In our experiments, we evaluated all three backbone models, and empirically selected PaSST as the main audio encoder. PaSST stands out with its self-attention mechanism on mel-spectrogram patches and its unique training approach, involving 'patchout', a dropout-like mechanism. Its training leverages DeiT's pre-trained weights from ImageNet, further adapted on AudioSet for enhanced audio classification capabilities. The model processes input mel-spectrograms, extracted using a window of 25ms and a hop length of 10ms, to produce feature vectors $\mathbf{x}^t_{\text{aud}}\in\mathbb{R}^{\hat{d}}$ for audio clips. The duration for these audio clips is set to 4 seconds, based on empirical observations.

During the development of our framework, we considered several alternatives to this ambisonics decoding strategy. One baseline fed the raw B-format channels $(W, Y, Z, X)$, as either waveforms or mel-spectrograms, directly into the model. However, this configuration consistently underperformed compared to our viewport-based decoding. The core advantage of our adopted strategy is perceptual alignment: saliency in $360^\circ$ scenes often relies on spatially localized audio cues, and decoding ambisonics into view-aligned signals exposes these cues in a geometry that directly matches the tangent visual representations. This alignment enables more effective audio–visual fusion, as features share a common spatial frame of reference.

We also experimented with learning a mapping from four global channels to eighteen spatial bins through an additional trainable module such as a cross-attention mapper. While this partially addressed the alignment issue, it increased both model size and inference time, without matching the efficiency or predictive accuracy of the viewport-based strategy. As a result, we selected the current design, which ensures a direct and efficient one-to-one correspondence between audio and visual streams and allows flexible integration of state-of-the-art audio backbones. Further implementation details and a comparative analysis of backbone choices are provided in Appendix~\ref{appdx-av}.

\begin{figure}[!t]
    \centering
    \includegraphics[width=\textwidth]{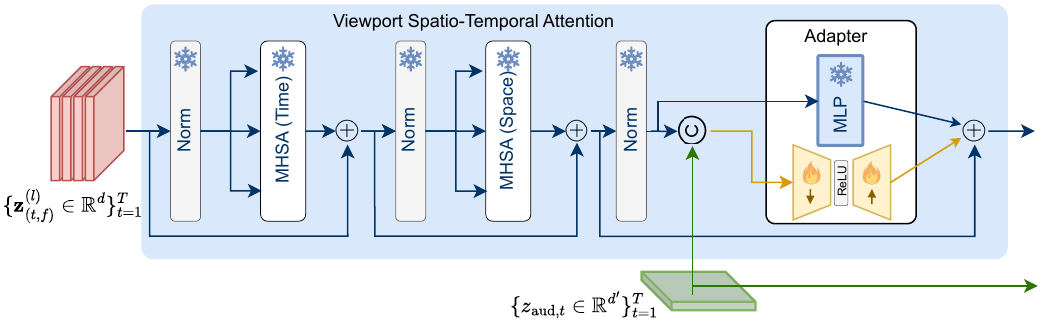}
    \caption{\textbf{Audio-Visual Adapter Fine-Tuning of \avsal.}
    For each of the $L$ transformer blocks in the frozen \salvit~pipeline, the visual tokens after Viewport Spatio-Temporal Attention are concatenated with audio tokens. The pre-trained MLPs at the end of each VSTA block are upgraded with \textbf{Audio-Visual MLP Adapters}, which consist of a scaled combination of the frozen MLP and trainable bottleneck module. $F, T, D$ denote the number of frames and tangent viewports tokens, clip length, and the embedding dimension, respectively.} 
    \label{fig:AVAdapter}
\end{figure}

\vspace{0.15cm}\noindent\textbf{Adapter Architecture.} As shown in Fig.~\ref{fig:AVAdapter}, \avsal~uses a parallel transformer adapter structure inspired by AdaptFormer~\cite{chen2022adaptformer}, with a frozen Viewport Spatio-Temporal Attention (VSTA) block from \salvit~and a lightweight bottleneck adapter for audio-visual fusion. The model processes audio tokens $\{{z}_{\text{aud}}\in\mathbb{R}^{\hat{d}}\}_{t=1}^T$ alongside visual tokens $\{\mathbf{z}^{\textit{(l)}}_{\text{(t, f)}}\in\mathbb{R}^{d}\}_{(t=1, f=1)}^{(T, F)}$. This dual-stream setup involves a stack of down-projection and up-projection layers, interconnected with ReLU non-linearity, functioning as a bottleneck with significantly lower dimensions than the original embeddings. Specifically, this stream is designed as a bottleneck module with a dimension of $k$, as the projections are performed by parameters $\mathbf{W}_{down}\in\mathbb{R}^{{(d+\hat{d})}\times k}$, and $\mathbf{W}_{up}\in\mathbb{R}^{k\times \hat{d}}$, where $k\ll d, \hat{d}$. The concatenated audio-visual embeddings $\{\bar{z}_{\text{av, t}}\in \mathbb{R}^{d+\hat{d}}\}_{t=1}^T$ for each tangent viewport undergo a transformation to generate audio-conditioning features $\hat{z}_{\text{av, t}}$ as:
\begin{equation}
    \hat{z}_{\text{av, t}} = \text{ReLU}(\text{LN}(\bar{z}_{\text{av, t}})\cdot\mathbf{W}_{down})\cdot\mathbf{W}_{up}\;\;,
\end{equation}
which are then fused with the visual features $\bar{z}_{\text{av, t}}$ to enhance the saliency prediction capability of the model:
\begin{equation}
    \mathbf{z}_{\text{av, t}} = \text{MLP}(\bar{z}_{\text{av, t}}) + {s} \cdot \hat{z}_{\text{av, t}} + \bar{z}_{\text{av, t}}
\end{equation}
Our adapter-based design ensures that when no audio input is provided such as in purely visual 360$^\circ$ saliency datasets, the audio branch transmits no signal to the adapter layers. In this case, SalViT360-AV behaves identically to SalViT360, and the model effectively reduces to single-modal visual saliency prediction without any loss of performance or the need for re-training. This flexibility allows our framework to generalize seamlessly across both single-modal (visual-only) and multi-modal (audio-visual) saliency prediction tasks.

\section{Experiments}
Our experimental evaluation was structured in a two-fold approach. In the first part, focusing on vision-only settings, we trained the \salvit~model using the VR-EyeTracking dataset~\cite{xu2018gaze}. This dataset comprises 134 videos for training and 74 videos for testing. For validation, we randomly selected a subset of 30 videos from the training set and assessed the performance of \salvit~on the test set. Additionally, to evaluate the model's generalization ability, we conducted cross-dataset evaluations using the PVS-HMEM~\cite{xu2018predicting}, 360AV-HM~\cite{chao2020audio} datasets, and the test split of our \bounvr~dataset. These datasets include 76, 21, and 27 videos, respectively, and each were viewed by 58, 15, and 15 subjects. 
In the second part of our experimentation, we focused on audio-visual tuning of the \avsal~model. We began by utilizing the pre-trained weights of \salvit~from the VR-EyeTracking dataset, which we kept frozen throughout this phase. The training of \avsal~was conducted using the training split of the \bounvr~dataset. We then evaluated \avsal’s performance on various datasets, including the test set of VR-EyeTracking, the \bounvr~test split, and the 360AV-HM dataset. 

\vspace{0.15cm}\noindent\textbf{Data Pre-processing.} The audio waveforms are resampled to a frequency of 32.0 kHz to align with the PaSST backbone. The duration of audio clips is empirically chosen to be a maximum of 4 seconds. This choice is based on discussions in \cite{cokelek2021leveraging}, which suggest that shorter clips tend to capture short-term audio saliency, while longer clips may encompass more than local saliency. 
A detailed comparative analysis of clip durations can be found in the Appendix.

After rotating the 4-channel ambisonics for each viewport, we decode them to extract a single-channel directional waveform $F$ for mono audio backbones (like PaSST and EZ-VSL), using the formula:
\begin{equation}
    {F} = (\sqrt{2}W + X)\cdot2.
\end{equation}\label{eqn:foa-decoding}

\vspace{-0.35cm}
\noindent\textbf{Evaluation Metrics and Loss Functions.} We evaluate the performance of the models using four commonly used metrics, namely, Normalized Scanpath Saliency (NSS), KL-Divergence (KLD), Pearson's Correlation Coefficient (CC), and Similarity Metric (SIM). We compute the training objective as a weighted differentiable combination of KLD, CC, and Selective-MSE (MSE on normalized saliency maps at only eye-fixation points~\cite{salfbnet}) as given below:
\begin{eqnarray}\label{loss}
\mathcal{L}(P, {Q}_{s},{Q}_{f}) & = & \text{KLD}(P,{Q}_s)+ \text{CC}(P, Q_s)\ \\\nonumber & + & \alpha\,\text{SMSE}(P, {Q}_s, {Q}_f)
\end{eqnarray}
where $P,\ {Q}_s,\ {Q}_f$ are the predicted saliency, ground truth density and fixation maps, respectively, and $\alpha = 0.05$.

\vspace{0.15cm}\noindent\textbf{Architecture and Optimization Details.}
We trained \salvit~for 100K iterations using AdamW~\cite{loshchilov2017decoupled} with momentum parameters $(\beta_1,\ \beta_2) = (0.9,\ 0.999)$, a weight decay of $1e-2$ and an initial learning rate of $1e-5$. A cosine scheduler was applied, decaying the lr to $2e-6$ over the course of training. The batch size was set to $16$, and all experiments were conducted on a single Tesla V100 GPU with 32 GB of memory. Subsequently, we performed parameter-efficient fine-tuning of \avsal~for an additional 20,000 steps using the same optimizer with a fixed learning rate of 2e-6.

\subsection{Empirical Evaluation Across Visual and Audio-Visual Saliency Benchmarks}
We conduct an extensive evaluation of our proposed \salvit~and \avsal~models, systematically examining their performance across a diverse set of benchmarks. Our analysis is structured into three main parts: (1) evaluation on datasets without spatial audio, which primarily assess visual saliency models; (2) assessment on our proposed \bounvr~dataset to evaluate the benefits of integrating spatial audio; and (3) cross-dataset generalization on unseen spatial audio-visual datasets.

\vspace{0.15cm}\noindent\textbf{Evaluation on Datasets without Spatial Audio.} We first evaluate the effectiveness of our visual transformer model \salvit~using the widely adopted VR-EyeTracking dataset~\cite{xu2018gaze}, which, despite containing mono audio, lacks spatial audio annotations and thus serves primarily as a visual-only benchmark. VR-EyeTracking provides an extensive test scenario, comprising 134 training and 74 testing videos. To comprehensively assess the model’s generalization beyond training data, we further evaluate its performance on the PVS-HMEM dataset~\cite{xu2018predicting}, which features purely visual annotations without any audio content.
Quantitative results provided in Table~\ref{table:vreye_pvshmem} clearly demonstrate that \salvit~achieves top-tier performance, frequently surpassing existing state-of-the-art visual-only saliency models. In particular, our transformer-based architecture shows substantial advantages in capturing complex visual contexts and accurately predicting viewer fixation patterns.

\begin{figure}[!t]
    \centering
    \includegraphics[trim={130 40 125 40},clip,width=0.99\textwidth]{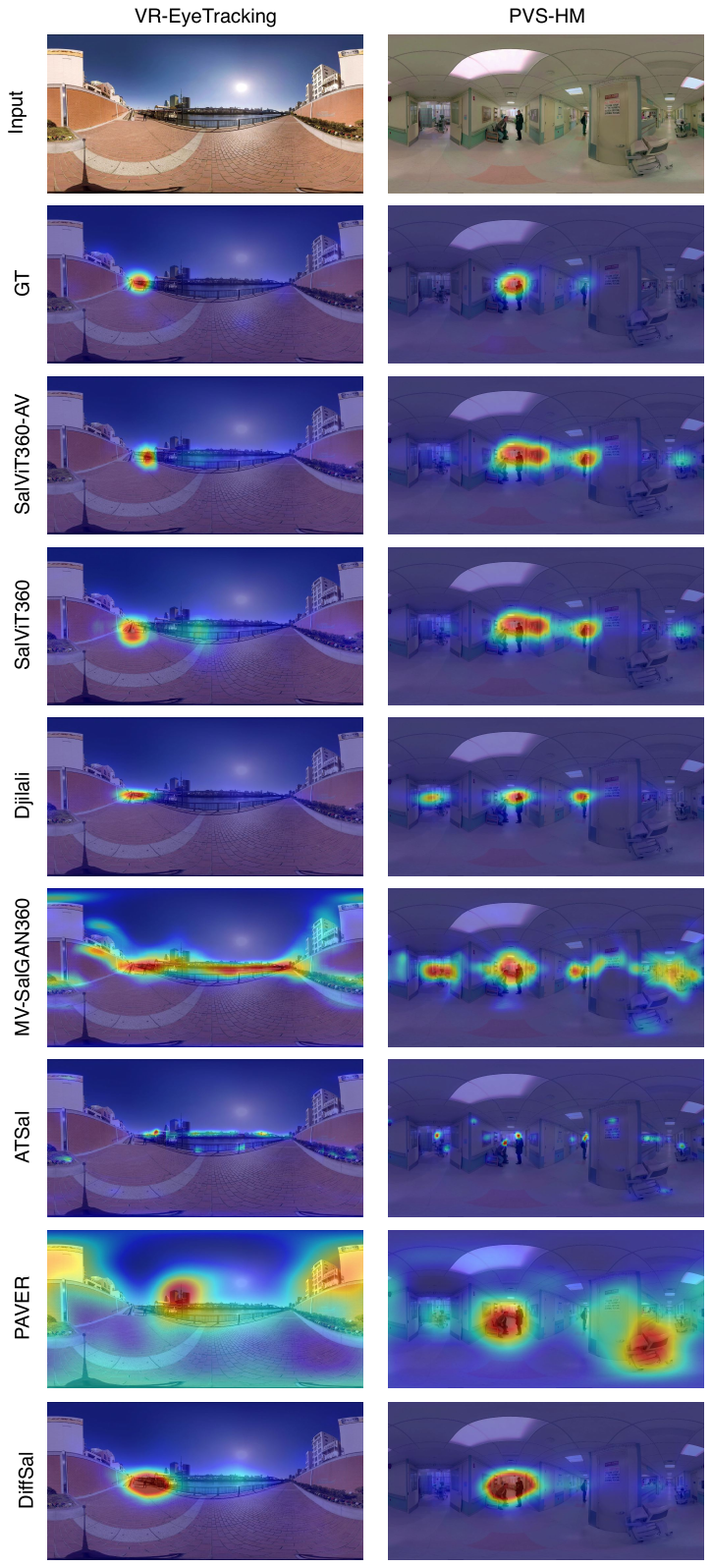}
    \caption{\textbf{Qualitative Comparison of Predicted Saliency Maps on Sample Frames from VR-EyeTracking and PVS-HMEM.} Both SalViT360 and SalViT360-AV accurately capture human gaze patterns, effectively highlighting salient cues in complex panoramic scenes. Competing visual-only models and the 2D audio-visual DiffSal model often miss or oversmooth key regions.}
    \label{fig:qualitative-VREye-PVS}
\end{figure}

Regarding the audio-visual evaluation on these datasets, we note that 360$^\circ$ audio-visual saliency models, apart from our SalViT360-AV, could not be directly evaluated on the PVS-HMEM dataset, as it entirely lacks audio data. Moreover, they were incompatible with the VR-EyeTracking dataset, as they require spatial audio to process. On the other hand, our SalViT360-AV model can effectively process mono audio waveforms or even no audio (see our discussion at the end of Sec. 4.2); thus, we can provide results on both datasets. As expected, in the presence mono audio, SalViT360-AV demonstrate improved predictive accuracy compared to existing 2D audio-visual models on the VR-EyeTracking dataset. As shown in Fig.~\ref{fig:qualitative-VREye-PVS}, SalViT360 and SalViT360-AV generate saliency maps that accurately localize human gaze across the provided panoramic scenes. Our models reliably focus on the most visually salient regions, even in cases where attention is narrowly concentrated. In contrast, previous visual-only 360$^\circ$ models as well as 2D audio-visual models often produce saliency maps that are either overly blurred or misplace attention, highlighting irrelevant parts of the scene. Our models are less prone to center or equator bias and more reliably highlight peripheral yet visually relevant regions, yielding predictions that are much closer to the ground truth fixations.

\begin{table}[!t]
    \centering
    \caption{\textbf{Quantitative Evaluation on VR-EyeTracking and PVS-HMEM.} Although the VR-EyeTracking dataset contains mono audio, it lacks spatial audio cues and is thus treated as a visual saliency benchmark. The PVS-HMEM dataset does not include sound. The models are grouped by category using background color: blue for 360$^\circ$ visual-only, orange for 2D audio-visual, and green for 360$^\circ$ audio-visual saliency models. \textbf{Bold} scores indicate the best performance, while \underline{underlined} scores denote the second best.}
    \resizebox{\linewidth}{!}{
    \begin{tabular}{l@{$\;\;$}c@{$\;\;$}c@{$\;\;$}c@{$\;\;$}c@{$\;\;\;$}c@{$\;\;$}c@{$\;\;$}c@{$\;\;$}c}
        \toprule
        & \multicolumn{4}{c}{{\bf VR-EyeTracking}} & \multicolumn{4}{c}{{\bf PVS-HMEM}} \\
        \cmidrule(lr){2-5} \cmidrule(lr){6-9}
        {\bf Method} & {\bf NSS$\uparrow$} & {\bf KLD$\downarrow$} & {\bf CC$\uparrow$} & {\bf SIM$\uparrow$} & {\bf NSS$\uparrow$} & {\bf KLD$\downarrow$} & {\bf CC$\uparrow$} & {\bf SIM$\uparrow$} \\
        \midrule
        \rowcolor{niceblue} CP-360 & 0.624 & 15.338 & 0.165 & 0.240 & 0.576 & 4.738 & 0.162 & 0.198 \\
        \rowcolor{niceblue} ATSal & 1.317 & 12.259 & 0.336 & 0.318 & 0.732 & 4.303 & 0.183 & 0.219 \\
        \rowcolor{niceblue} SalGAN360 & 1.753 & 10.845 & 0.370 & 0.355 & 1.513 & 4.394 & 0.314 & 0.291 \\
        \rowcolor{niceblue} MV-SalGAN360 & 1.818 & 8.713  & 0.382 & 0.357 & 1.546 & 4.112 & 0.316 & 0.295 \\
       \rowcolor{niceblue}  PAVER & 1.511 & 13.267 & 0.307 & 0.294 & 0.750 & 3.736 & 0.224 & 0.269 \\
       \rowcolor{niceblue}  Djilali~et~al. & \textbf{3.183} & \underline{6.570} & \underline{0.565} & \underline{0.475} & \underline{1.688} & \underline{2.430} & \underline{0.447} & \underline{0.404} \\
       \rowcolor{niceblue}  \textbf{\salvit}                          & \underline{2.630} & \textbf{5.744} & \textbf{0.586} & \textbf{0.492} & \textbf{2.191} & \textbf{1.841} & \textbf{0.626} & \textbf{0.495} \\
       \midrule
        \rowcolor{niceorange} DAVE    & 1.821 & 12.199 & 0.331 & 0.304 & 1.644 & 4.252 & 0.309 & 0.273 \\
        \rowcolor{niceorange} STAViS  & 1.642 & 13.148 & 0.316 & 0.297 & 1.539 & 4.320 & 0.306 & 0.266 \\
        \rowcolor{niceorange} DiffSal & 2.599 & 5.979 & 0.566 & 0.473 & \underline{1.926} & 2.539 & 0.443 & 0.399 \\
      \rowcolor{niceblue2} 
      \midrule
       
       \rowcolor{nicegreen}  \textbf{\avsal}                   & \textbf{2.821} & \textbf{5.334} & \textbf{0.599} & \textbf{0.511} & \textbf{2.191} & \textbf{1.841} & \textbf{0.626} & \textbf{0.495}\\
        \bottomrule
    \end{tabular}
    }
    \label{table:vreye_pvshmem}
\end{table}

\begin{table}[!t]
    \centering
    \caption{\textbf{Evaluation on \bounvr~to Assess the Impact of Spatial Audio Integration.} We compare visual-only, 2D audio-visual, 360$^\circ$ audio-only, and 360$^\circ$ audio-visual saliency models. Categories are color-coded: blue for 360$^\circ$ visual, orange for 2D audio-visual, green for 360$^\circ$ audio-visual, and gray for the 360$^\circ$ audio-only models.} 
    
    \resizebox{0.8\textwidth}{!}{
    \begin{tabular}{lcccc}
        \toprule
        \textbf{Method} & \textbf{NSS$\uparrow$} & \textbf{KLD$\downarrow$} & \textbf{CC$\uparrow$} & \textbf{SIM$\uparrow$} \\
        \midrule
        \rowcolor{niceblue} CP-360  & 1.041 & 22.356 & 0.120 & 0.118 \\
        \rowcolor{niceblue} ATSal & 1.711 & 13.948 & 0.242 & 0.244 \\
        \rowcolor{niceblue} SalGAN360 & 1.415 & 14.412 & 0.236 & 0.239 \\
        \rowcolor{niceblue} MV-SalGAN360 & 1.433 & 14.381 & 0.244 & 0.246 \\
        \rowcolor{niceblue} PAVER & 1.093 & 14.830 & 0.254 & 0.226 \\
        \rowcolor{niceblue} Djilali~et~al. & \underline{1.738} & \underline{11.085} & \underline{0.369} & \underline{0.326} \\
        \rowcolor{niceblue} \textbf{\salvit}                        & \textbf{2.346} & \textbf{9.861} & \textbf{0.484} & \textbf{0.373} \\
        \midrule
        \rowcolor{niceorange} DAVE & 2.290 & 8.858 & 0.462 & 0.363 \\
        \rowcolor{niceorange} STAViS & 1.535 & 12.039 & 0.343 & 0.294 \\
        \rowcolor{niceorange} DiffSal                           & \underline{2.293} & 8.458 & \underline{0.477} & \underline{0.383} \\
        \midrule
        \rowcolor{niceblue2} SSSL & 1.212 & 13.398 & 0.214 & 0.244 \\
        \midrule
        \rowcolor{nicegreen} AVS360 & 2.170 & \textbf{7.801} & 0.457 & 0.374 \\
        \rowcolor{nicegreen} SVGC-AVA                        & 2.191 & 8.860 & 0.488 & 0.399 \\
        \rowcolor{nicegreen}\textbf{SalViT360-AV~(w/ raw FOA)}   & \textbf{2.443} & \underline{8.539} & \textbf{0.508} & \textbf{0.406} \\
        \rowcolor{nicegreen}\textbf{\avsal~(proposed)}       & \textbf{2.449} & \underline{8.341} & \textbf{0.512} & \textbf{0.407} \\
        \bottomrule
    \end{tabular}}
    \label{table:yt360}
\end{table}

\begin{figure*}[!t]
    \centering
 \includegraphics[width=\textwidth]{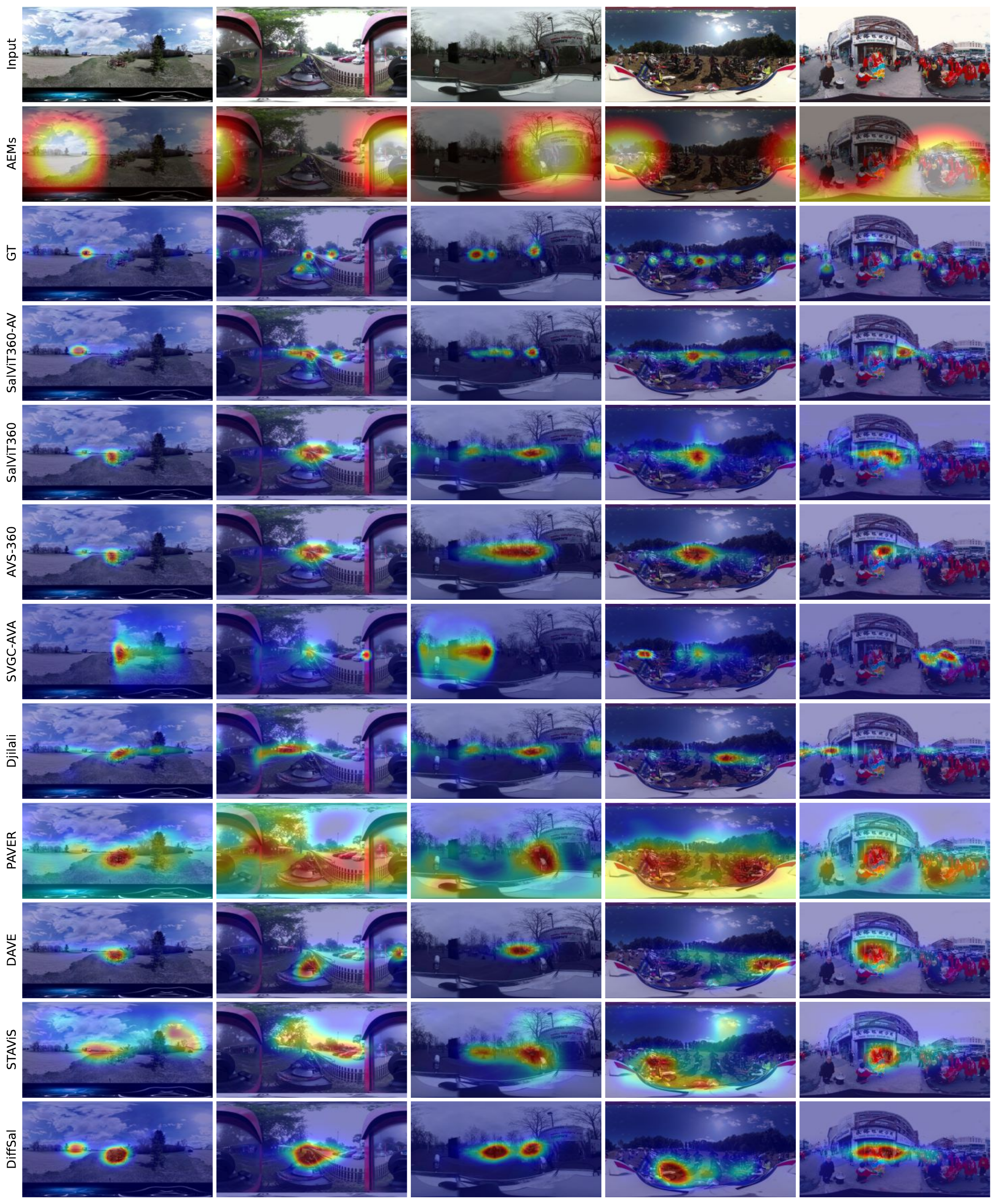}
    \caption{\textbf{Predicted Saliency Maps Illustrating the Integration of Spatial Audio Cues in the YT360-EyeTracking Dataset.} SalViT360-AV clearly leverages ambisonic sound to pinpoint visually salient regions strongly influenced by audio events (e.g., approaching vehicles, off-screen conversations, and environmental sounds). Compared to visual-only or 2D audio-visual baselines, our model produces richer and contextually accurate attention maps aligned closely with human~fixations.}
    \label{fig:qualitative-YT360}
\end{figure*}

\begin{table*}[!htb]
    \centering
    \caption{\textbf{Cross-Dataset Generalization Results on 360AV-HM, AVS-ODV, and SVGC-AVA.} All~datasets include ambisonic spatial audio. The models are grouped by category using background color: blue for 360$^\circ$ visual saliency, orange for 2D audio-visual saliency, gray for the 360$^\circ$ audio-only saliency, green for 360$^\circ$ audio-visual saliency prediction.}
   \resizebox{0.9\linewidth}{!}{
    \begin{tabular}%
    {lcccccccccccccc}
        \toprule
        & \multicolumn{4}{c}{{\bf 360AV-HM}} & & \multicolumn{4}{c}{{\bf AVS-ODV}} & & \multicolumn{4}{c}{{\bf SVGC-AVA}}\\
        \cmidrule(lr){2-5} \cmidrule(lr){7-10} \cmidrule(lr){12-15}
        {\bf Method} & {\bf NSS$\uparrow$} & {\bf KLD$\downarrow$} & {\bf CC$\uparrow$} & {\bf SIM$\uparrow$} & & {\bf NSS$\uparrow$} & {\bf KLD$\downarrow$} & {\bf CC$\uparrow$} & {\bf SIM$\uparrow$} & & {\bf NSS$\uparrow$} & {\bf KLD$\downarrow$} & {\bf CC$\uparrow$} & {\bf SIM$\uparrow$} \\
        \midrule        
        \rowcolor{niceblue} ATSal & 1.322 & \textbf{14.141} & 0.156 & 0.121 & & 1.430 & 13.159 & 0.233 & 0.229 & & 1.510 & 11.886 & 0.314 & 0.311 \\
        \rowcolor{niceblue} PAVER & 1.019 & 19.136 & 0.176 & 0.115 & & 1.619 & 13.252 & 0.251 & 0.241 & & 1.641 & 11.302 & 0.422 & 0.396 \\
        \rowcolor{niceblue} Djilali~et~al. & \underline{2.050} & 16.618 & \underline{0.308} & \underline{0.209} & & 2.014 & 10.167 & 0.344 & 0.300 & & 2.020 & 10.239 & 0.511 & 0.419 \\ 
        \rowcolor{niceblue} \textbf{\salvit}  & \textbf{2.285} & 15.879 & \textbf{0.349} & \textbf{0.246} & & 2.192 & \textbf{8.626} & \textbf{0.390} & \textbf{0.321} & & 2.179 & \textbf{9.320} & \textbf{0.557} & \textbf{0.434} \\
        \midrule 
        \rowcolor{niceorange} DAVE & 2.178 & 14.693 & 0.344 & 0.256 & & 2.050 & 12.065 & 0.341 & 0.279 & & 1.997 & 10.451 & 0.446 & 0.401 \\
        \rowcolor{niceorange} STAViS & 1.479 & 17.934 & 0.237 & 0.162 & & 2.012 & 12.195 & 0.340 & 0.277 & & 1.991 & 10.543 & 0.443 & 0.400 \\
        \rowcolor{niceorange} DiffSal & 2.460 & 13.991 & \underline{0.376} & 0.271 & & 2.368 & 11.866 & \textbf{0.398} & 0.317 & & 2.276 & \textbf{8.274} & 0.541 & 0.408 \\
        \midrule
        \rowcolor{niceblue2} SSSL & 1.181 & 14.192 & 0.203 & 0.184 & & 1.019 & 13.225 & 0.198 & 0.184 & & 1.190 & 11.469 & 0.211 & 0.187 \\
        \midrule
        \rowcolor{nicegreen} AVS360 & \textbf{2.501} & \textbf{13.581} & \textbf{0.380} & \textbf{0.290} & & 2.386 & \textbf{8.130} & {0.394} & \underline{0.332} & &2.320 & {9.245} & 0.559 & 0.433 \\
        \rowcolor{nicegreen} SVGC-AVA & 2.448 & 14.012 & 0.367 & 0.266 & & \underline{2.387} & \underline{8.379} & {0.394} & \textbf{0.334} & & \textbf{2.371} & \underline{9.241} & \underline{0.560} & \underline{0.434} \\
    \rowcolor{nicegreen} \textbf{\avsal} & \underline{2.473} & \underline{13.830} & \underline{0.379} & \underline{0.278} & & \textbf{2.389} & 8.481 & \textbf{0.397} & \underline{0.332} & & \underline{2.342} & 9.251 & \textbf{0.564} & \textbf{0.440} \\
        \bottomrule
    \end{tabular}
    }
    \label{table:spaudio_cross-results}
\end{table*}

\vspace{0.15cm}\noindent\textbf{Effectiveness of Integrating Spatial Audio.} Next, we systematically investigate the benefit of spatial audio integration. Leveraging our newly introduced \bounvr~dataset, which explicitly captures viewer attention under spatial audio conditions, we fine-tune our audio-visual \avsal~model, initialized with weights from \salvit~pretrained on VR-EyeTracking. Our transformer-based audio adapters effectively incorporate ambisonic spatial audio cues into the visual backbone.
Quantitative outcomes detailed in Table~\ref{table:yt360} highlight that \avsal~consistently outperforms visual-only and state-of-the-art 2D audio-visual models across all metrics. Notably, the results emphasize the significant advantage of our audio-adaptive strategy, clearly demonstrating its ability to harness spatial audio information effectively. 
As shown in Fig.~\ref{fig:qualitative-YT360}, SalViT360-AV stands out in aligning attention with salient audio-driven events such as off-screen conversations, approaching vehicles, or spatialized environmental sounds. Compared to both visual-only and other 2D/360$^\circ$ audio-visual baselines, SalViT360-AV’s maps are more precise, especially in challenging situations where salient cues are strongly dictated by audio. Competing methods tend to produce fragmented or overly diffused predictions under such conditions, failing to adapt their attention when the salient event is not visually dominant but audibly salient. In addition, we evaluate a model variant that directly uses raw FOA (4-channel B-format) signals without applying our proposed ambisonics rotation and decoding. As shown in Table~\ref{table:yt360}, this baseline (SalViT360-AV w/ raw FOA) delivers competitive results; however, \avsal~consistently outperforms it across all metrics. We attribute this improvement to two key factors: (1) while FOA channels offer a compact representation, they do not explicitly encode directional information, whereas decoding into 18 directional waveforms more closely replicates how spatial cues are perceived in VR environments; and (2) using only four channels in the cross-attention integration reduces the size of the attention matrix, leading to coarser audio-visual correspondences. In contrast, our decoded directional signals support finer, view-specific alignments. These findings confirm the importance of spatially aligning audio inputs prior to fusion for more accurate and robust saliency prediction.

\begin{figure*}[!t]
    \centering
    \includegraphics[height=0.75\textwidth]{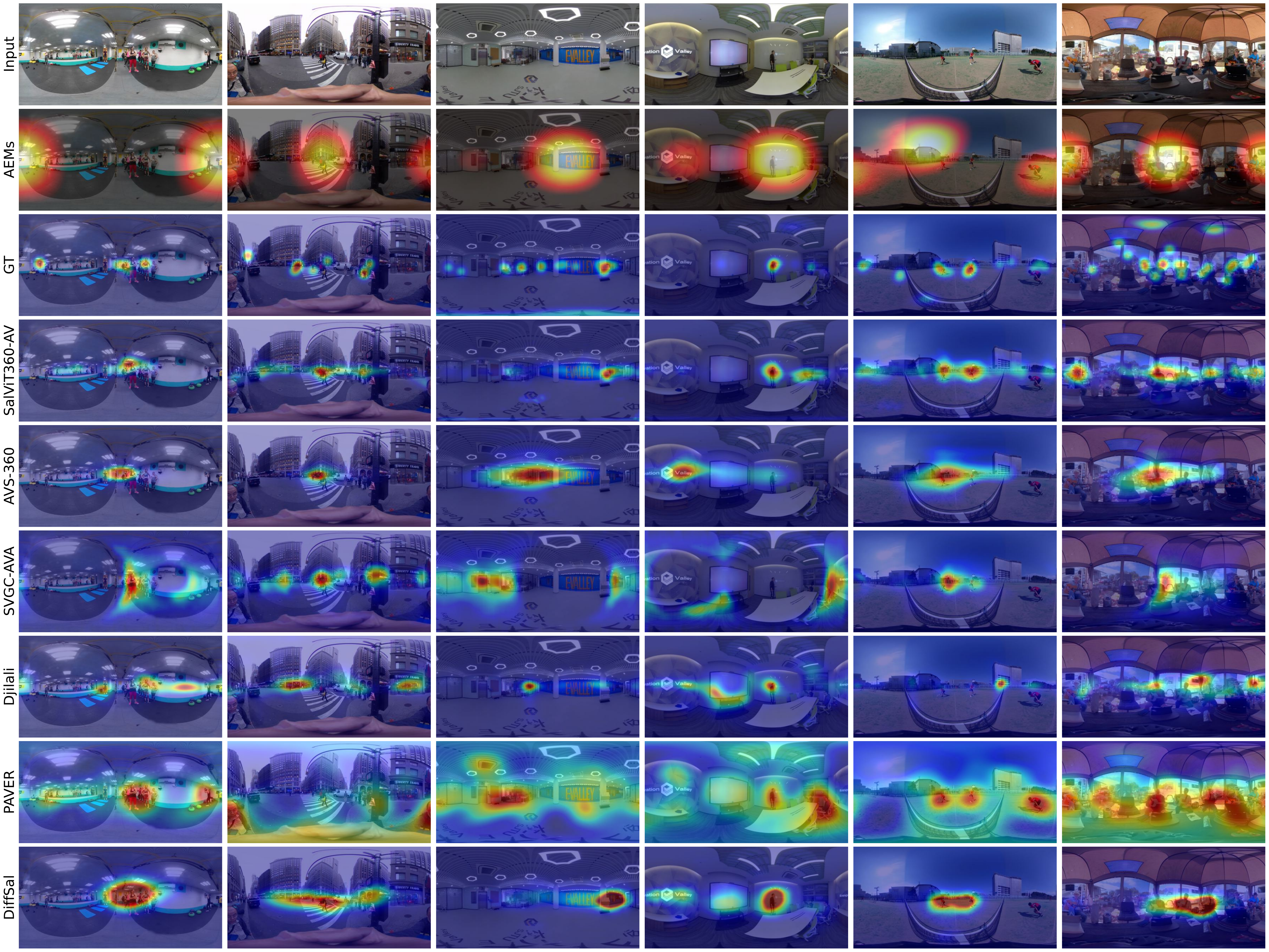}
    \caption{\textbf{Cross-Dataset Generalization Results on 360AV-HM, AVS-ODV, and SVGC-AVA Datasets.} Qualitative saliency predictions on sample frames from  360AV-HM (columns 1–2), AVS-ODV (columns 3–4), and SVGC-AVA (columns 5–6) datasets. These datasets span diverse indoor and outdoor scenes with varying spatial audio complexity. Our model consistently localizes salient regions influenced by spatial audio cues, outperforming recent state-of-the-art models.}
    \label{fig:qualitative-spaudio-cross}
\end{figure*}

\vspace{0.15cm}\noindent\textbf{Cross-Dataset Generalization under Spatial Audio.} Finally, we examine the robustness and generalizability of our approach by evaluating the fine-tuned \avsal~model on three additional datasets containing spatial audio cues: 360AV-HM\cite{chao2020audio}, AVS-ODV~\cite{11003418}, and SVGC-AVA~\cite{yang2023svgc}. These datasets vary substantially in content complexity, audio-visual dynamics, and participant viewing conditions, thus providing rigorous tests of generalization. As summarized in Table~\ref{table:spaudio_cross-results}, our model consistently achieves top-level performance, outperforming or closely matching recent state-of-the-art 360$^\circ$ audio-visual methods.\footnote{We note that the AVS360 model is trained on 360AV-HM dataset, and, thus, obtains the highest scores on the test set of 360AV-HM.} These results confirm the broad adaptability of \avsal~to unseen datasets characterized by complex spatial audio environments. The qualitative results in Fig.~\ref{fig:qualitative-spaudio-cross} further highlight the robustness and adaptability of SalViT360-AV. Even when tested on datasets it was not trained on, our model consistently identifies salient audio-visual regions. For example, in the AVS-360 scene where a vehicle passes behind the camera, SalViT360-AV accurately shifts attention to the rear-facing tangent viewport, unlike competing models that either miss the cue or overemphasize the front-facing view. Similarly, in the SVGC-AVA clip with multiple sound sources, our model selectively highlights the dominant auditory event while suppressing distractions. These examples demonstrate that SalViT360-AV effectively understands subtle interactions between auditory and visual modalities. Unlike earlier methods, which tend to blur attention or focus on visually prominent but semantically irrelevant areas, our model produces sharper, context-aware predictions that align with human-like attention shifts across diverse content types.

\begin{table}[!t]
    \centering
    \caption{\textbf{Ablation Study of SalViT360 Components on  VR-EyeTracking.} We evaluate the impact of core architectural components. Replacing 1D positional embeddings with spherical ones improves performance. Introducing our VSTA module yields further gains, and the VAC module—with masking—achieves the best results, demonstrating the benefit of spatial priors and temporal context.} 
    \resizebox{\textwidth}{!}{
    \begin{tabular}{@{}l@{$\;\;$}c@{$\;\;$}c@{$\;\;$}c@{$\;\;$}c@{$\;\;$}c@{$\;\;$}c@{}}
        \toprule
        {\bf Method} & {\bf \#\ params} & {\bf NSS$\uparrow$} & {\bf KLD$\downarrow$} & {\bf CC$\uparrow$} & {\bf SIM$\uparrow$}  \\
        \midrule
        VSA (w/\ {1D\ Pos.\ Emb.})                    & 18.76M & 2.518 & 6.445 & 0.560 & 0.472 \\
        {\ \ \ \ }+\ Spherical\ Pos.\ Emb.            & 19.26M & 2.575 & 6.221 & 0.563 & 0.475 \\
        VSTA (w/\ {Sph. \ Pos.\ Emb.})                   & 25.56M & \textbf{2.664} & 6.174 & 0.570 & 0.479 \\
        {\ \ \ \ } \ +\ VAC\ (w/o\ mask)              & 25.56M & {2.624} & {6.011} & {0.576} & {0.490} \\
        {\ \ \ \ } \ +\ VAC\ (w/\ mask)                & 25.56M & \underline{2.630} & \underline{5.744} & \underline{0.586} & \underline{0.492} \\
        \bottomrule
    \end{tabular}
    }
    \label{table:ablation}
\end{table}

\subsection{Ablation Studies}
To provide a more comprehensive analysis of each component in our audio-visual saliency pipeline, we perform experiments on the validation splits of VR-EyeTracking and \bounvr~datasets, for our \salvit~and \avsal models, respectively.

\vspace{0.15cm}\noindent\textbf{Spherical Position Embeddings.} We compare the performance of our proposed \textit{spherical geometry-aware spatial position embeddings} with regular 1D learnable position embeddings in Table~\ref{table:ablation}. The results on all four metrics show that our proposed embedding method outperforms it, demonstrating that it is more suitable for processing spherical data with Vision Transformers.

\begin{table}[!t]
    \centering
    \caption{\textbf{Comparison of spatio-temporal modeling strategies.} We compare our VSTA mechanism against a 2+1D CNN and offline EMA aggregation. VSTA outperforms both alternatives across all metrics, confirming the advantage of end-to-end learnable spatio-temporal attention in 360$^\circ$ video saliency prediction.} 
    \resizebox{\textwidth}{!}{
    \begin{tabular}{@{}l@{$\;\;$}c@{$\;\;$}c@{$\;\;$}c@{$\;\;$}c@{$\;\;$}c@{$\;\;$}c@{}}
        \toprule
        {\bf Method} & {\bf \#\ params} & {\bf NSS$\uparrow$} & {\bf KLD$\downarrow$} & {\bf CC$\uparrow$} & {\bf SIM$\uparrow$}  \\
        \midrule
        VSA $\ +\ $ {2+1D-CNN Enc.}                    & 17.31M & 2.568 & \textbf{5.915} & 0.568 & 0.477 \\
        VSA $\ +\ $ {Offline EMA}                         & 19.27M & 2.591 & 6.018 & 0.566 & 0.477 \\
        VSTA                                  & 25.56M & \textbf{2.664} & {6.174} & \textbf{0.570} & \textbf{0.479} \\
        \bottomrule
    \end{tabular}
    }
    \label{table:stmodelling}
\end{table}

\vspace{0.15cm}\noindent\textbf{Spatio-Temporal modelling.} 
In Table~\ref{table:stmodelling}, we compare Viewport Spatial Attention (VSA) and Viewport Spatio-Temporal Attention (VSTA) blocks to assess the contribution of temporal information processing in omnidirectional videos. We conducted several experiments to assess the effectiveness of our proposed VSTA mechanism. We include two distinct approaches, {2+1D-CNN}~\cite{morgado2020learning} backbone and {Offline EMA} in this analysis. {2+1D-CNN} backbone is an R2+1D model~\cite{8578773} which performs convolution over consecutive frames, pre-trained on undistorted {normal-FOV} crops in $360^\circ$ videos. We replace ResNet-18 + VSTA with {2+1D-CNN}+VSA to introduce temporal features for spatial self-attention. 

In the other setting, we keep ResNet-18 and VSA and apply a weighted exponential moving average on $F$ consecutive predictions for temporal aggregation. We also experiment on transformer depth, which shows that our VSTA blocks gradually learn better spatio-temporal representations in deeper layers. Our results show that the proposed VSTA mechanism outperforms the spatial-only setting and the other two spatio-temporal approaches. We refer the reader to the appendix for comprehensive experiments on the transformer depth, and temporal window size $F$, along with a comparison of joint spatio-temporal attention.

\vspace{0.15cm}\noindent\textbf{Impact of Ground-Truth Audio Modality} The saliency datasets having spatial audio are collected under (1) muted, (2) mono, and (3) ambisonics audio conditions, each audio modality viewed by a different subject group to prevent biases. We evaluate our video-only baseline on each ground truth separately. The results in Table~\ref{avsal_inter_subject_correlation} demonstrate that the increasing representative power of audio modality yields higher inter-subject correlation on our video-only model.

\begin{table}[!t]
    \centering
    \caption{\textbf{Effect of audio modality in ground truth saliency.} We evaluate SalViT360 using saliency maps collected under mute, mono, and ambisonics audio. Performance improves with richer audio conditions, highlighting how spatial audio influences human attention—even when the model is trained without audio.}
    \resizebox{\linewidth}{!}{
    \begin{tabular}{@{}l@{$\;\;$}c@{$\;\;$}c@{$\;\;$}c@{$\;\;$}c@{$\;\;\;$}c@{$\;\;$}c@{$\;\;$}c@{$\;\;$}c@{}}
        \toprule
        & \multicolumn{4}{c}{{\bf 360AV-HM}} & \multicolumn{4}{c}{{\bf \bounvr}} \\
        \cmidrule(lr){2-5} \cmidrule(lr){6-9}
        {\bf Audio} & {\bf NSS$\uparrow$} & {\bf KLD$\downarrow$} & {\bf CC$\uparrow$} & {\bf SIM$\uparrow$} & {\bf NSS$\uparrow$} & {\bf KLD$\downarrow$} & {\bf CC$\uparrow$} & {\bf SIM$\uparrow$} \\
        \midrule
        {Mute}        & 1.961 & 16.819 & 0.289 & 0.225 & 2.115 & \textbf{9.798} & 0.461 & 0.370\\
        {Mono}        & 2.230 & 16.356 & 0.329 & 0.238 & 2.315 & 9.993 & 0.478 & 0.371 \\
        {Ambisonics}  & \textbf{2.285} & \textbf{15.879} & \textbf{0.349} & \textbf{0.246} & \textbf{2.346} & 9.861 & \textbf{0.484} & \textbf{0.373} \\

        \bottomrule
    \end{tabular}
    }
    \label{avsal_inter_subject_correlation}
\end{table}

\section{Conclusion}
In this study, we introduced \salvit and \avsal. two models for predicting saliency in $360^\circ$ video and audio-visual contexts. \salvit, leverages an encoder-transformer-decoder architecture that processes undistorted tangent image representations using a spatio-temporal attention mechanism and spherical geometry-aware position embeddings. Evaluations on four benchmark datasets, supported by ablation studies, confirm consistent improvements over state-of-the-art methods.

To model the auditory dimension of attention, SalViT360-AV extends SalViT360 with a parameter-efficient adapter fine-tuning strategy, frozen pre-trained backbones, and a spatial audio pipeline that decodes and rotates ambisonics per viewport. This integration significantly boosts performance on datasets with audio–visual content.

Beyond saliency prediction, both models have broader applications, including improving audio-visual quality assessment~\cite{Cao2023AudioVisualQA,Zhu2023OAVQA,Min2020AVQA}, automatically generating viewing paths for immersive navigation~\cite{Cha2020Enhanced360}, and guiding compression to prioritize perceptually important regions~\cite{Chiang2021SaliencyRDO}.

\section*{Acknowledgment}
This work was supported in part by the KUIS AI Center Research Award, Unvest R\&D Center, TÜBİTAK-1001 Program (No. 120E501), the TÜBA-GEBİP 2018 Award to E. Erdem, and the BAGEP 2021 Award to A. Erdem.

\ifCLASSOPTIONcaptionsoff
  \newpage
\fi

{\small
\bibliographystyle{unsrt}
\bibliography{references}
}

\vskip -2\baselineskip plus -1fil
\begin{IEEEbiography}
[{\includegraphics[width=1in,height=1.25in,clip,keepaspectratio]{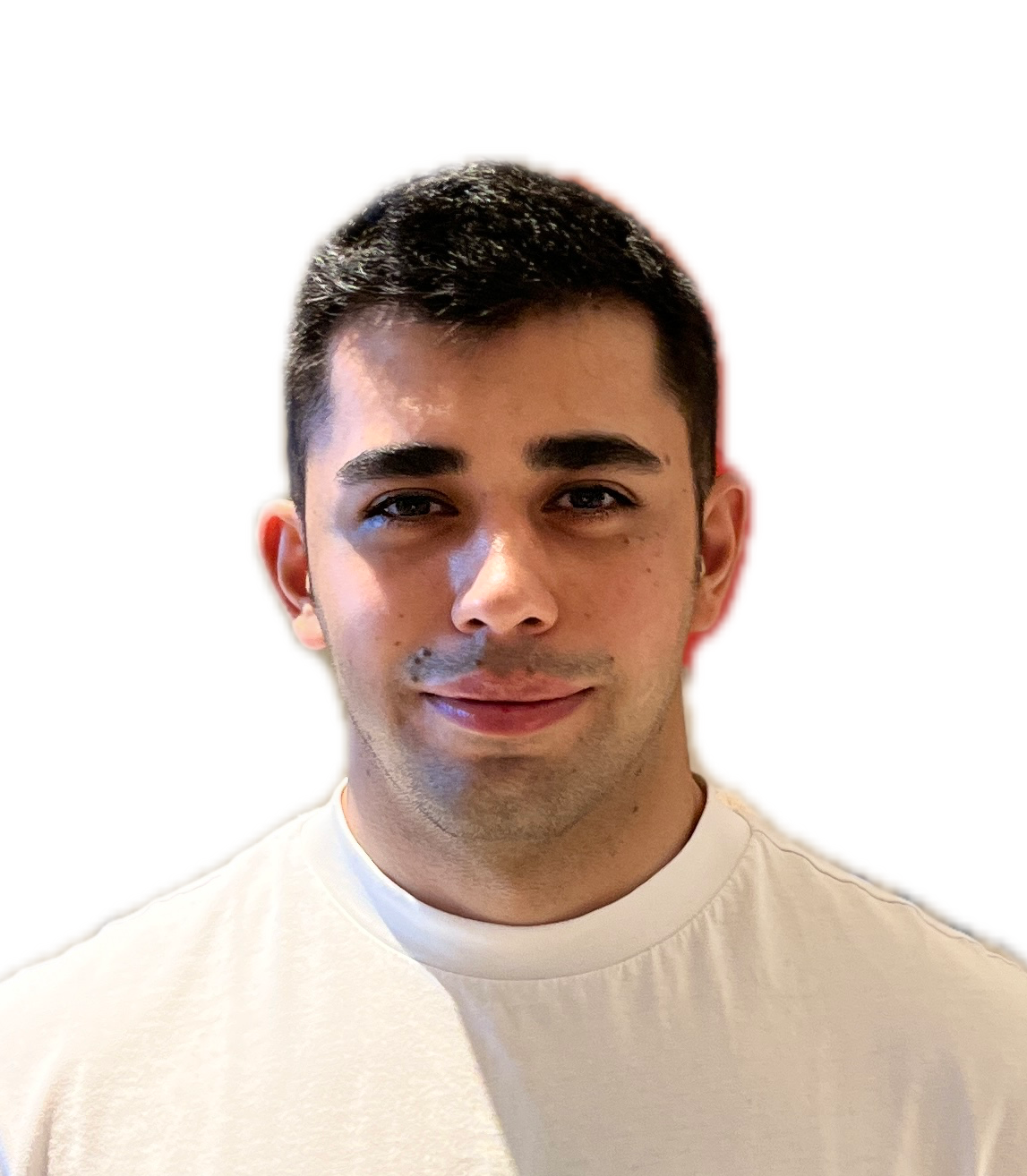}}]
{Mert Cokelek} received his Bachelor's degree in Computer Science from Hacettepe University, Ankara, Turkey, in 2021, and his Master's degree in Computer Science from Ko\c{c} University, Istanbul, Turkey, in 2023. His research interests include computer vision focused on omnidirectional videos, audio-visual saliency prediction and generative AI.
\end{IEEEbiography}

\vskip -2\baselineskip plus -1fil
\begin{IEEEbiography}
[{\includegraphics[width=1in,height=1.25in,clip,keepaspectratio]{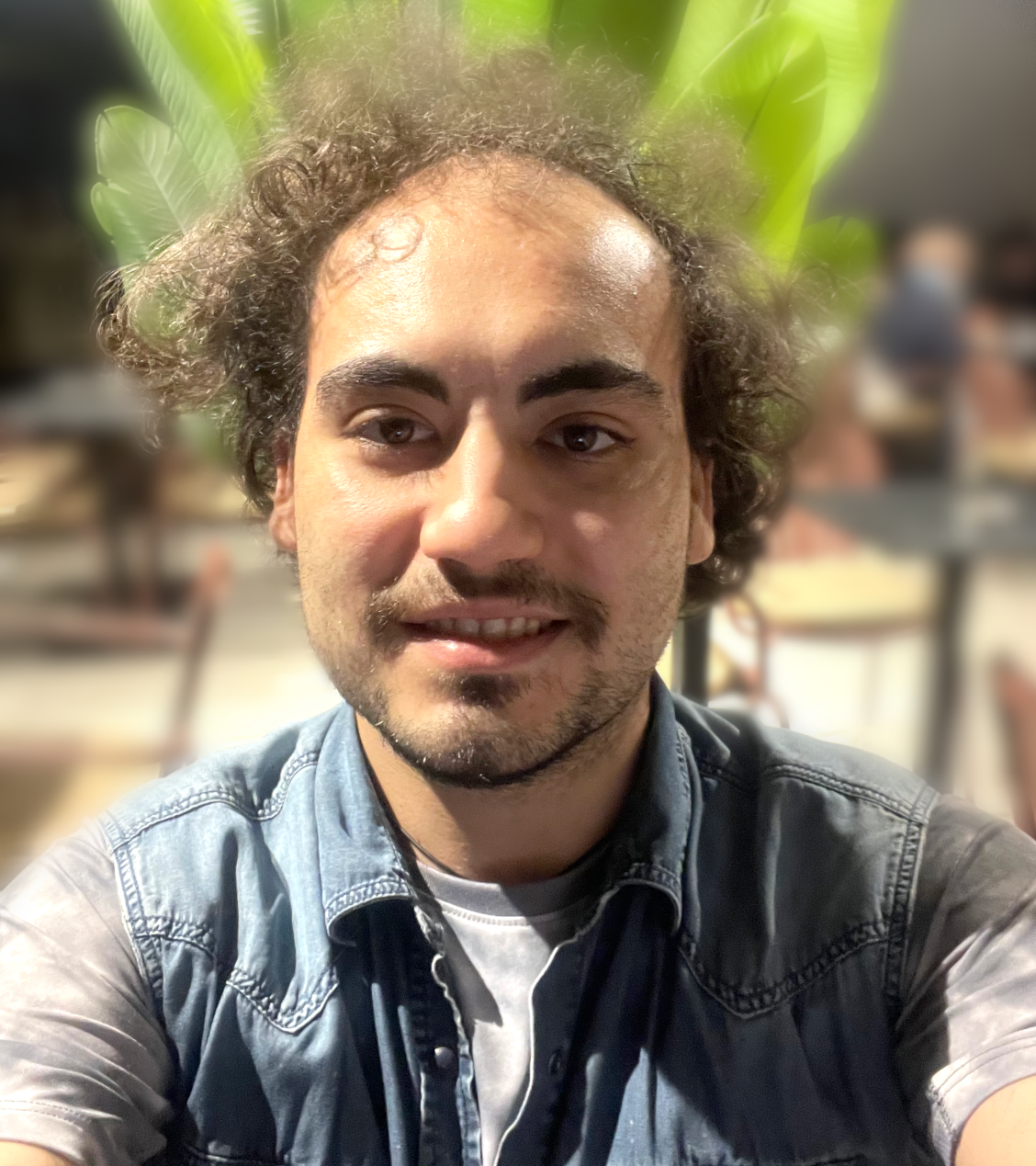}}]
{Halit Ozsoy} is currently pursuing his M.A. degree in the Cognitive Science Program at Boğaziçi University, specializing in psychology and computer science. He completed his undergraduate studies in Computer Engineering at the same university. His research primarily focuses on analyzing the impact of specific semantic contextual information on audio-visual saliency in omnidirectional environments.
\end{IEEEbiography}

\vskip -2\baselineskip plus -1fil
\begin{IEEEbiography}
[{\includegraphics[width=1in,height=1.25in,clip,keepaspectratio]{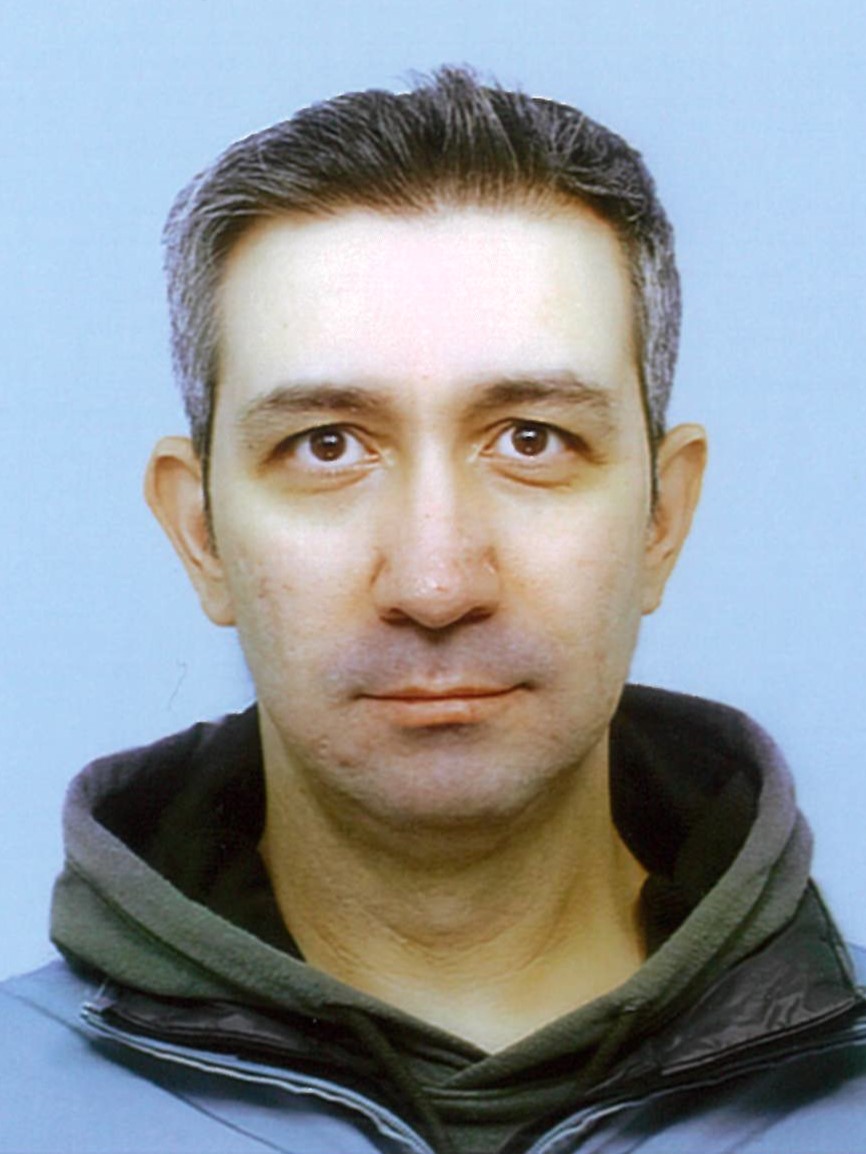}}]
{Nevrez Imamoglu} is a Senior Researcher at National Institute of Advanced Industrial Science and Technology (AIST), Japan since 2016. Before AIST, he was with RIKEN BSI as a JSPS Foreign Postdoctoral Fellow from 2015 to 2016. He received Ph.D. from Chiba University, Japan in 2015. He was also with Nanyang Technbological University, Singapore as a Research Associate from 2010 to 2011. His research interests are applications of signal image processing, computer vision and machine learning.
\end{IEEEbiography}

\vskip -2\baselineskip plus -1fil
\begin{IEEEbiography}
[{\includegraphics[width=1in,height=1.25in,clip,keepaspectratio]{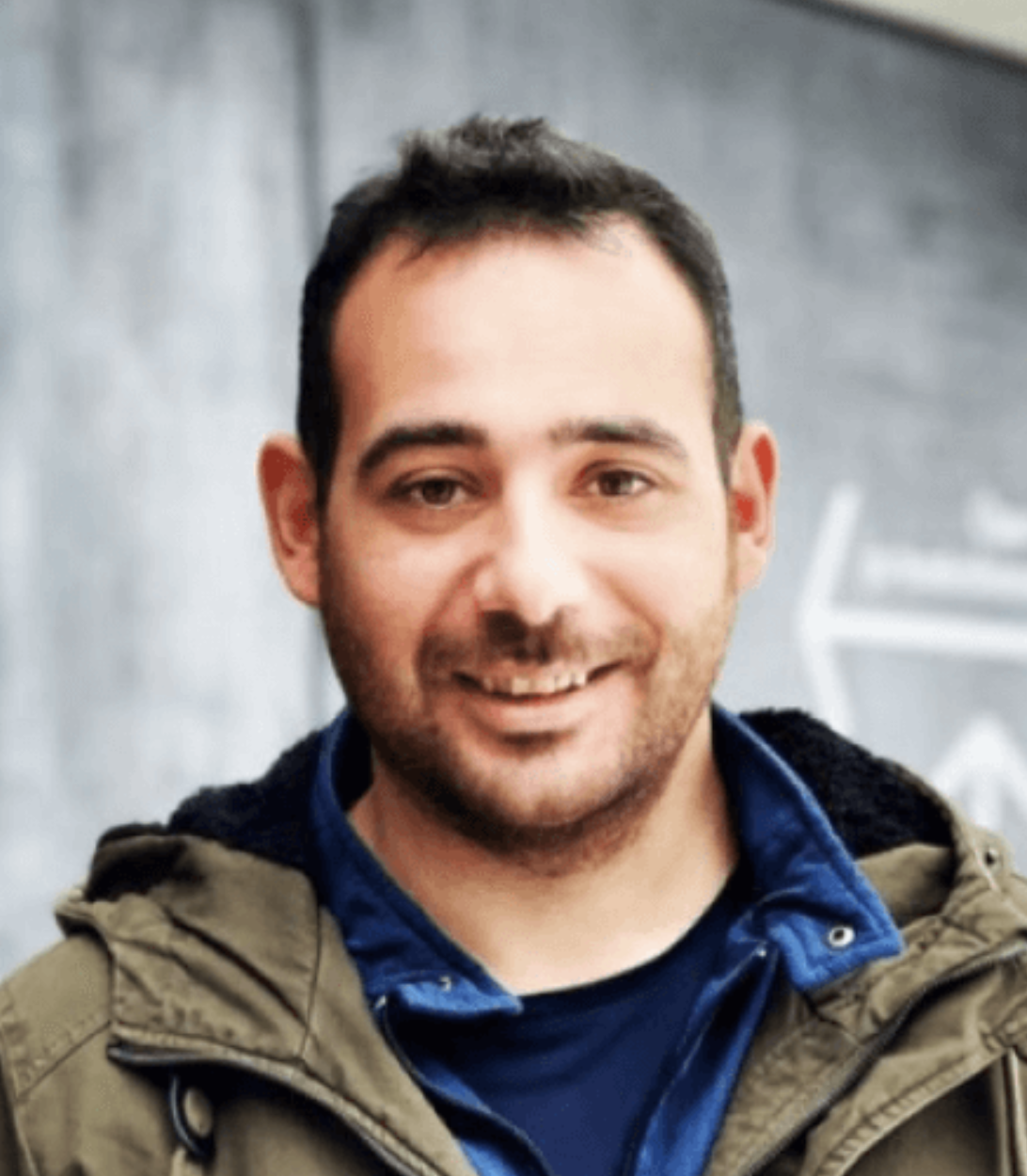}}]
{Cagri Ozcinar} is a machine learning scientist with expertise in image processing, video streaming, and computer vision solutions for embedded devices. His expertise also extended to immersive media technologies and speech processing areas. He held academic research positions in Ireland, UK, and EU Universities. He earned his M.Sc. and Ph.D. degrees from the University of Surrey in the UK. He authored/co-authored over 100 scientific works, including journal and conference papers, as well as books.
\end{IEEEbiography}

\vskip -2\baselineskip plus -1fil
\begin{IEEEbiography}
[{\includegraphics[width=1in,height=1.25in,clip,keepaspectratio]{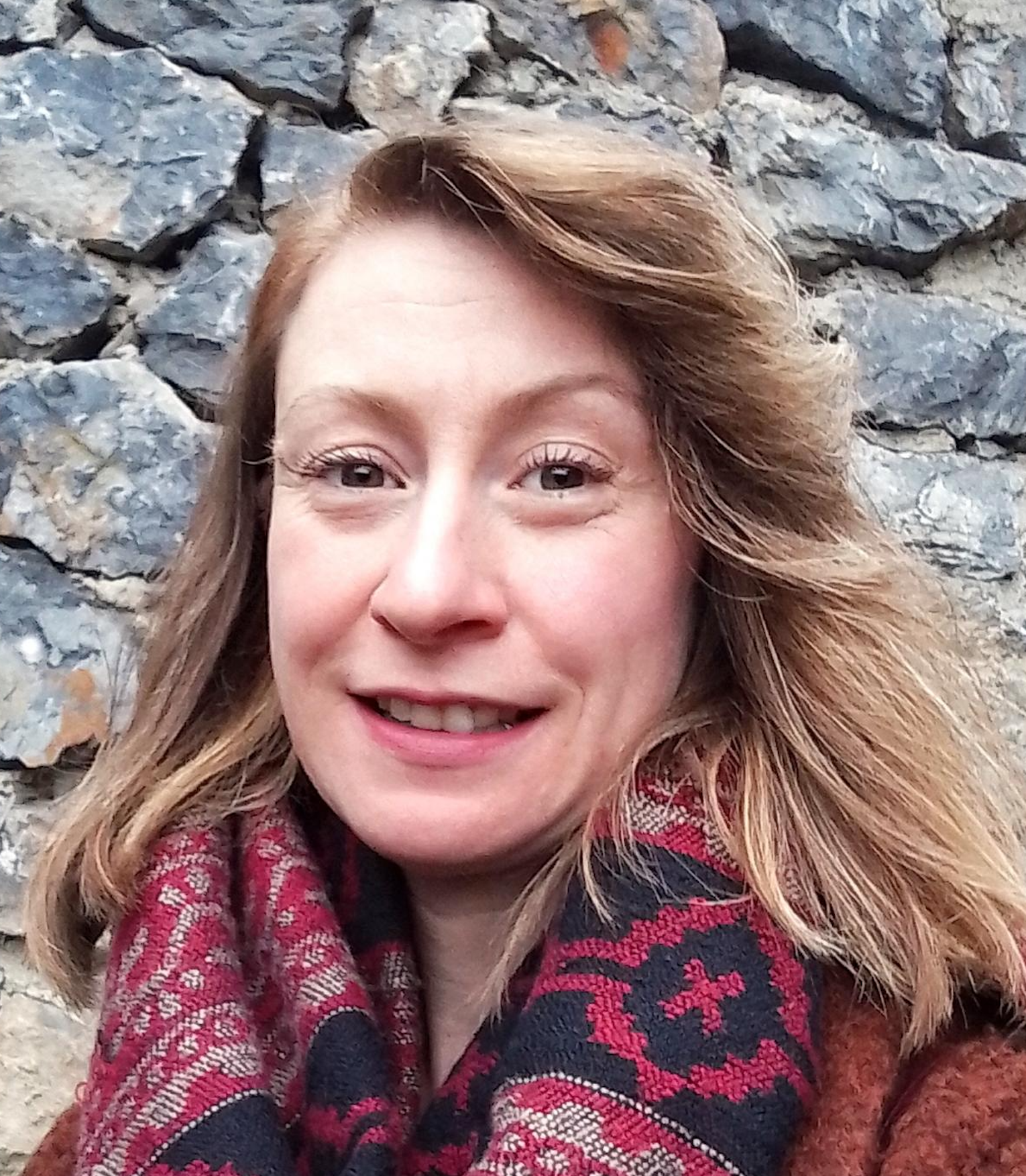}}]
{Inci Ayhan} is an Associate Professor in the Department of Psychology at Boğaziçi University, Turkey. She received her Ph.D. from University College London, UK, in 2010. Following her doctorate, she held postdoctoral positions at University College London and University of London. Her research primarily explores visual perception, temporal processing in the visual system, time perception, and the interactions between the visual and motor systems. 
\end{IEEEbiography}

\vskip -2\baselineskip plus -1fil
\begin{IEEEbiography}
[{\includegraphics[width=1in,height=1.25in,clip,keepaspectratio]{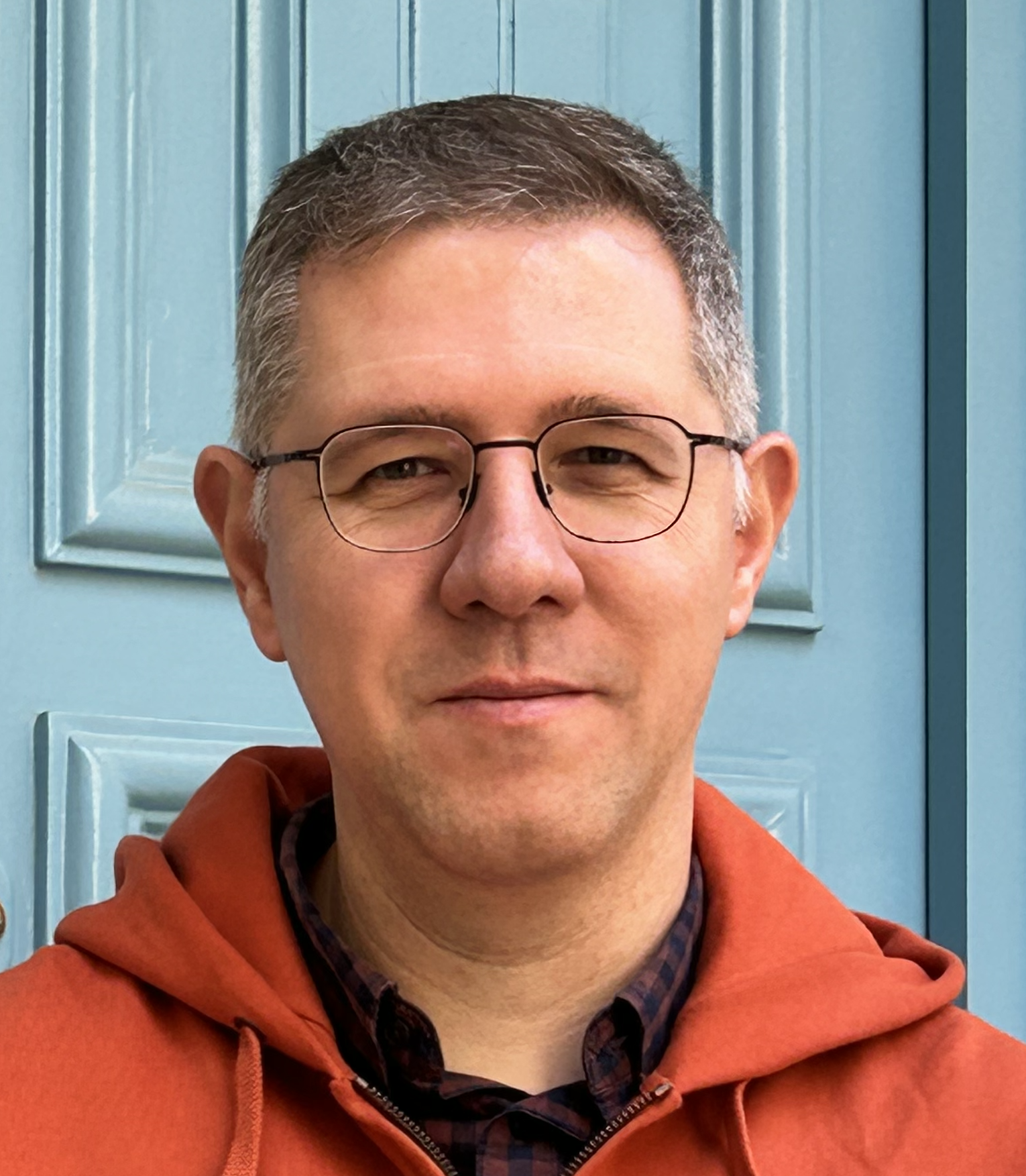}}]
{Erkut Erdem} is a Professor with the Department of Computer Engineering, Hacettepe University, Turkey. He received his Ph.D. degree from Middle East Technical University in 2008. After completing his Ph.D., he continued his post-doctoral studies with Télécom ParisTech, École Nationale Supérieure des Télécommunications, France, from 2009 to 2010. His research interests include semantic image editing, visual saliency prediction, and multimodal machine learning.
\end{IEEEbiography}

\vskip -2\baselineskip plus -1fil
\begin{IEEEbiography}
[{\includegraphics[width=1in,height=1.25in,clip,keepaspectratio]{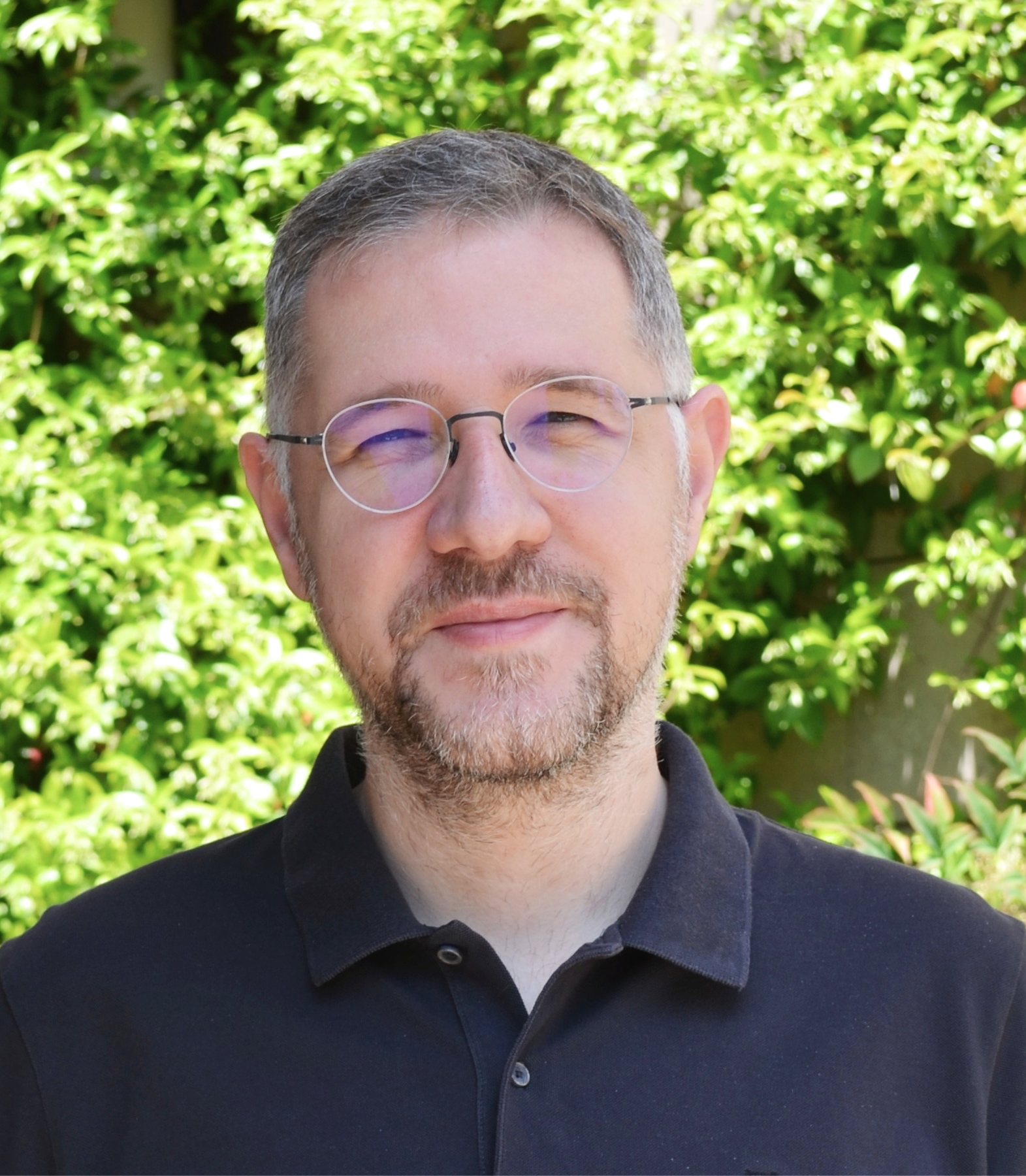}}]
{Aykut Erdem} is an Associate Professor of Computer Science at Ko\c{c} University. He received his Ph.D. from Middle East Technical University in 2008 and was a post-doctoral researcher at the Ca'Foscari University of Venice from 2008 to 2010. Previously, he was with Hacettepe University. His research interests lie in computer vision and machine learning, with a particular focus on semantic image editing and vision–language integration.
\end{IEEEbiography}

\clearpage

\renewcommand\thefigure{A.\arabic{figure}}    
\renewcommand\thetable{A.\arabic{table}}    
\renewcommand\theequation{A.\arabic{equation}}    

\setcounter{figure}{0}    
\setcounter{table}{0}    
\setcounter{equation}{0}    

\appendices
\section*{Supplementary Material}

\textbf{Overview.} The supplementary material has the following structure:
\begin{itemize}
\item Appendix~\ref{appdx-salvit360} provides more details about our proposed audio-blind SalViT360 model, offering extensive analyses to understand its performance and underlying mechanics.
\item Appendix~\ref{appdx-av} gives details about the implementation aspects of our SalViT360-AV model and conduct thorough analyses on the influence and effectiveness of the audio backbone in enhancing our model's capabilities.
\item Appendix~\ref{appdx-dataset} presents additional analysis on our YT360-EyeTracking dataset, including a breakdown of fixation patterns across audio categories and an investigation into the role of audio modalities in guiding attention and inter-subject consistency.
\item Appendix~\ref{appdx-vqa} evaluates the performance of our SalViT360 model through omnidirectional image quality assessment task.

\end{itemize}
\section{SalViT360 Additional Experiments} \label{appdx-salvit360}
This section presents additional experiments on the proposed {Viewport Spatio-Temporal Attention (VSTA)} mechanism and the implementation details of our proposed {Viewport Augmentation Consistency (VAC)} loss.
\subsection{Temporal Window Size} 
\begin{figure*}[!t]
        \centering
        \begin{subfigure}[b]{0.24\textwidth}
            \centering
            \includegraphics[width=\textwidth]{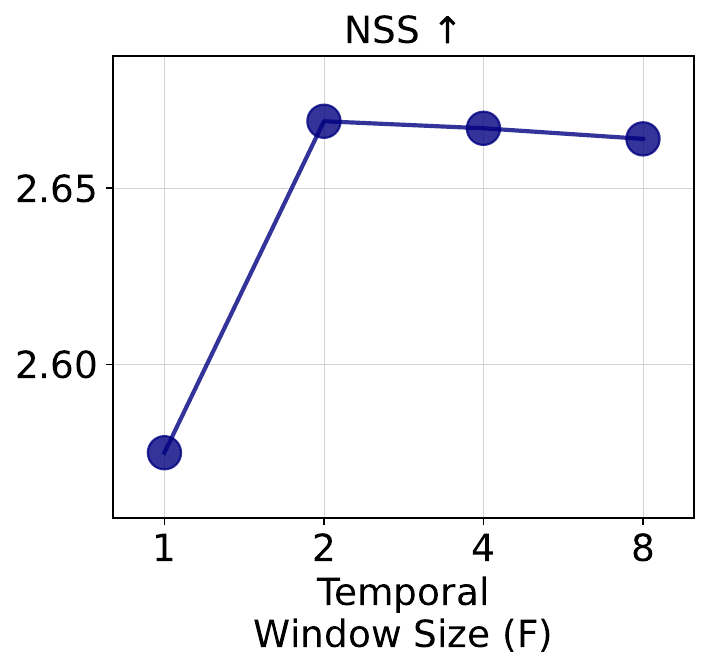}
        \end{subfigure}
        \begin{subfigure}[b]{0.24\textwidth}  
            \centering 
            \includegraphics[width=\textwidth]{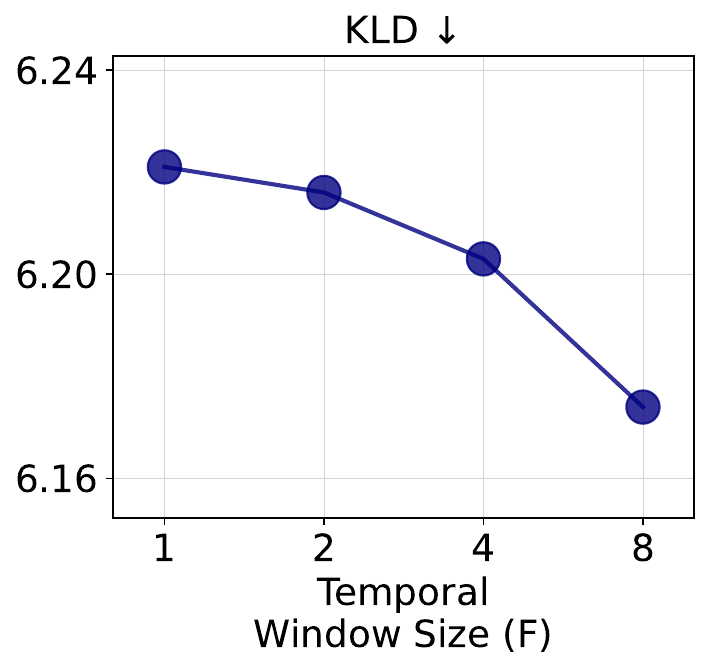}
        \end{subfigure}
        \begin{subfigure}[b]{0.25\textwidth}   
            \centering 
            \includegraphics[width=\textwidth]{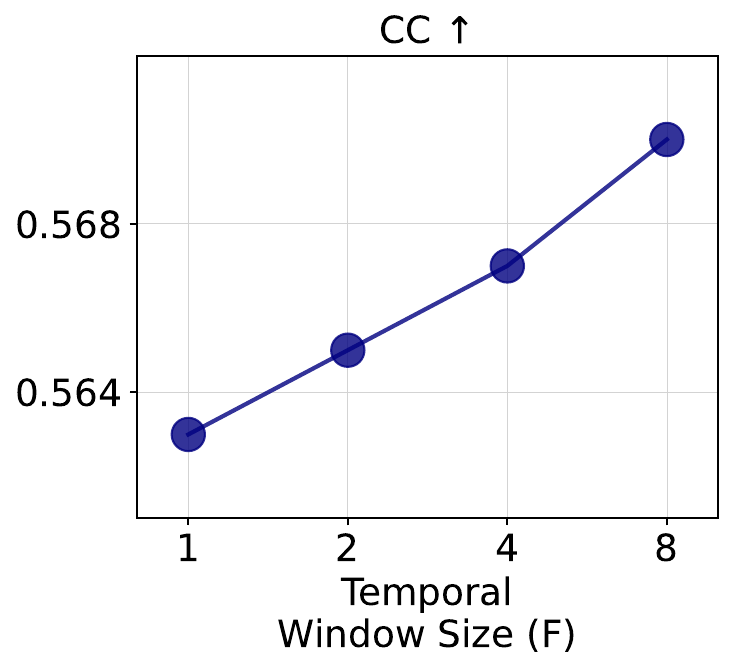}
        \end{subfigure}
        \begin{subfigure}[b]{0.25\textwidth}   
            \centering 
            \includegraphics[width=\textwidth]{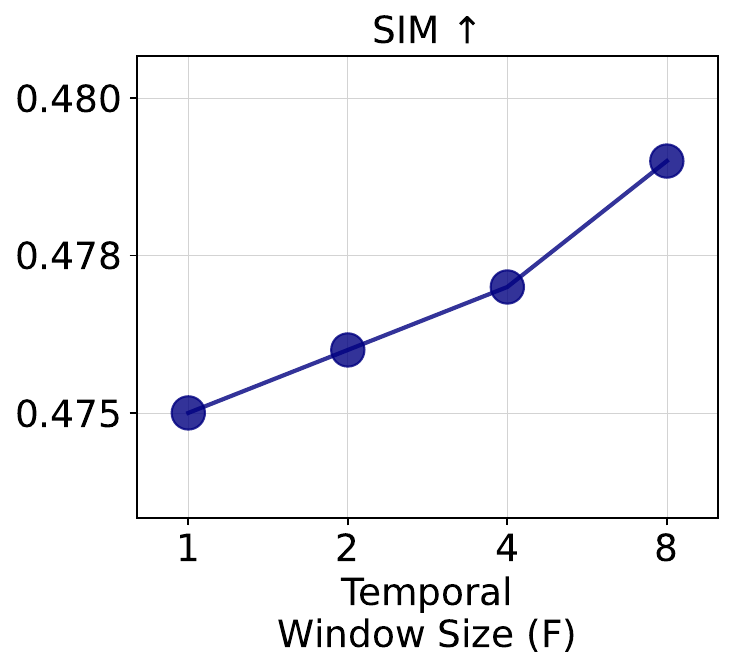}
        \end{subfigure}
        \caption{Performance of our ODV saliency prediction model in terms of four evaluation metrics {(NSS, KLD, CC, SIM)} as a function of temporal window size ($F$) on the validation split of VR-EyeTracking~\cite{xu2018gaze} dataset.} 
        \label{fig:TWS_performance}
\end{figure*}

To investigate the impact of temporal window size ($F$) on the representational power of our omnidirectional video saliency prediction model, we vary the number of frames in a video clip and analyze its effect. These experiments are performed on our VSTA baseline, which consists of 6 transformer blocks with an embedding dimension of $D=512$ and 8 attention heads. We present the results of our experiments using four saliency evaluation metrics in Fig.~\ref{fig:TWS_performance}, providing insights into the performance of our model across different temporal window sizes.

Fig.~\ref{fig:TWS_performance} shows that increasing temporal window size gradually leads to a performance boost in three metrics and an insignificant performance drop in NSS for $F > 2$. We conclude our experiments at $F=8$, considering the memory limit of a single Tesla V100 GPU.

\subsection{Transformer Depth} 
\begin{figure*}[!ht]
        \centering
        \begin{subfigure}[b]{0.24\textwidth}
            \centering
            \includegraphics[width=\textwidth]{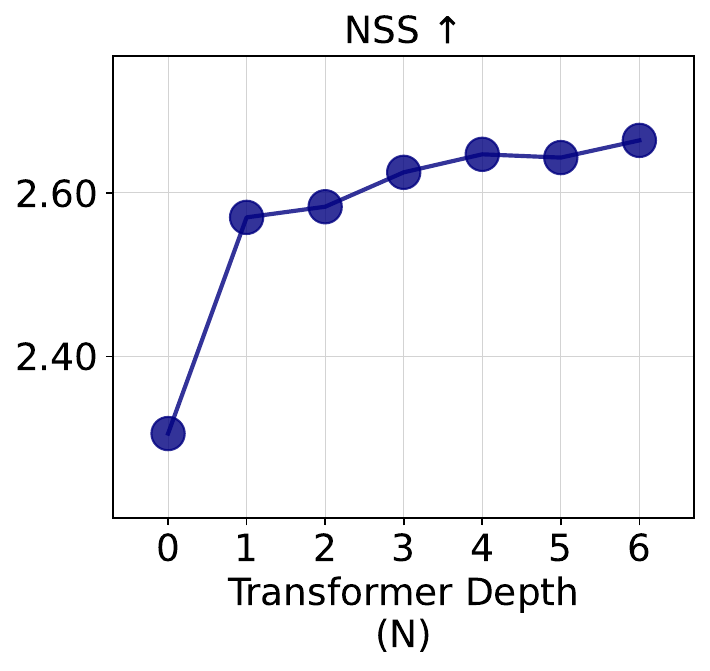}
        \end{subfigure}
        \begin{subfigure}[b]{0.24\textwidth}  
            \centering 
            \includegraphics[width=\textwidth]{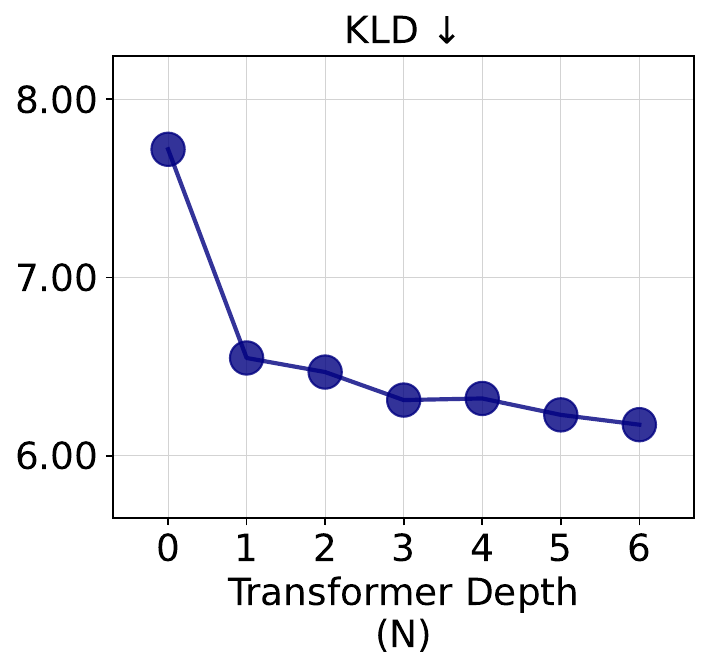}
        \end{subfigure}
        \begin{subfigure}[b]{0.24\textwidth}   
            \centering 
            \includegraphics[width=\textwidth]{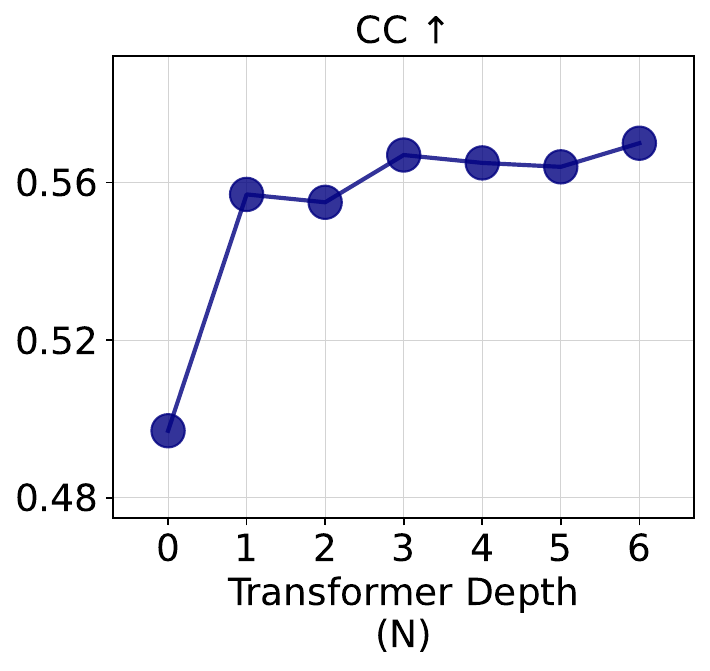}
        \end{subfigure}
        \begin{subfigure}[b]{0.24\textwidth}   
            \centering 
            \includegraphics[width=\textwidth]{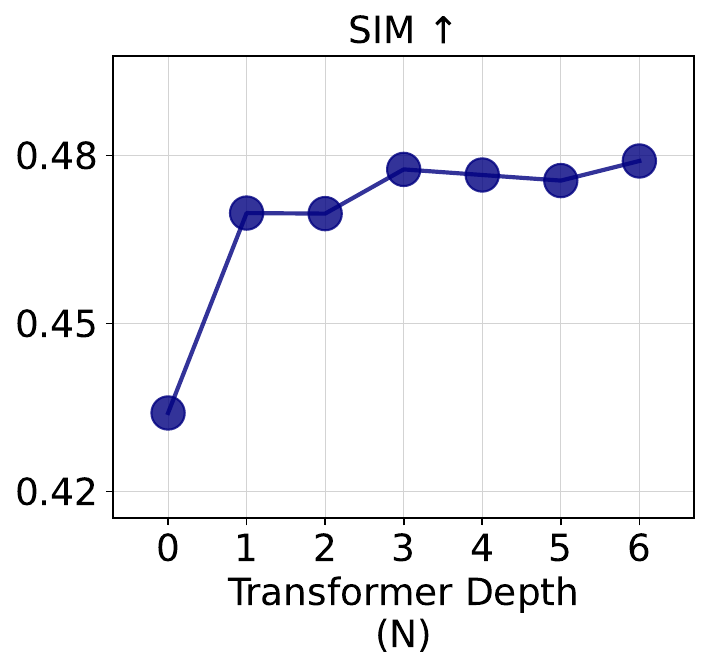}
        \end{subfigure}
        \caption{Performance of our ODV saliency prediction model in terms of four evaluation metrics {(KLD, NSS, CC, SIM)} as a function of VSTA depth ($N$) on the validation split of VR-EyeTracking dataset.} 
        \label{fig:Depth_performance}
\end{figure*}

We analyze the influence of the number of transformer blocks on the performance and the computational complexity of our saliency prediction model for omnidirectional videos. In Fig.~\ref{fig:Depth_performance}, we present our experimental results, showing how the performance varies with different numbers of transformer blocks.

In Fig.~\ref{fig:Depth_performance}, the performance gain from $N=0$ to $N=1$, highlights the effectiveness of our proposed VSTA mechanism for saliency prediction in omnidirectional videos. Notably, even with a single VSTA transformer block, our model shows the ability to capture rich 360$^\circ$ spatio-temporal features. As the depth of the transformer blocks increases, the model performance continues to improve. However, we conclude our experiments after $N=6$ transformer blocks, taking into consideration the model size as reported in Table~\ref{vsta_complexity}.

\subsection{Comparison with Joint Spatio-Temporal Attention} 
Table~\ref{vsta_complexity} compares our proposed VSA and VSTA mechanisms with joint spatio-temporal attention, which computes self-attention among all frames and tokens in a video clip. Since VSTA computes spatio-temporal attention in two stages (time and space), the model size becomes larger than VSA/JSTA. Alternatively, VSTA is computationally more efficient as its complexity grows linearly with respect to temporal window size, which grows quadratically in JSTA.

\begin{table*}[!ht]
            \caption{\textbf{Quantitative comparison} for space-only attention, the proposed Viewport Spatio-Temporal Attention and existing Joint Spatio-Temporal attention for omnidirectional video saliency prediction on the validation split of VR-EyeTracking dataset. Due to memory limitations, JSTA model could not be trained on our GPU.}%
    \centering

        \begin{tabular}{c c c | c c c c} \toprule
            {Attention}  & \# params & GFLOPs & NSS$\uparrow$ & KLD$\downarrow$ & CC$\uparrow$ & SIM$\uparrow$ \\\midrule
            None & 11.81M & 0.00 &2.306 & 7.718 & 0.497 & 0.434\\ 
            VSA  & 30.78M & 57.08 & 2.575 & 6.221 & 0.563 & 0.475\\ 
            VSTA & 37.07M & 63.30 & 2.664 & 6.174 & 0.570 & 0.479\\
            JSTA & 30.78M & 77.57 & {n/a} & {n/a} & {n/a} & {n/a}\\\bottomrule
        \end{tabular}       
        \label{vsta_complexity}
\end{table*}

\subsection{Viewport Augmentation Consistency (VAC)}\label{sec:vac}
\begin{figure*}[!ht]
    \centering
    \includegraphics[trim={0 50 0 10}, width=0.85\textwidth]{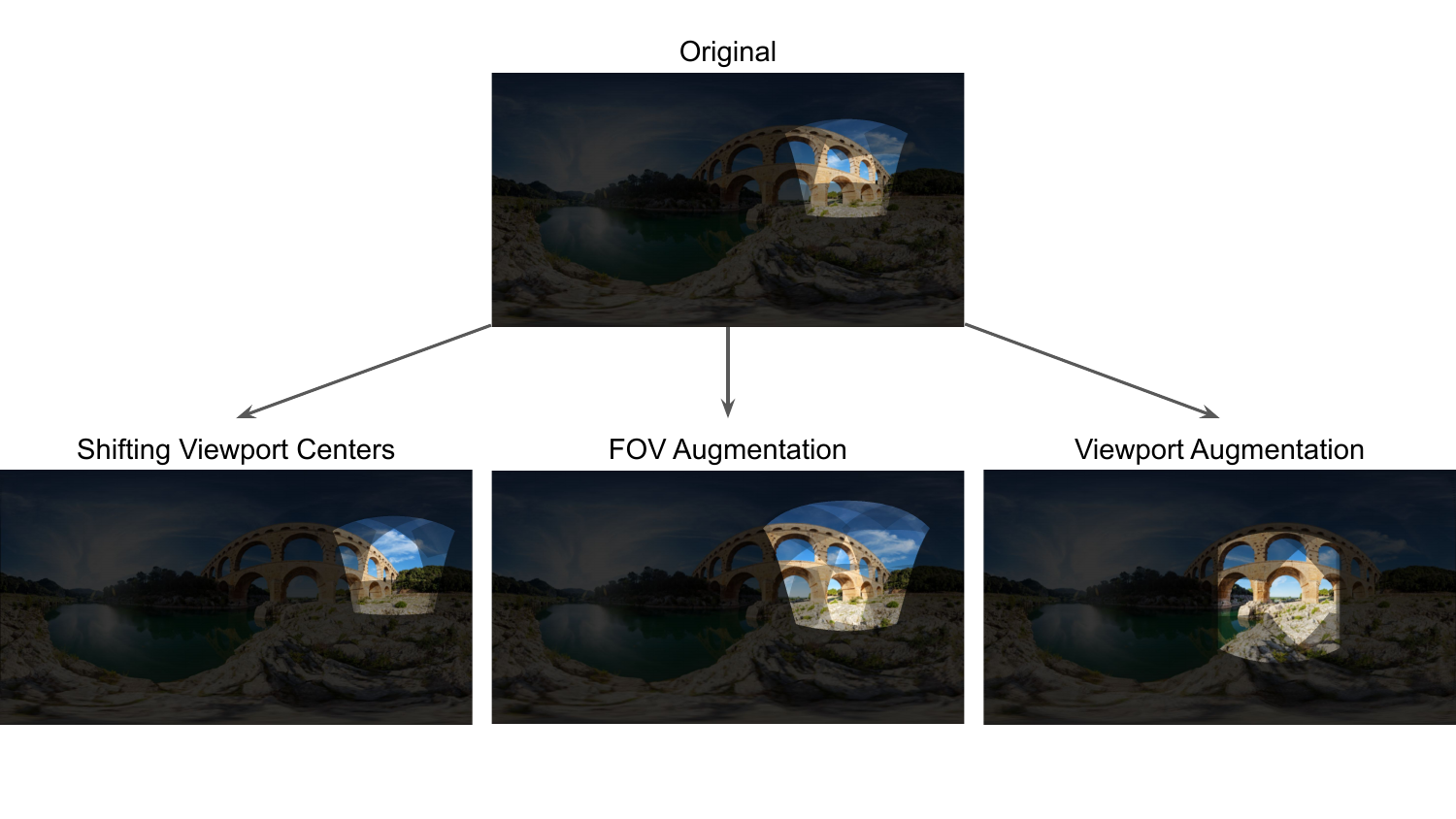}
    \caption{A tangent viewport from the original projection, highlighted on Equirectangular Projection (ERP) {(top)}, compared with three augmentation methods {(bottom)}.}
    \label{fig:vac_approaches}
\end{figure*}

In this section, we describe the proposed {Viewport Augmentation Consistency} in more detail. To address the discrepancies in overlapping regions of tangent predictions, we propose to use a second \textit{augmented} tangent image set and minimize the difference between the predictions of these pairs with an additional loss term. \noindent Following \cite{li2022omnifusion}, we sampled $T=18$ tangent images at four latitudes: $-67.5^\circ, -22.5^\circ, 22.5^\circ, 67.5^\circ$ for the original set. The tangent images are sampled for each latitude level with $90^\circ$ apart in longitude. We extracted each tangent image with a resolution of $224\times224$ and field-of-view (FOV) of $80^\circ$. We generated the augmented tangent image set under three configurations: (1) horizontally shifting viewports, (2) using a larger FOV for each viewport, and (3) varying the number, position, and FOV of the tangent viewports (Fig.~\ref{fig:vac_approaches}). 

\noindent\textbf{Shifting Viewport Centers.} In this setting, we keep the number and FOV of tangent images the same. We obtain the shifted tangent image set by applying a $45^\circ$ horizontal shift on each viewport. 

\noindent\textbf{FOV Augmentation.} In the second set, we keep the position of each tangent image the same and generate the augmented set by increasing their FOV to $120^\circ$. Augmented FOV also provides the model with a multi-scale representation for the same input. 

\noindent\textbf{Viewport Augmentation.} In the last setting, we generate the second set with $T'=10$ tangent images with a FOV of $120^\circ$, located in three latitudes: $-60^\circ, 0^\circ, 60^\circ$. We sample $3, 4, 3$ viewports for each latitude.

\begin{figure*}[!h]       
        \begin{subfigure}[b]{0.22\textwidth}
            \centering
            \includegraphics[width=\textwidth]{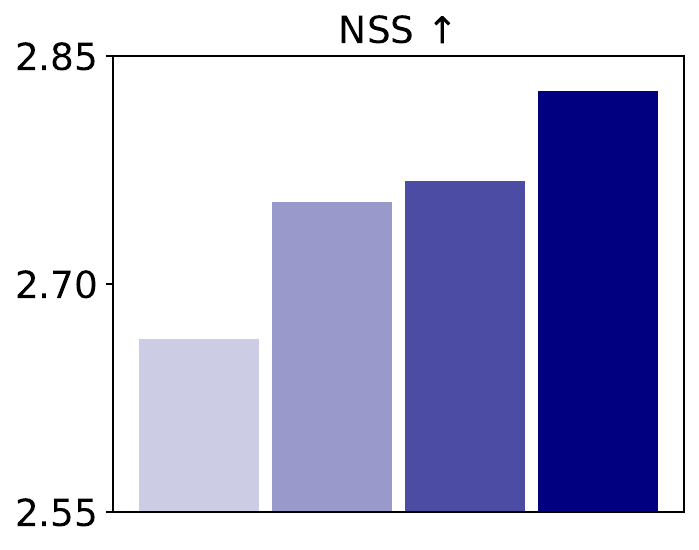}
        \end{subfigure}
        \begin{subfigure}[b]{0.22\textwidth}  
            \centering 
            \includegraphics[width=\textwidth]{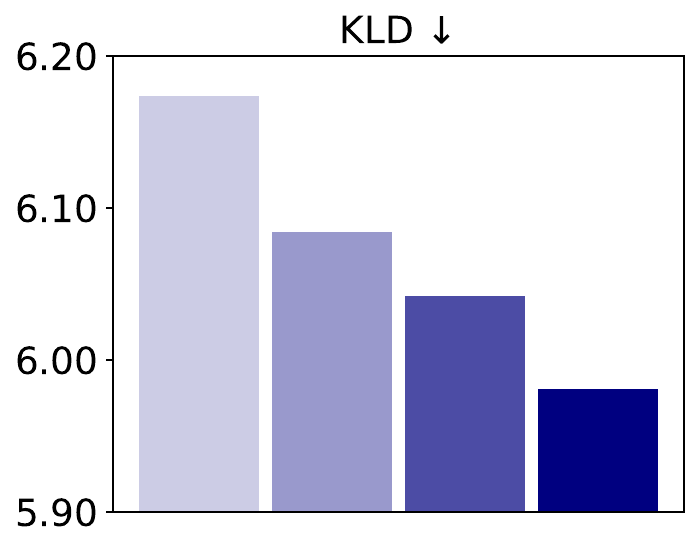}
        \end{subfigure}
        \begin{subfigure}[b]{0.22\textwidth}   
            \centering 
            \includegraphics[width=\textwidth]{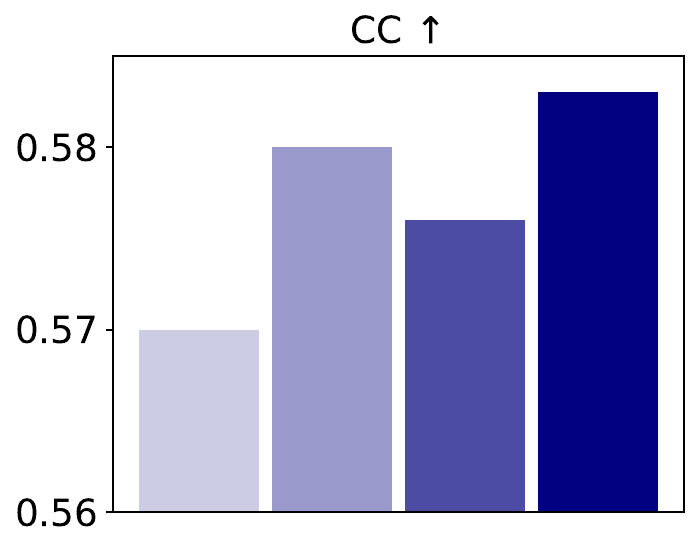}
        \end{subfigure}
        \begin{subfigure}[b]{0.22\textwidth}   
            \centering 
            \includegraphics[width=\textwidth]{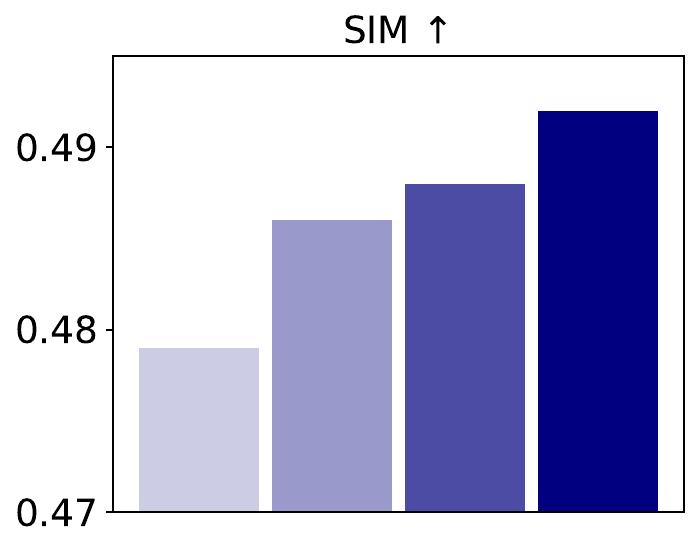}
        \end{subfigure}
                \centering
        \begin{subfigure}[b]{1.0\textwidth}
                \centering
                \includegraphics[width=0.65\textwidth]{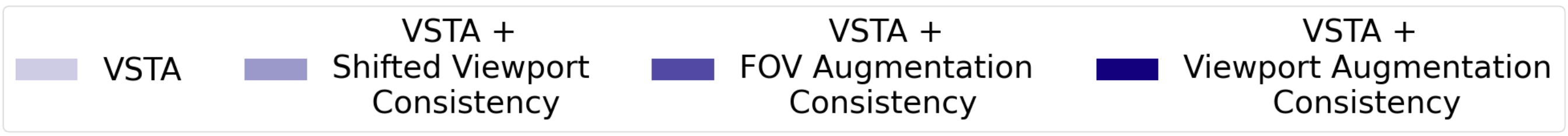}
        \end{subfigure}
        \caption{Performance of our VSTA baseline compared with three augmentation consistency methods on four metrics, on the validation split of VR-EyeTracking dataset.} 
        \label{fig:VAC}
\end{figure*}

It is important to emphasize that the augmented tangent images share weights with the original set, which neither requires extra parameters nor increases model complexity during training. Our experimental results in Fig.~\ref{fig:VAC} show that each augmentation method improves model consistency significantly.

\begin{figure}[!h]
    \includegraphics[width=0.75\textwidth]{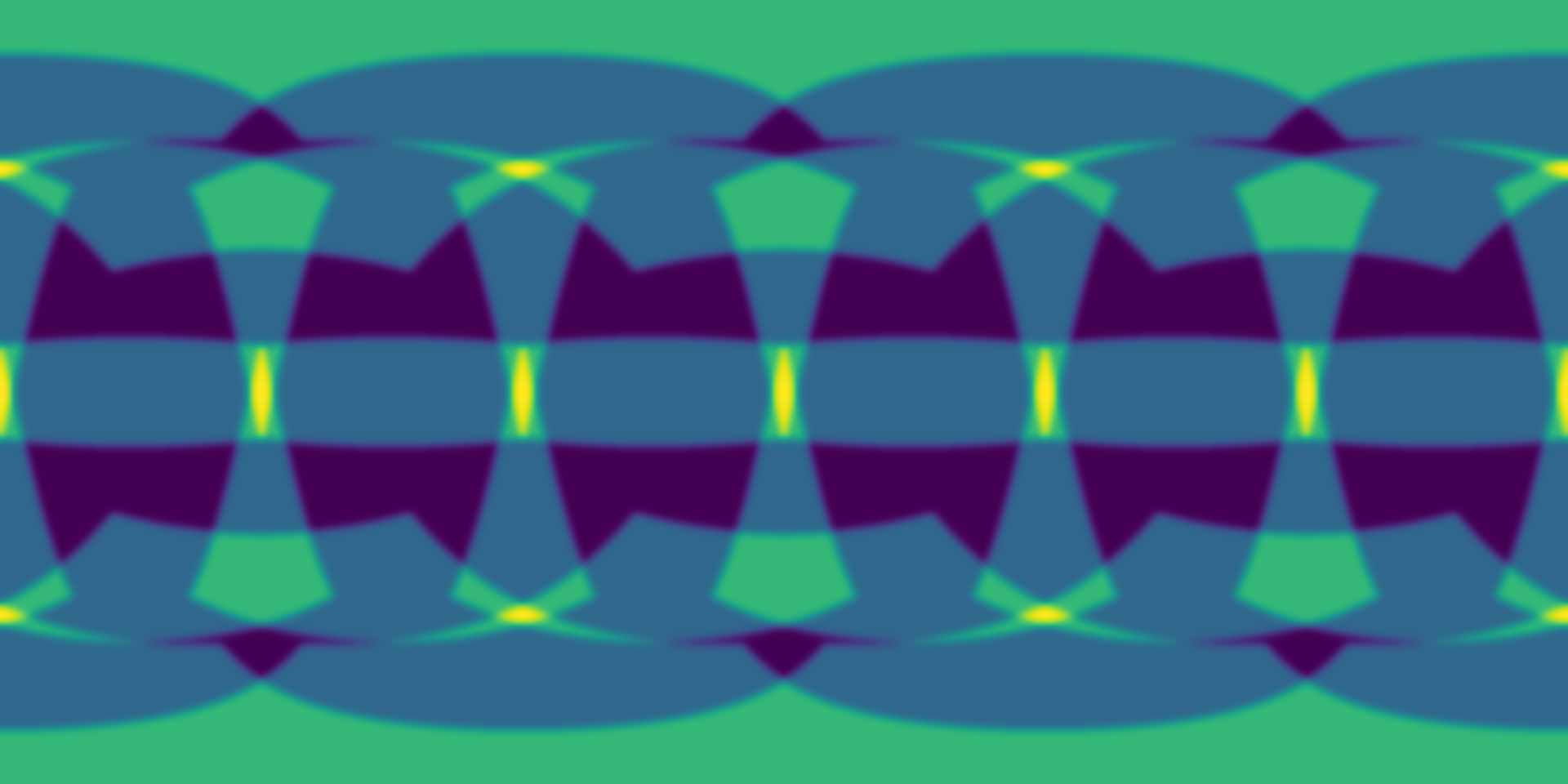}

 \caption{\textbf{Weight mask used for VAC Loss.} Each pixel coordinate in ERP takes a value based on the number of tangent viewports it is projected onto (max. 4). Brighter colors represent increasing overlaps.} 
      \label{mask}
    \end{figure}

\noindent\textbf{Mask-weighted VAC Loss.}
We use an optional weight for the proposed $\mathcal{L}_{\text{VAC}}(P,\ P')$ loss to increase consistency, especially on the overlapping regions on ERP. In Fig.~\ref{mask}, we provide the weight mask computed from the gnomonic projection for the original tangent image set. The performance comparison in Table~\ref{masktable} demonstrates the effectiveness of the proposed masking operation.

    \begin{table}[!t]
         \caption{\textbf{Quantitative comparison} for the proposed VAC Loss with and without mask, on the validation split of the VR-EyeTracking dataset. }%
        \begin{tabular}{l  c@{$\;$}c@{$\;$}c@{$\;$}c} \toprule
            {Model} & NSS$\uparrow$ & KLD$\downarrow$ & CC$\uparrow$ & SIM$\uparrow$ \\\midrule
            \shortstack{VSTA $+$ VAC \\{\it (w/o mask)}}  & 2.624 & 6.011 & 0.576 & 0.490\\\midrule
            \shortstack{VSTA $+$ VAC \\{\it (w/ mask)}}   & 2.630 & 5.744 & 0.586 & 0.492\\\bottomrule
        \end{tabular}            \label{masktable}

    \end{table}

\renewcommand\thefigure{B.\arabic{figure}}    
\renewcommand\thetable{B.\arabic{table}}    
\renewcommand\theequation{B.\arabic{equation}}

\setcounter{figure}{0}    
\setcounter{table}{0}    
\setcounter{equation}{0}  

\section{SalViT360-AV Details}\label{appdx-av}
In this section, we first formally describe the rotation and decoding of first-order ambisonics, the spatial audio encoding method used in our audio-visual pipeline. Then, we compare three audio backbones described in our main paper and investigate the effect of audio clip duration on saliency prediction performance.

\subsection{Spatial Audio}
First-order Ambisonics (FOA) is a sound encoding technique that captures and reproduces audio, allowing immersive listening experiences. 
This method diverges from the traditional stereo or surround sound systems, which rely on discrete speakers arranged around the listener to simulate spatial audio. Instead, Ambisonics adopts a spherical approach to sound representation, enabling the playback of audio from any direction surrounding the listener. FOA integrates sound information from multiple directions into a single audio signal. This signal can then be decoded and played back from any direction, offering a versatile listening experience. The encoding process thus captures the direction of sound waves by recording and mixing the sound sources according to the positions of each source relative to a reference point on the sphere. Ambisonics approximates the sound pressure field at a single point in space using a spherical harmonic decomposition. 

More specifically, an audio signal $f((\theta, \phi), t)$ is represented by a spherical harmonic expansion of order $N$
\begin{equation}
    f((\theta, \phi), t) = \sum\nolimits_{n=0}^{N} \sum\nolimits_{m=-n}^{n} Y_n^m(\theta, \phi) \alpha_n^m(t)\;,
\label{eqn:foa-signal}
\end{equation}
where $\theta$ and $\phi$ denote the zenith and the azimuth angles at time $t$, respectively, $Y_n^m(\theta, \phi)$ is the spherical harmonic of order $n$ and degree $m$, and $\alpha_n^m(t)$ are the expansion coefficients. For simplicity, (Eq.~\ref{eqn:foa-signal}) can be written as $f((\theta, \phi), t) = \bm{\mathit{y}}_N \bm{\alpha}_N(t).$

\noindent\textbf{Encoding.} 
Given a set of \textit{k} audio signals $s_1(t), \dots, s_k(t)$ from directions $\theta_1, \dots, \theta_k$,
\begin{equation}
    \bm{\alpha}_N(t) = \sum\nolimits_{i=1}^k\bm{\mathit{y}}_N((\theta_i, \phi_i))s_i(t).
\end{equation}

The spherical harmonic coefficients, denoted as $\bm{\alpha}_N$, or ambisonic channels ($\bm{\alpha}_w, \bm{\alpha}_x, \bm{\alpha}_y, \bm{\alpha}_z$), are integral to decoding the captured sound into a speaker array, facilitating the reconstruction of the sound field. FOA employs four distinct channels to represent the first-order coefficients of the spherical harmonic expansion: $\alpha_0^0$, $\alpha_0^{-1}$, $\alpha_1^0$, and $\alpha_1^1$. These channels encapsulate the sound information in a way that allows for a spatially immersive playback, reproducing sound from any direction on the sphere.

\noindent\textbf{Rotation and Decoding.} 
Adjusting the orientation of the sound field to match the listener's perspective involves rotating the spherical harmonic coefficients, $\bm{\alpha}_N$, with a rotation matrix, $\mathbf{R}$. For FOA, this rotation is executed using a 3x3 matrix tailored to the specific rotation defined by Euler angles (yaw, pitch, and roll). This process, applied prior to decoding, modifies the ambisonic channels directly. By manipulating these coefficients, it is possible to precisely alter the directionality of the sound field, ensuring that it aligns with the target orientation
\begin{equation}
\bm{\alpha}_N'(t) = \mathbf{R} \cdot \bm{\alpha}_N(t) \;.
\end{equation}\label{eqn:appdx-foa-rotation}

\begin{table*}[!t]
    \centering
    \caption{\textbf{Performance comparison of three pre-trained audio models} as the audio backbone of \avsal, on VR-EyeTracking, 360AV-HM, \bounvr~datasets.}
    \renewcommand{\arraystretch}{1.5} %
    \resizebox{\textwidth}{!}{%
    \begin{tabular}{>{\raggedright\arraybackslash}p{3cm}cccccccccccc}
        \toprule
        & \multicolumn{4}{c}{{\bf VR-EyeTracking~\cite{xu2018gaze}}} & \multicolumn{4}{c}{{\bf 360AV-HM~\cite{9301766}}} & \multicolumn{4}{c}{{\bf YT360-EyeTracking}}\\
        \cmidrule(lr){2-5} \cmidrule(lr){6-9} \cmidrule(lr){10-13}
        {\bf Method} & {\bf NSS$\uparrow$} & {\bf KLD$\downarrow$} & {\bf CC$\uparrow$} & {\bf SIM$\uparrow$} & {\bf NSS$\uparrow$} & {\bf KLD$\downarrow$} & {\bf CC$\uparrow$} & {\bf SIM$\uparrow$} & {\bf NSS$\uparrow$} & {\bf KLD$\downarrow$} & {\bf CC$\uparrow$} & {\bf SIM$\uparrow$} \\
        \midrule        
        AVSA~\cite{morgado2020learning}     & n/a & n/a & n/a & n/a & 2.499 & 12.916 & 0.381 & 0.284 & 2.501 & 7.768 & 0.516 & 0.417 \\
        EZ-VSL~\cite{morgado2020learning}   & 2.817 & 5.342 & 0.598 & 0.509 & 2.472 & 13.833 & 0.380 & 0.277 & 2.449 & 8.288 & 0.511 & 0.407 \\
        PaSST~\cite{koutini22passt}         & 2.821 & 5.334 & 0.599 & 0.511 & 2.473 & 13.830 & 0.379 & 0.278 & 2.449 & 8.341 & 0.512 & 0.407 \\
        \bottomrule
    \end{tabular}%
    }
    \label{table:appdx-audio-backbones}
\end{table*}

\begin{figure*}[!t]
        \centering
        \begin{subfigure}[b]{0.23\textwidth}
            \centering
            \includegraphics[width=\textwidth]{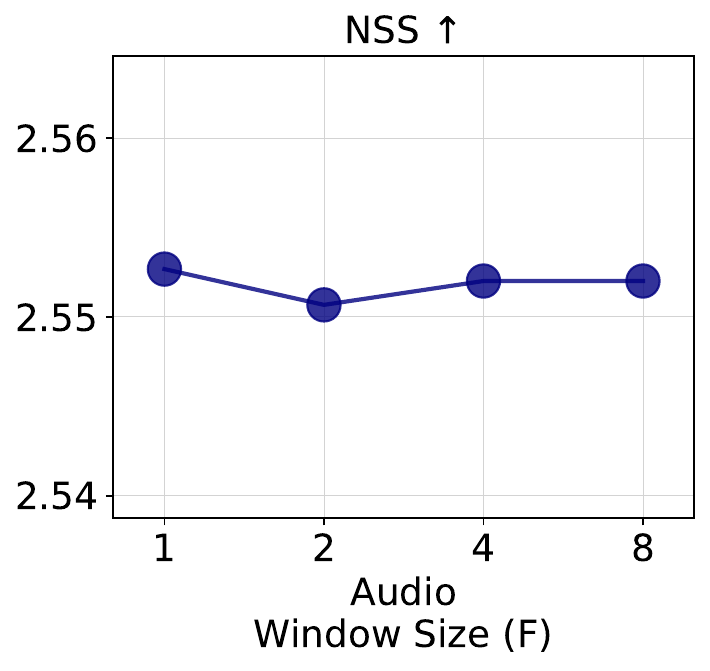}
        \end{subfigure}
        \begin{subfigure}[b]{0.23\textwidth}  
            \centering 
            \includegraphics[width=\textwidth]{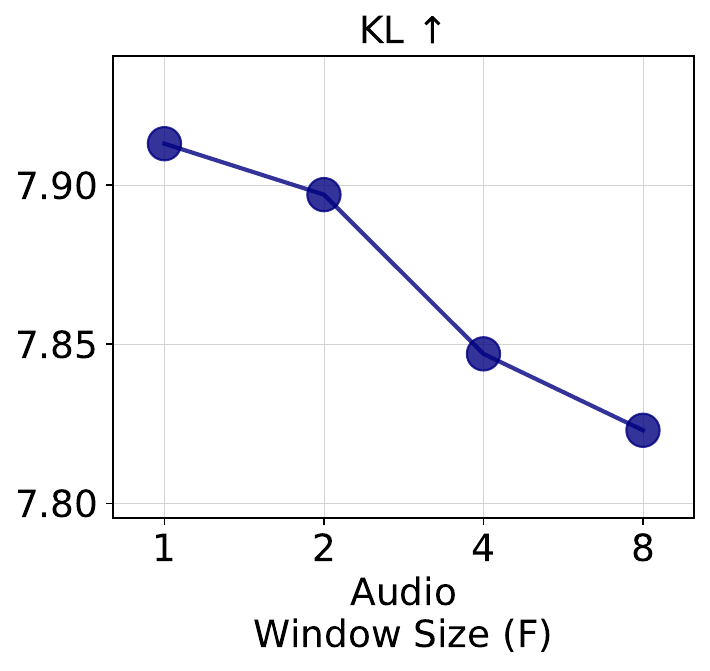}
        \end{subfigure}
        \begin{subfigure}[b]{0.24\textwidth}   
            \centering 
            \includegraphics[width=\textwidth]{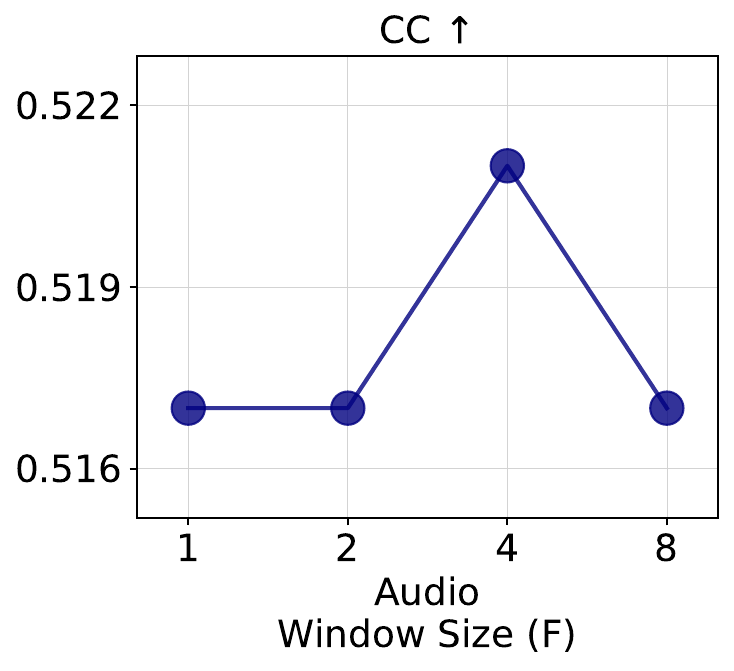}
        \end{subfigure}
        \begin{subfigure}[b]{0.24\textwidth}   
            \centering 
            \includegraphics[width=\textwidth]{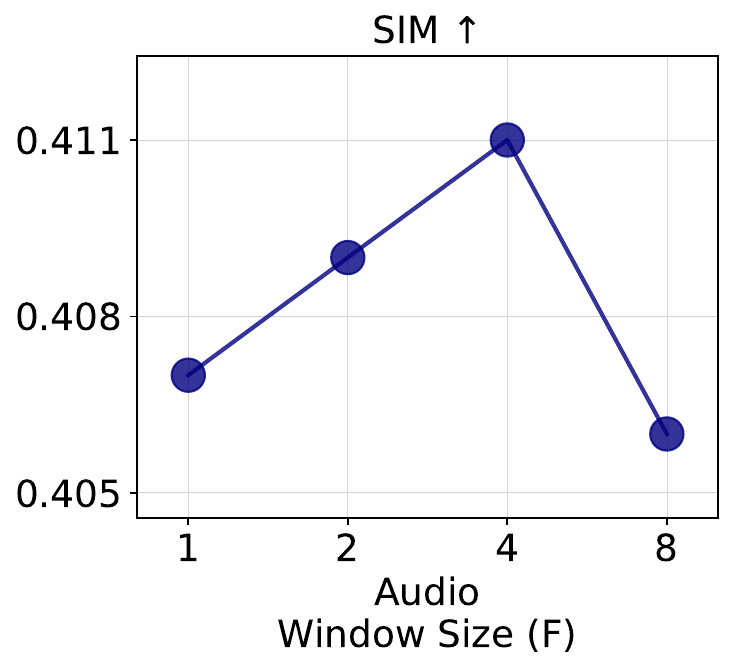}
        \end{subfigure}
        \caption{Performance of our ODV saliency prediction model in terms of four evaluation metrics {(NSS, KLD, CC, SIM)} as a function of audio window size ($F$) on the validation split of \bounvr~dataset.} 
        \label{fig:appdx-audio-window-size}
\end{figure*}
In \salvit~and \avsal, we use $T=18$ tangent viewports located around the sphere centered at $(\theta_t, \phi_t))_{t=1}^T$. For each viewport, we apply the ambisonics channel rotation given in (Eq.~\ref{eqn:appdx-foa-rotation}) to generate new ambisonics. Then, we decode the ambisonics waveforms for the forward direction using (Eq.~\ref{eqn:foa-decoding} of the main manuscript). This way, we are able to obtain viewport-specific waveforms in mono format, capturing the sound sources located in the direction of each tangent viewport.

\subsection{Additional Experiments}
\subsubsection{On Audio Backbones}
\label{sec:exp_audio_backbones}
We compare the performance of three audio backbones in our \avsal~pipeline, namely, PaSST~\cite{koutini22passt}, EZ-VSL~\cite{mo2022localizing}, and AVSA~\cite{morgado2020learning} in Table~\ref{table:appdx-audio-backbones}. PaSST~\cite{koutini22passt} is an Audio Spectrogram Transformer trained for audio classification, distinguished by its self-attention mechanism on mel-spectrogram patches and its patchout-based regularization strategy. PaSST further benefits from initialization with DeiT’s pre-trained ImageNet weights, which are subsequently adapted to AudioSet for improved audio classification. EZ-VSL is a ResNet-18 model fine-tuned on mel-spectrograms for visual sound localization in 2D, while PaSST, an Audio Spectrogram Transformer variant, is trained for audio classification. AVSA employs a 9-layer 2D CNN that operates on mel-spectrograms in the time-frequency domain, specifically optimized for audio-visual spatial alignment within $360^\circ$ environments. While the AVSA encoder, which is specifically trained for first-order ambisonics, yields the best quantitative results, we used PaSST in our main experiments due to its availability on mono audio. 

\subsubsection{On Audio Window Size}
We report the impact of audio window size (in seconds) on saliency prediction performance in Fig.~\ref{fig:appdx-audio-window-size}. The findings suggest that there is an optimal audio window size range that marginally improves the model's ability to predict visual attention across different metrics. Notably, a window size of approximately 4 secs tends to yield slightly better or comparable results relative to smaller or larger window sizes. This analysis highlights the importance of audio temporal context in enhancing the accuracy of \avsal.

\renewcommand\thefigure{C.\arabic{figure}}    
\renewcommand\thetable{C.\arabic{table}}    
\renewcommand\theequation{C.\arabic{equation}}

\setcounter{figure}{0}    
\setcounter{table}{0}    
\setcounter{equation}{0}  

\section{Additional Analysis on Our YT360-EyeTracking Dataset} \label{appdx-dataset}

\subsection{Fixation Distributions Across Audio Categories}

\begin{figure*}[!t]
    \centering
    \includegraphics[trim={0 0 0 0}, clip, height=5.8cm]{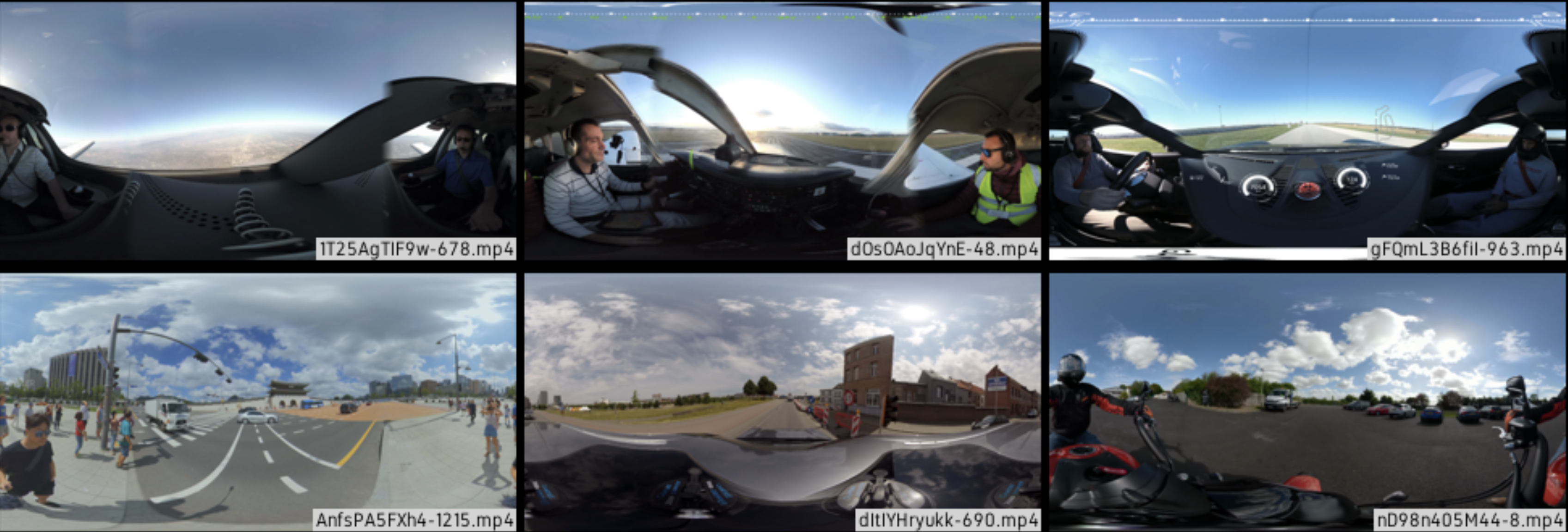}
    \caption{Sample frames from the \textit{Vehicle Sound} category. Scenes typically feature forward-facing driving perspectives, leading to a stronger center bias in fixations as demonstrated in Fig.~\ref{fig:audio-modalities-on-fixation-vehicle}.}
    \label{data:frames-vehicle}
\end{figure*}

\begin{figure*}[!t]
    \centering
    \includegraphics[trim={0 0 0 0}, clip, height=5.8cm]{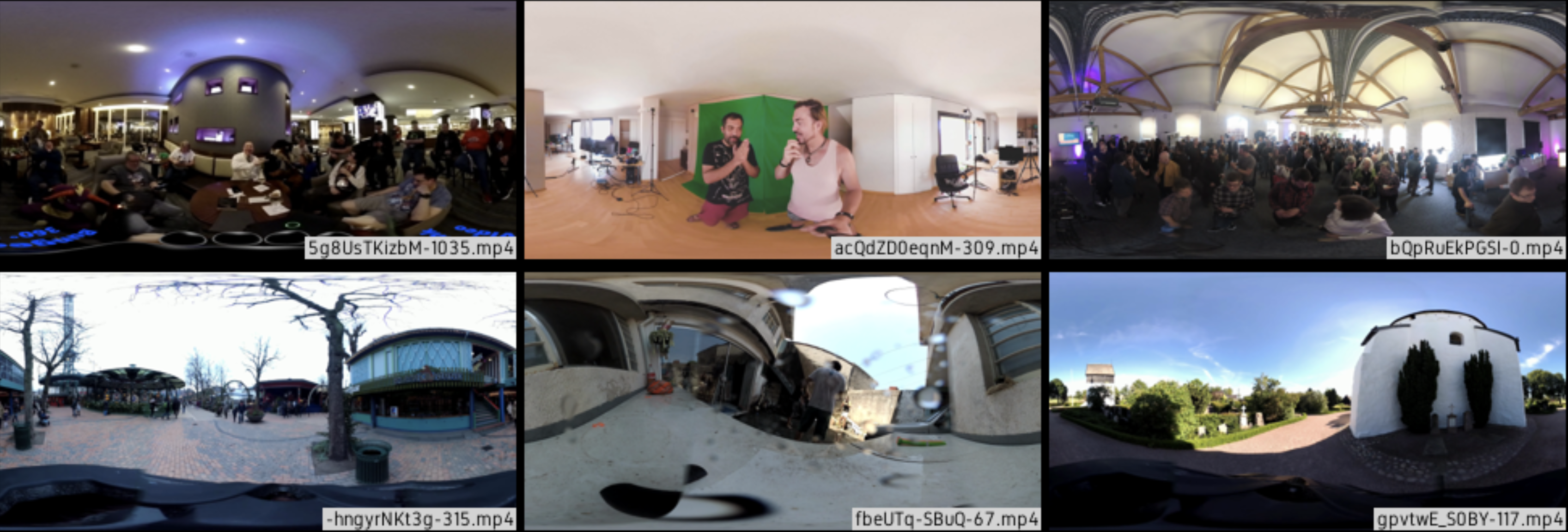}
    \caption{Sample frames from the \textit{Human Speech} category. Scenes involve multi-speaker conversations and interactions in diverse settings. Spatial audio promotes more focused fixations around the speaking individuals, resulting in more concentrated attention patterns, as further illustrated in Fig.~\ref{fig:audio-modalities-on-fixation-speech}.}
    \label{data:frames-speech}
\end{figure*}

\begin{figure*}[!t]
    \centering
    \includegraphics[trim={0 0 0 0}, clip, height=5.8cm]{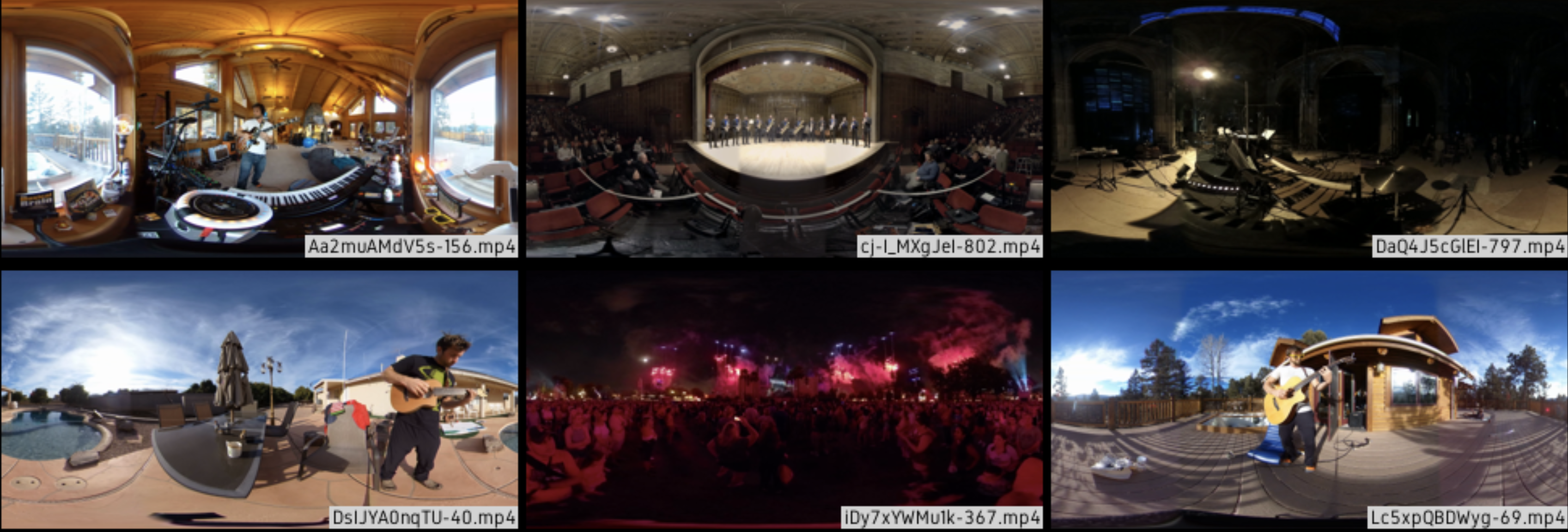}
    \caption{Sample frames from the \textit{Music} category. These include live performances and concerts with highly dynamic visuals and distributed sound sources. Spatial audio encourages more focused fixations around relevant sound sources, as illustrated in Fig.~\ref{fig:audio-modalities-on-fixation-music}.}
    \label{data:frames-music}
\end{figure*}

\begin{figure*}[!t]
    \centering
    \includegraphics[trim={20 0 0 0}, clip, height=6cm]{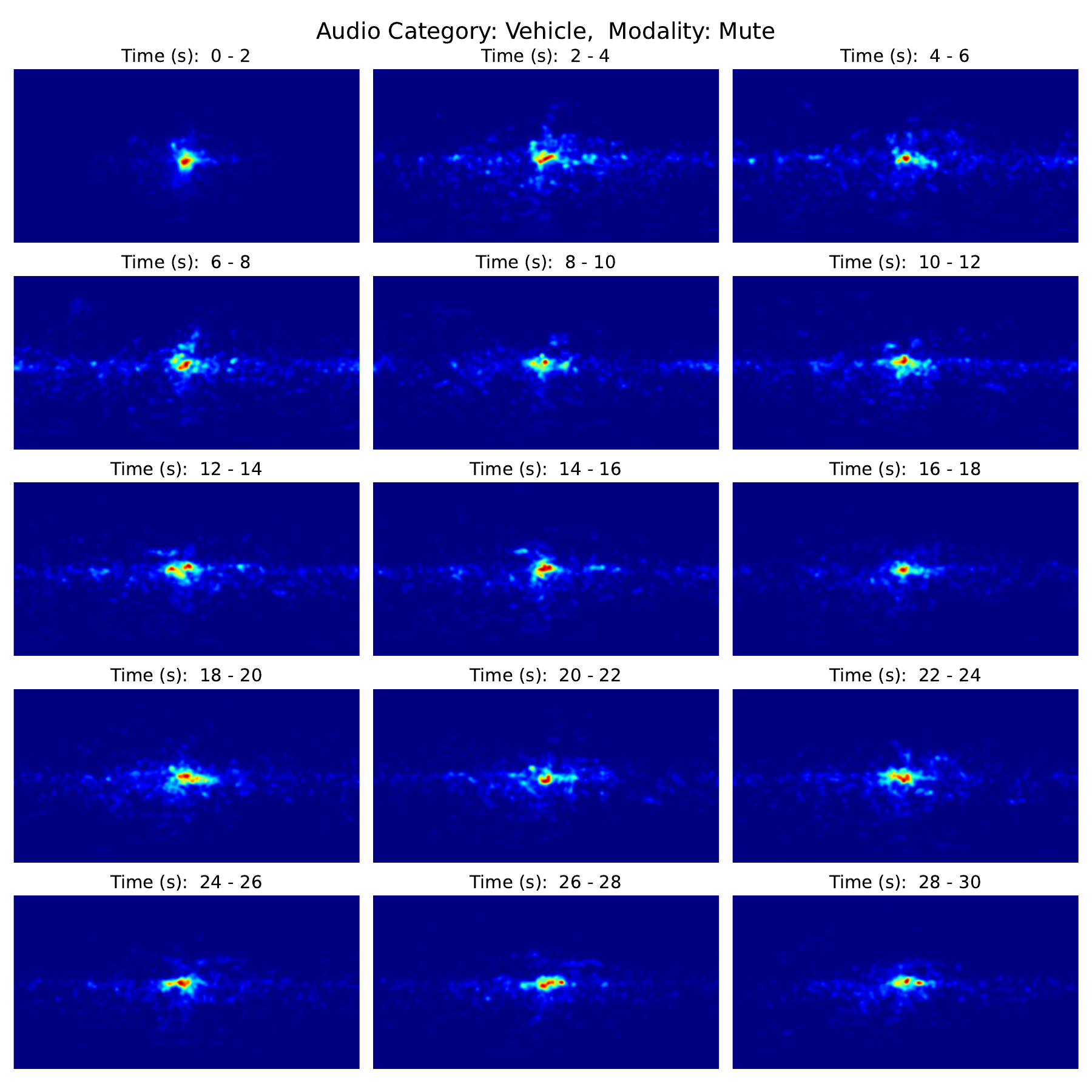}
    \includegraphics[trim={0 0 0 0}, clip, height=6cm]{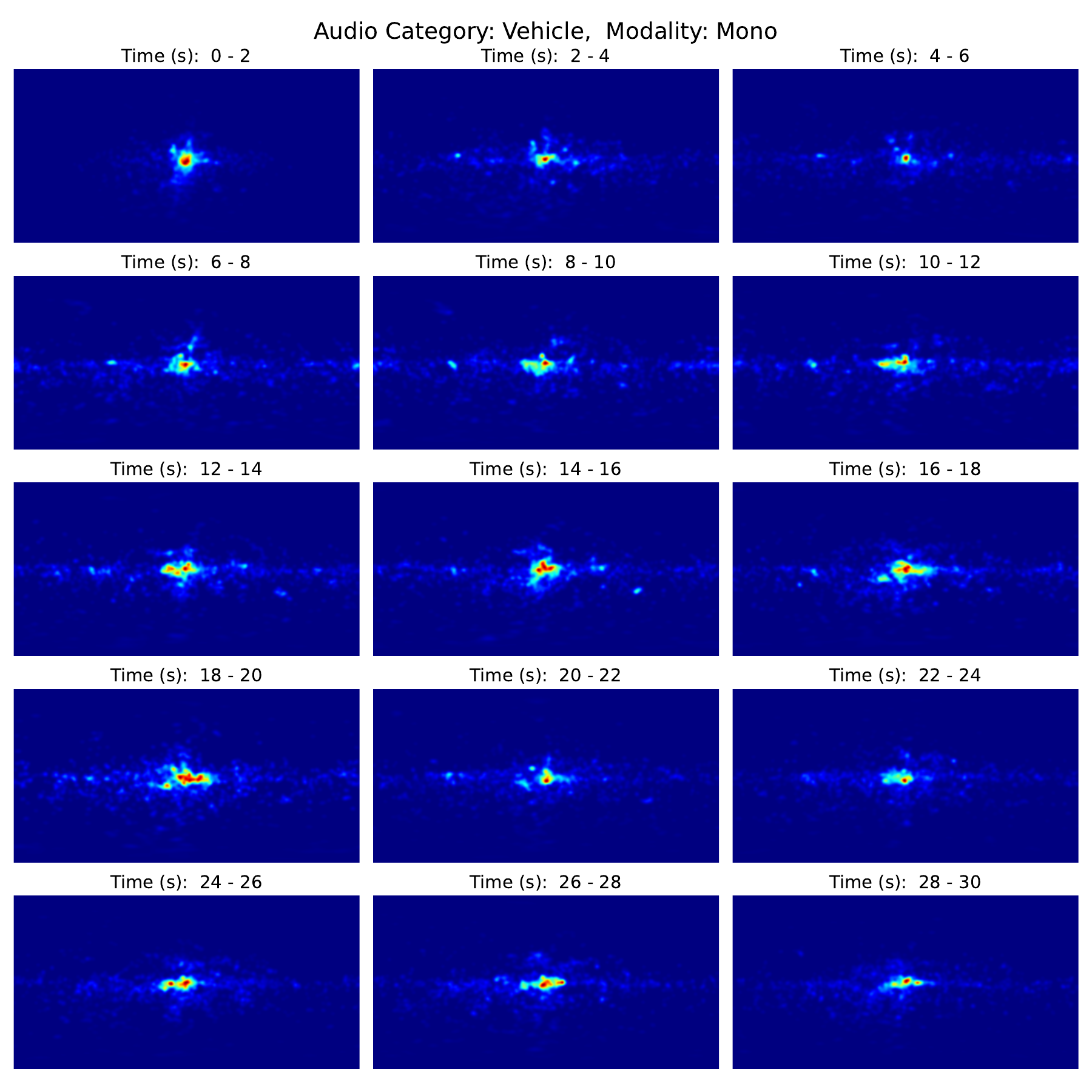}
    \includegraphics[trim={0 0 20 0}, clip, height=6cm]{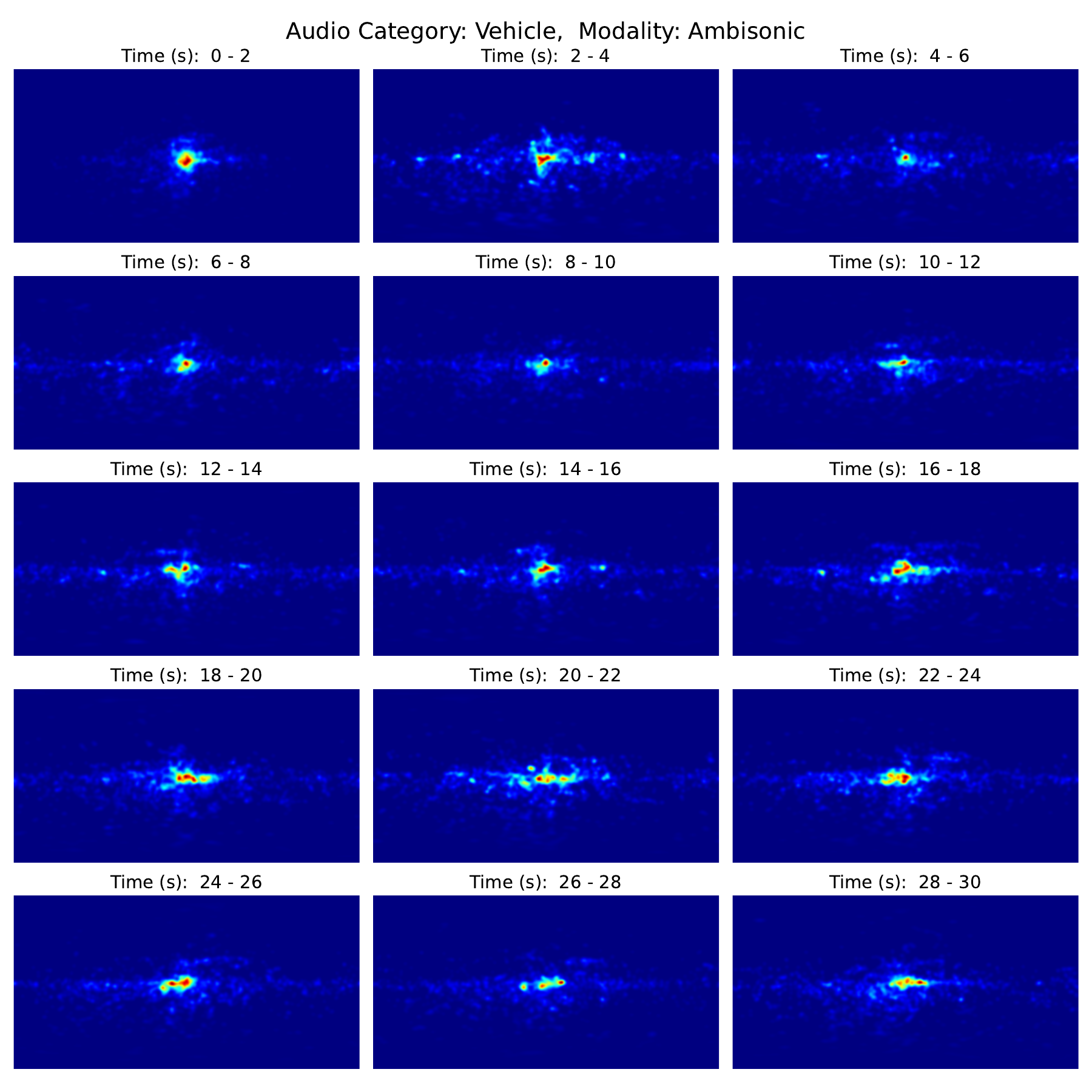}
    \caption{Average fixation density maps for the \textit{vehicle} category across different time intervals (2-secs windows) under \textit{mute}, \textit{mono}, and \textit{ambisonics} audio conditions. Fixations mainly suggest a stronger center bias due to the forward-facing driving behavior.}
    \label{fig:audio-modalities-on-fixation-vehicle}
\end{figure*}

\begin{figure*}[!t]
    \centering
    \includegraphics[trim={20 0 0 0}, clip, height=6cm]{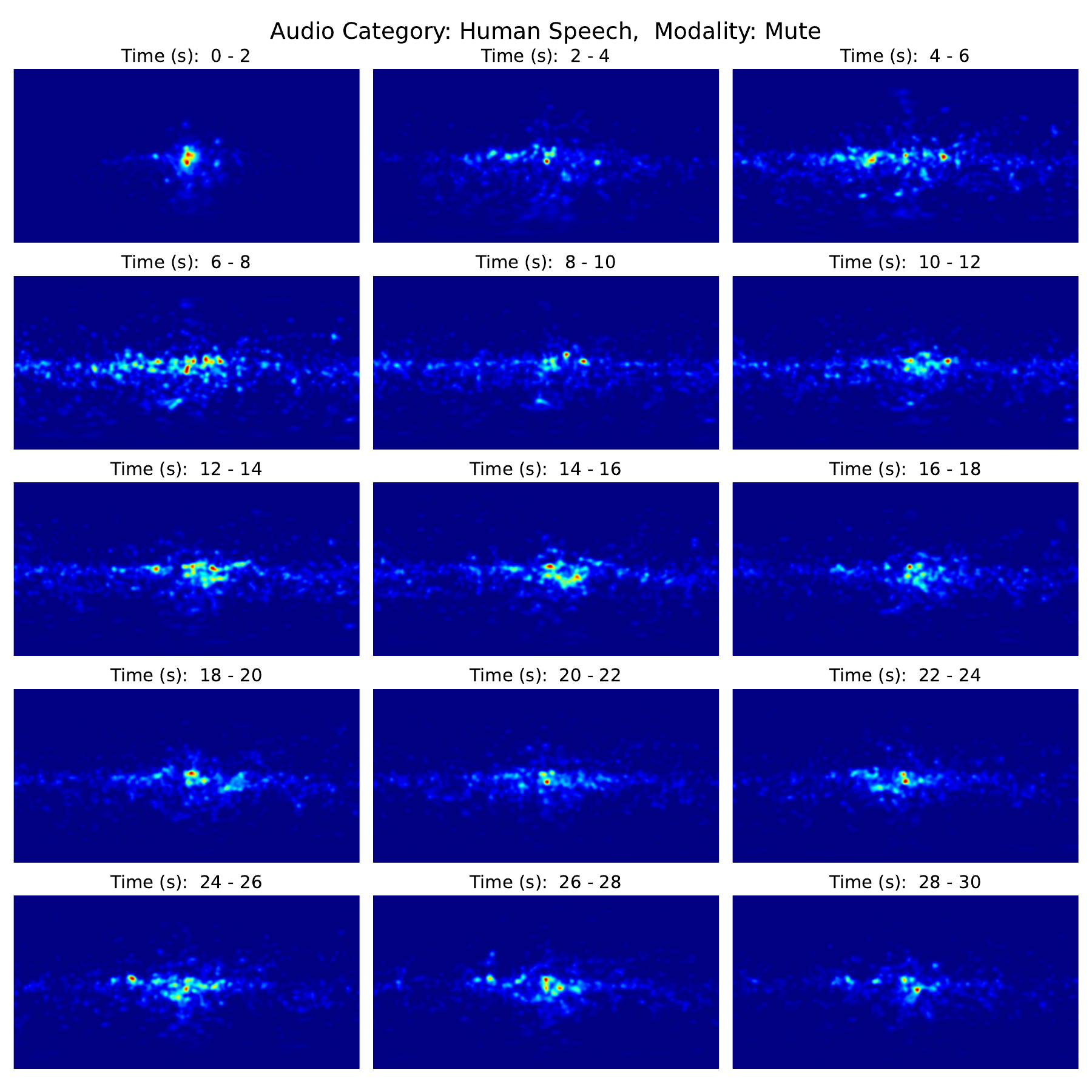}
    \includegraphics[trim={0 0 0 0}, clip, height=6cm]{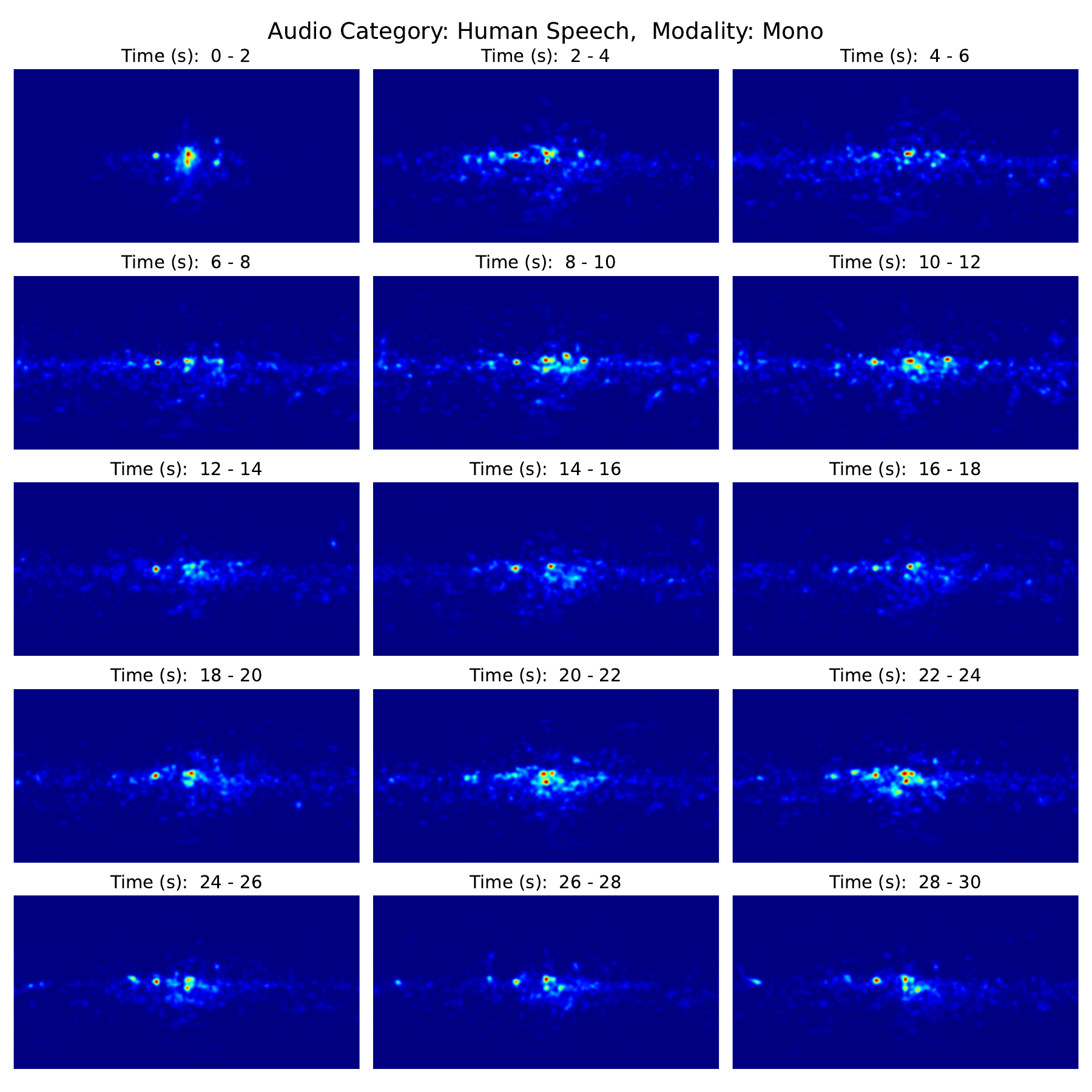}
    \includegraphics[trim={0 0 20 0}, clip, height=6cm]{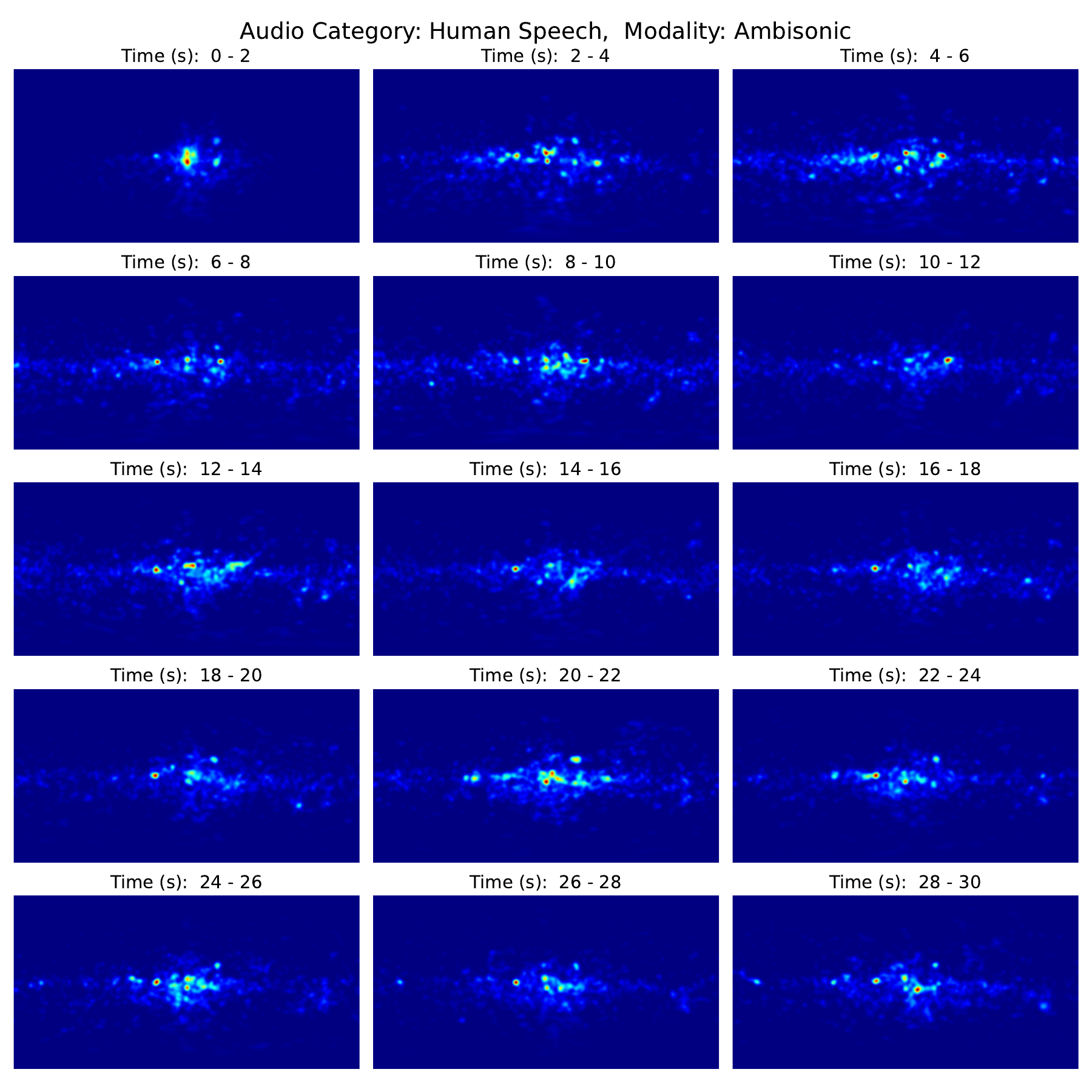}
    \caption{Average fixation density maps for the \textit{speech} category across different time intervals (2-secs windows) under \textit{mute}, \textit{mono}, and \textit{ambisonics} audio. Compared to vehicle scenes, fixations here are more evenly distributed and driven by the existence of mono and spatial audio.}
    \label{fig:audio-modalities-on-fixation-speech}
\end{figure*}

\begin{figure*}[!t]
    \centering
    \includegraphics[trim={20 0 0 0}, clip, height=6cm]{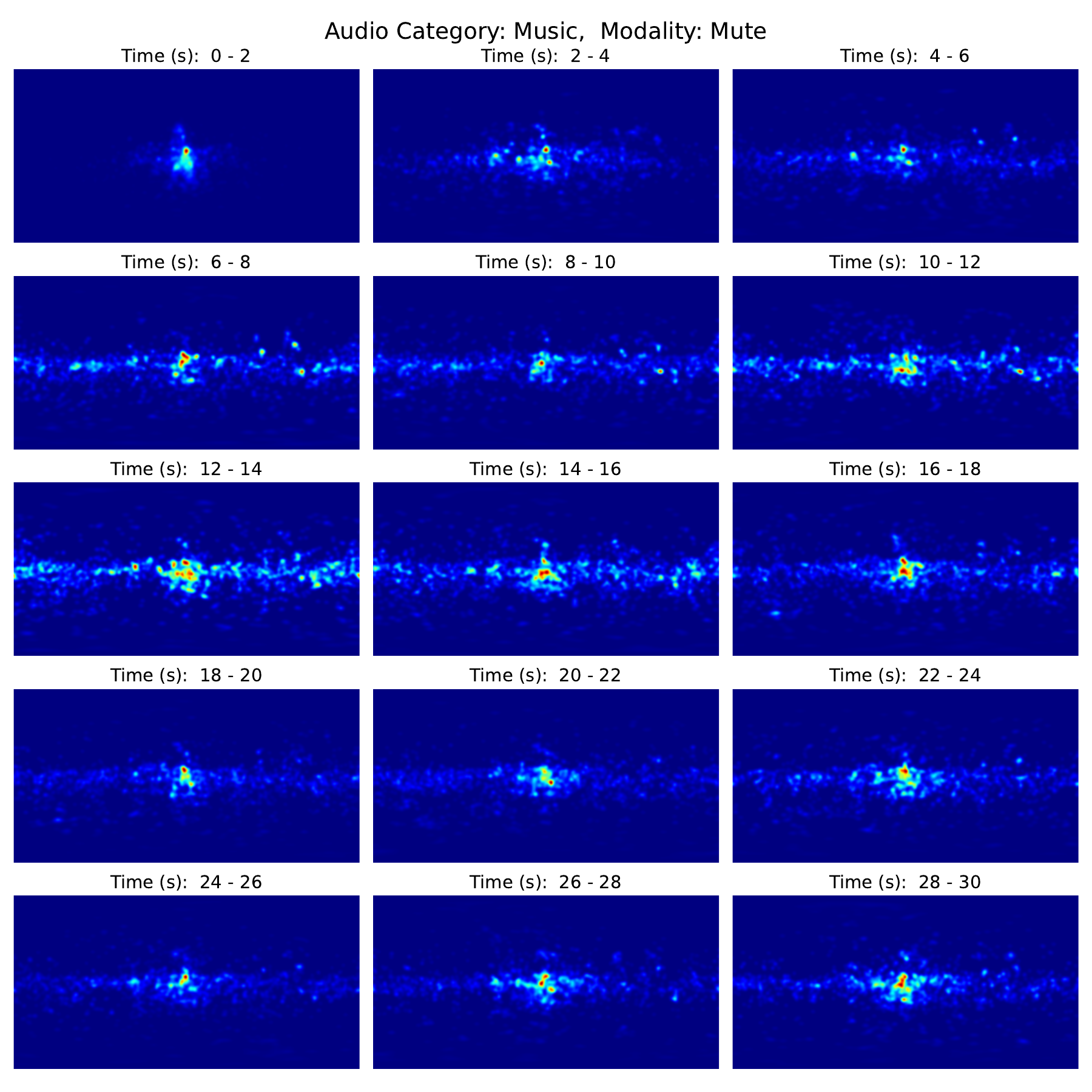}
    \includegraphics[trim={0 0 0 0}, clip, height=6cm]{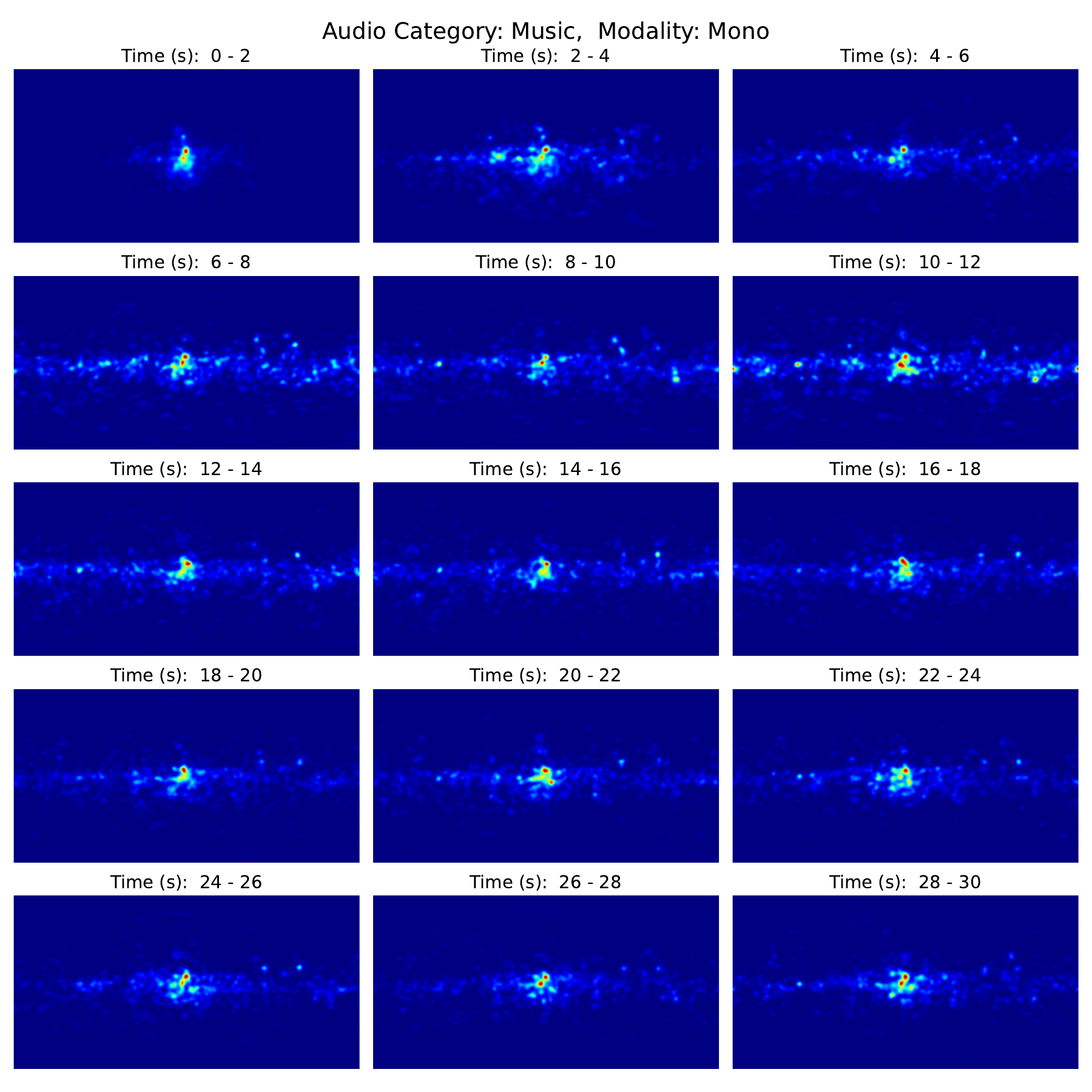}
    \includegraphics[trim={0 0 20 0}, clip, height=6cm]{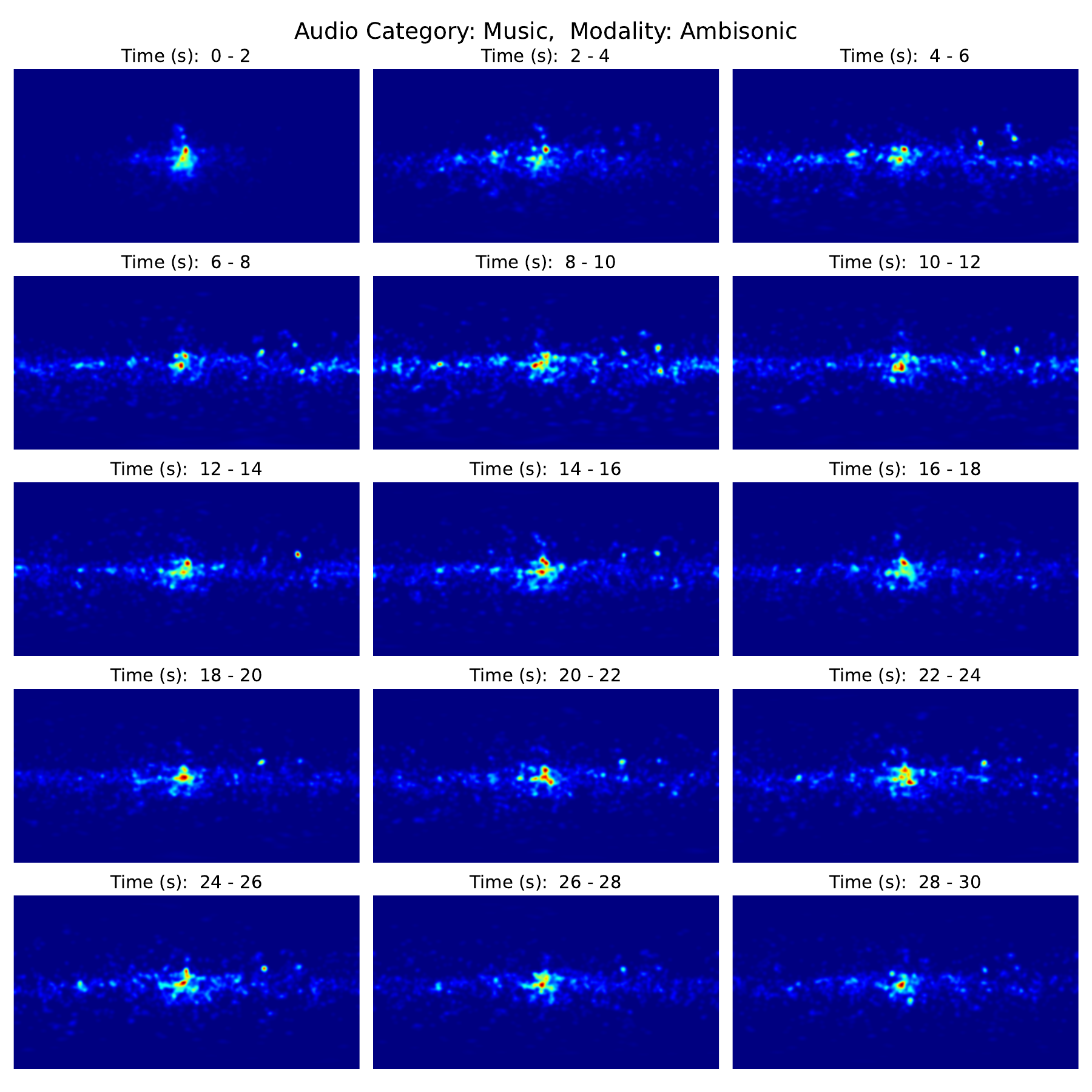}
    \caption{Average fixation density maps for the \textit{music} category across different time intervals (2-secs windows) under \textit{mute}, \textit{mono}, and \textit{ambisonics} audio. The presence of spatial audio helps guide attention toward musicians or instruments, whereas fixations under mute audio are more diffuse.}
    \label{fig:audio-modalities-on-fixation-music}
\end{figure*}

To further examine the presence and variability of center bias in our dataset, we visualize example frames and fixation patterns across three distinct audio categories: \textit{vehicle}, \textit{speech}, and \textit{music}. As shown in Fig.~\ref{data:frames-vehicle}-\ref{data:frames-music}, each category captures diverse visual contexts, ranging from structured environments (e.g., car interiors) to more unconstrained scenes (e.g., live concerts or social gatherings). In the \textit{vehicle} category, the camera is typically placed at the front of the vehicle, facing the road ahead. This layout introduces a strong content-driven center bias, as participants tend to fixate in the forward driving direction, near the horizon. Average fixation density maps given in Fig.~\ref{fig:audio-modalities-on-fixation-vehicle} confirm this behavior, especially under spatial audio. In contrast, the \textit{speech} and \textit{music} categories feature more dynamic and visually diverse content, often with multiple people or performers distributed across the scene. As reflected in the average fixation maps given in Fig.~\ref{fig:audio-modalities-on-fixation-speech} and~\ref{fig:audio-modalities-on-fixation-music}), subjects explore wider regions of the visual field, and the fixation distribution becomes more dispersed.

\subsection{Effect of Audio Modalities on Fixation}

To quantify the role of auditory context in directing attention, we compare fixation patterns under three audio modalities: \textit{mute}, \textit{mono}, and \textit{ambisonics}. Across all content categories, spatial audio (ambisonics) consistently results in more concentrated and semantically aligned fixations, typically near the sound sources. In contrast, mute conditions often lead to less focused exploration, particularly around the equator, indicating a lack of auditory cues to guide viewer attention.

\begin{figure*}[!t]
    \centering
    \includegraphics[trim={30 25 40 35}, clip, width=0.775\textwidth]{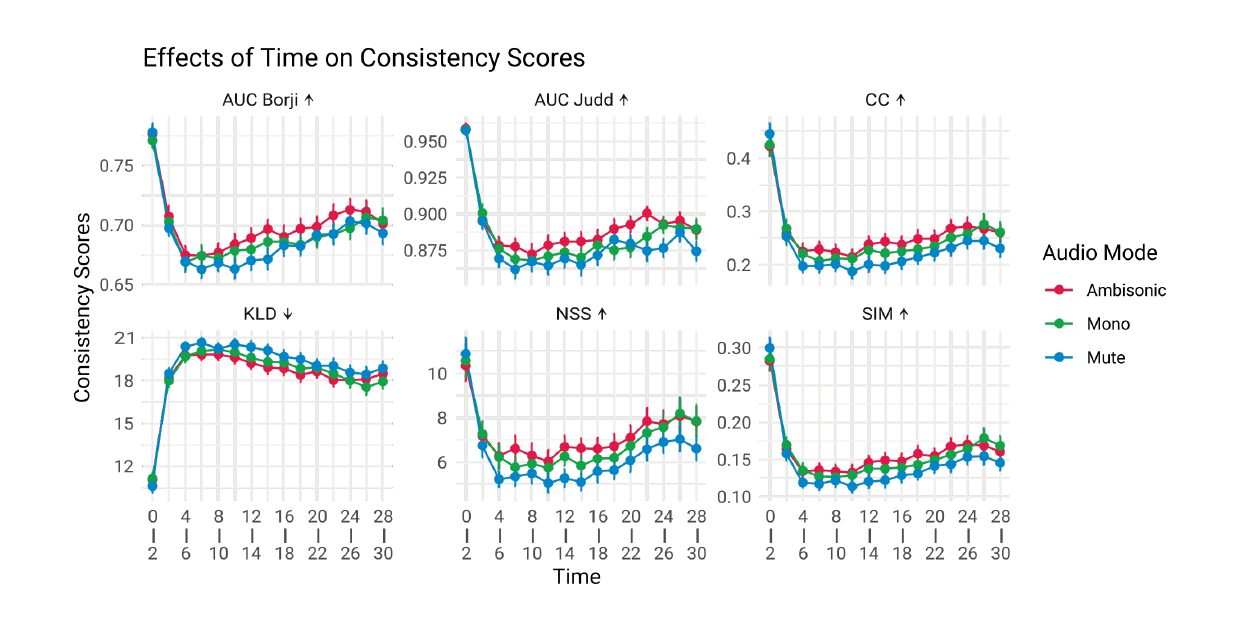}
    \caption{\textbf{Inter-subject consistency scores of viewer fixations under three different audio conditions, mute, mono, and ambisonics, measured over 2-secs time windows.} Higher consistency in spatial audio (ambisonics) highlights its stronger influence on guiding collective visual attention.}
    \label{fig:audio-modalities-on-fixation}
\end{figure*}

To further quantify the influence of audio modality on collective viewing behavior, we analyze inter-subject consistency across viewers under different audio conditions. As shown in Fig.~\ref{fig:audio-modalities-on-fixation}, consistency scores are computed over 2-second time windows across all participants. The results reveal a consistent trend: spatial audio (ambisonics) yields significantly higher inter-subject consistency compared to mono and mute conditions. This suggests that directional auditory cues not only enhance individual saliency but also help synchronize visual attention across viewers. Notably, consistency scores are highest at the onset of each video, likely due to the standardized ERP-centered initial viewport. A noticeable dip in consistency occurs between 4 and 10 seconds, reflecting a phase of exploratory behavior as viewers diverge in their gaze patterns. This is followed by a gradual realignment of attention, particularly under the influence of spatial audio, indicating that spatial cues guide viewers back toward semantically relevant regions. These temporal dynamics highlight the crucial role of audio modality in modulating both the distribution and convergence of attention over time, further emphasizing the importance of incorporating spatial audio in 360° saliency modeling.

\renewcommand\thefigure{D.\arabic{figure}}    
\renewcommand\thetable{D.\arabic{table}}    
\renewcommand\theequation{D.\arabic{equation}}

\setcounter{figure}{0}    
\setcounter{table}{0}    
\setcounter{equation}{0}  

\section{Use Case: Saliency-Guided Omnidirectional Video Quality Assessment}\label{appdx-vqa}
This section shows the evaluation of our method and the competing approaches on a downstream task, which involves assessing the visual quality of omnidirectional videos. 

In Table~\ref{vqaodv}, we report the performance of two PSNR variants compared to ground-truth DMOS values in the VQA-ODV~\cite{Li:2018:BGV:3240508.3240581} dataset. Each row corresponds to saliency weights that supply human-perceptual information to PSNR and WS-PSNR metrics. The ground truth head movement (HM) maps refer to the viewports that human subjects have viewed while rating the visual quality of ODVs. Predicting saliency maps that better capture human head movements will result in better performance in the PSNR metrics. The table demonstrates that our proposed saliency prediction model better highlights the perceptually important regions in $360^\circ$ videos compared to the state-of-the-art, for omnidirectional video quality assessment downstream task.

Following the prior work~\cite{yun2022panoramic}, the saliency-weighted PSNR and WS-PSNR values are calculated as:
\begin{equation}\label{eqn:psnr}
\begin{split}
  \text{PSNR}_{\textit{}}  = & 10\log_{10} \left( \frac{Y_{max}^2 \cdot \sum_{p\in\mathbb{P}}w_{sal}(p)}{\sum_{p\in\mathbb{P}}(Y(p)- Y'(p))^2 \cdot w_{sal}(p)} \right)
\end{split}
\end{equation}
\begin{equation}\label{eqn:ws-psnr}
\begin{split}
  \text{WS-PSNR}_{\textit{}} = & 10\log_{10} \left( \frac{Y_{max}^2 \cdot \sum_{p\in\mathbb{P}}w_{sal}(p) \cos{\theta}_p}{\sum_{p\in\mathbb{P}}(Y(p)- Y'(p))^2 \cdot w_{sal}(p)\cos{\theta}_p} \right)
\end{split}
\end{equation}
where $Y_{max}$ is the maximum intensity of the frames, $Y(p)$ and $Y'(p)$ denote the intensities for of pixel $p$ in the reference and impaired videos, and $\theta_p$ is the latitude at pixel~$p$.

\begin{table}[!t]
\resizebox{\linewidth}{!}{ 
  \begin{tabular}{l@{$\;$}c@{$\;$}c@{$\;$}cc@{$\;$}c@{$\;$}c}
    \toprule
    \multirow{3}{*}{{Weight}} &
      \multicolumn{3}{c}{PSNR} &
      \multicolumn{3}{c}{WS-PSNR~\cite{Sun2017WeightedtoSphericallyUniformQE}} \\
    & PCC$\uparrow$ & SRCC$\uparrow$ & RMSE$\downarrow$ & PCC$\uparrow$ & SRCC$\uparrow$ & RMSE$\downarrow$ \\
    \midrule
    None                & 0.650 & 6.664 & 7.502 & 0.671 & 0.686 & 7.233\\
    PAVER~\cite{yun2022panoramic}               & 0.661 & 0.667 & 7.481 & 0.679 & 0.691 & 6.914\\
    Djilali et.~al.~\cite{Djilali_2021_ICCV}    & 0.648 & 0.721 & 7.336 & 0.684 & 0.721 & 6.829\\
    SalViT360 (ours)    & 0.688 & 0.733 & 7.295 & 0.689 & 0.737 & 6.673\\
    \midrule
    HM (Supervised)     & 0.764 & 0.759 & 6.601 & 0.759 & 0.756 & 6.612\\
    \bottomrule
  \end{tabular}}
  \caption{\textbf{Comparison with state-of-the-art} saliency models for \textit{saliency-guided omnidirectional video quality assessment} on VQA-ODV~\cite{Li:2018:BGV:3240508.3240581} dataset, as a downstream task.}
  \label{vqaodv}
\end{table}

\end{document}